\documentclass[10pt,twocolumn,letterpaper]{article}

\usepackage{cvpr}              % To produce the CAMERA-READY version

\newcommand{\Equation}{Eq.~}

\newcommand{\Img}{\ensuremath{I}}                      % Input images
\newcommand{\HImg}{\ensuremath{H}}                     % Image Height
\newcommand{\WImg}{\ensuremath{W}}                     % Image Widith
\newcommand{\HCoarse}{\ensuremath{h}}                     % Image Height
\newcommand{\WCoarse}{\ensuremath{w}}                     % Image Widith
\newcommand{\NBlocks}{\ensuremath{L}}                  % Number of transformer blocks

\newcommand{\DMaps}{\ensuremath{D}}         % Depth maps
\newcommand{\Token}[1]{\ensuremath{t_{#1}}} % Token

\newcommand{\SNormal}{\ensuremath{\mathbf{n}}}         % Surface normal

\newcommand{\Patch}{\ensuremath{J}}         % Patches
\newcommand{\HPatch}{\ensuremath{H'}}        % Patch Height
\newcommand{\WPatch}{\ensuremath{W'}}        % Patch Widith
\newcommand{\Offset}[1]{\ensuremath{\Delta^{#1}}} 
\newcommand{\NPatch}{\ensuremath{M}}        % Number of Patches
\newcommand{\Prob}[1]{\ensuremath{\mathcal{P}}\big( \text{config} = {#1}\times {#1} \big)}    %  Probability law
\newcommand{\vProb}[1]{\ensuremath{p}_{#1}}        % Probability of a configuration

\newcommand{\Loss}[1] {\mathcal{L}_{\text{#1}}}

\usepackage{bm} 

\newcommand{\vect}[1]{\bm{#1}}

\newcommand{\patchidx}{k}

\newcommand{\RoPE}{\mathrm{RoPE}}

\definecolor{cvprblue}{rgb}{0.21,0.49,0.74}
\usepackage[pagebackref,breaklinks,colorlinks,allcolors=cvprblue]{hyperref}
\usepackage{multirow}
\usepackage{tikz}
\usepackage{graphicx}
\usepackage{overpic}
\usepackage{adjustbox}
\usepackage{xcolor}
\usetikzlibrary{positioning}

\usepackage{makecell}
\usepackage{graphicx}
\usepackage{amsmath, bm}
\usepackage{tabularx,booktabs}
\usepackage{adjustbox}
\usepackage{multirow}
\usepackage{graphicx,xcolor,colortbl} 
\usepackage{array}
\newcolumntype{C}[1]{>{\centering\arraybackslash}p{#1}}
\newcolumntype{L}[1]{>{\arraybackslash}p{#1}}
\newcolumntype{M}[1]{>{\centering\arraybackslash}m{#1}}

\def\paperID{} % *** Enter the Paper ID here
\def\confName{CVPR}
\def\confYear{2026}
\usepackage{indentfirst}

\title{Any Resolution Any Geometry: From Multi-View To Multi-Patch}

\author{
Wenqing Cui$^{1}$, 
Zhenyu Li$^{1\dagger}$, 
Mykola Lavreniuk$^{2\dagger}$,
Jian Shi$^1$, 
Ramzi Idoughi$^1$,\\ 
Xiangjun Tang$^1$, 
Peter Wonka$^1$\\
\\
$^1$KAUST, $^2$Space Research Institute NASU-SSAU \\
{\small $^\dagger$Equal contribution} \\
{\small \url{https://dreamaker-mrc.github.io/Any-Resolution-Any-Geometry}} \\
}

\begin{document}
\twocolumn[{%
\renewcommand\twocolumn[1][]{#1}%
\maketitle
\vspace{-30pt}
\begin{center}
    \centering
    \captionsetup{type=figure}
    \includegraphics[width=\textwidth]{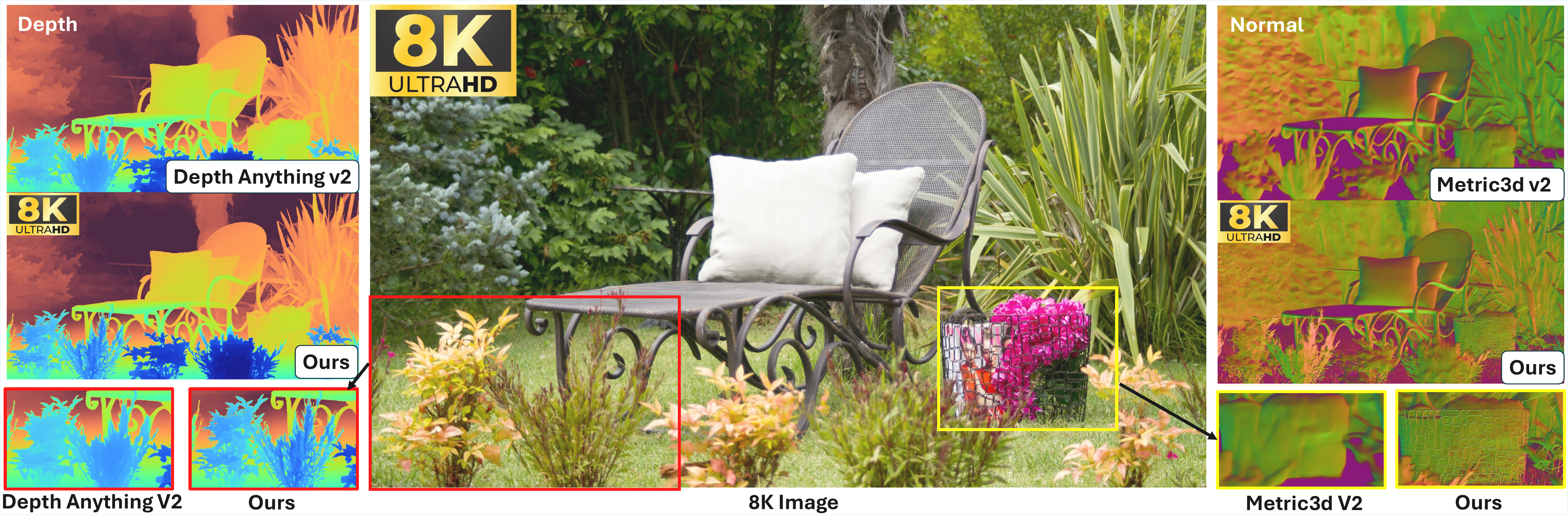}
    \label{fig:teaser}
    \vspace{-20pt}
    \caption{
    \textbf{Teaser.} On an in-the-wild 8K image, our method produces high-resolution depth and normal maps with sharp boundaries and globally consistent geometry. Compared to zero-shot foundation models such as DepthAnything V2 ~\cite{yang2025depthanythingv2} and Metric3D V2~\cite{hu2025metric3dv2}, our approach better recovers thin structures and high-frequency textures. The zoom-in regions below highlight the improved fine-detail preservation and depth–normal consistency achieved by our method.
    }

\end{center}%
}]

\begin{abstract}
Joint estimation of surface normals and depth is essential for holistic 3D scene understanding, yet high-resolution prediction remains difficult due to the trade-off between preserving fine local detail and maintaining global consistency. To address this challenge, we propose the \textbf{U}ltra \textbf{R}esolution \textbf{G}eometry \textbf{T}ransformer \textbf{(URGT)}, which adapts the Visual Geometry Grounded Transformer (VGGT) into a unified \emph{multi-patch} transformer for monocular high-resolution depth--normal estimation. A single high-resolution image is partitioned into patches that are augmented with coarse depth and normal priors from pre-trained models, and jointly processed in a single forward pass to predict refined geometric outputs. Global coherence is enforced through \emph{cross-patch attention}, which enables long-range geometric reasoning and seamless propagation of information across patches within a shared backbone. To further enhance spatial robustness, we introduce a \emph{GridMix patch sampling} strategy that probabilistically samples grid configurations during training, improving inter-patch consistency and generalization. Our method achieves state-of-the-art results on UnrealStereo4K, jointly improving depth and normal estimation, reducing AbsRel from 0.0582 to 0.0291, RMSE from 2.17 to 1.31, and lowering mean angular error from 23.36$^\circ$ to 18.51$^\circ$, while producing sharper and more stable geometry. The proposed multi-patch framework also demonstrates strong zero-shot and cross-domain generalization and scales effectively to very high resolutions, offering an efficient and extensible solution for high-quality geometry refinement.

\end{abstract}
    
\vspace{-20pt}
\section{Introduction}
\label{sec:intro}

High-resolution prediction of depth maps and surface normals is crucial for modern 3D scene understanding, yet remains fundamentally challenging. Depth maps capture the global spatial layout of a scene, while surface normals encode fine-grained surface orientations; together, these complementary cues support a wide range of downstream tasks such as 3D reconstruction~\cite{dai2024high, turkulainen2025dn} and scene segmentation~\cite{fan2020sne}. However, producing \emph{high-resolution} geometry requires simultaneously preserving sharp object boundaries and thin structures while maintaining globally consistent depth and normal fields across the image. In practice, most recent joint depth--normal estimation models~\cite{hu2025metric3dv2, fu2024geowizard} operate at relatively low input resolutions due to memory and computational constraints, which limits their ability to recover fine-scale details and leads to noticeable degradation when applied to high-resolution imagery.

Patch-based refinement methods, such as PatchFusion~\cite{li2024patchfusion} and PatchRefiner~\cite{li2024patchrefiner}, have been developed to leverage high-resolution imagery and enhance geometric quality. Nonetheless, these approaches remain limited: PatchRefiner refines patches \emph{iteratively} and largely in isolation, leading to inconsistent boundaries across neighboring patches. In addition, existing high-resolution pipelines are typically designed for a depth prediction only, making joint depth--normal estimation difficult to realize at scale. These limitations highlight the need for a unified, efficient \emph{high-resolution joint} depth--normal model that goes beyond depth-only patch refinement, performing global reasoning across all regions while maintaining pixel-level geometric precision and seamless inter-patch continuity.

In parallel, multi-view transformer architectures have recently achieved strong results in dense geometry estimation. Models such as DUSt3R and the Visual Geometry Grounded Transformer (VGGT)~\cite{wang2025vggt, wang2024dust3rgeometric3dvision} jointly encode many views within a unified backbone and use attention to propagate information globally, enabling accurate 3D reconstruction while efficiently handling multiple inputs. This set-based formulation demonstrates that geometry prediction can scale by reasoning across a collection of inputs rather than processing each one independently. However, these models are primarily designed for multi-view scenarios at moderate resolutions and do not directly tackle joint high-resolution depth--normal estimation from a single image. In this work, we draw inspiration from their design and reinterpret a single high-resolution image as a set of patches, effectively treating patches as \emph{virtual views}, and transfer the multi-view paradigm to a \emph{multi-patch} setting that enables global reasoning across all regions while preserving high-fidelity local geometry.

We address these challenges by adapting VGGT~\cite{wang2025vggt} into a unified backbone for predicting geometric coupled high-resolution depth and normal from a single image. On top of the matured multi-view frameworks, we propose a \emph{multi-patch} formulation, where a high-resolution image is partitioned into patches that are jointly processed in a single forward pass. Each patch is augmented with coarse depth and normal priors from pre-trained models, and the transformer refines these into high-resolution geometric outputs. Crucially, cross-patch attention enables long-range information exchange across all patches, allowing the network to enforce global consistency while recovering fine-grained details. Building on this design, we propose a unified model \textbf{U}ltra \textbf{R}esolution \textbf{G}eometry \textbf{T}ransformer \textbf{(URGT)} that simultaneously predicts geometric coupled depth and surface normals, leveraging their inherent geometric relationships to improve accuracy and enforce physically consistent scene geometry.

The main contributions of this paper are as follows:
\begin{itemize}
    \item \textbf{A Unified Model for geometric coupled High-Resolution Geometry Prediction}.  
    We propose a novel multi-patch transformer that predicts geometric coupled high-resolution depth and surface normals from a single high-resolution image (\eg 4K).
    \item \textbf{GridMix Patch Sampling Strategy}.  
    Due to the scarcity of high-resolution data, we introduce a GridMix patch sampling strategy that probabilistically samples different patch grid configurations during training.
    \item \textbf{Scalability to Arbitrary Resolutions}: Our method  supports the prediction of geometric coupled high-resolution (\eg 4K, 6K, 8K) depth maps and surface normals, offering flexibility and practicality for real-world applications without compromising accuracy or requiring resolution-specific training.
    \item \textbf{Empirical Validation and Insights}.  
    Through extensive experiments, we demonstrate that our approach outperforms state-of-the-art methods in both estimation quality and computational efficiency on high-resolution benchmarks, and we provide ablation studies that highlight the benefits of multi-patch reasoning, cross-patch attention, and GridMix training.
\end{itemize}

\vspace{-8pt}

\section{Related work}
\label{sec:related work}

%-------------------------------------------------------------------------
\subsection{High-Resolution Depth Estimation}

High-resolution depth estimation seeks to recover fine details while maintaining globally consistent scene geometry. SMD-Net~\cite{u4k} employed local implicit functions~\cite{chen2021learning} to enhance boundary accuracy, while Dai et al.~\cite{dai2023multi} introduced Poisson-based fusion in a self-supervised framework. SDDR~\cite{li2025self} leveraged self-distilled edge refinement to reduce local noise and improve structural consistency.

Patch-based strategies have become a dominant solution for scaling to 2K–4K inputs. BoostingDepth~\cite{miangoleh2021boosting}, PatchFusion~\cite{li2024patchfusion}, PatchRefiner~\cite{li2024patchrefiner}, and PatchRefiner V2~\cite{li2025patchrefinerv2fastlightweight} divide images into patches and merge their predictions. While effective, such methods often produce depth discontinuities near patch boundaries due to limited cross-patch interaction. Existing remedies, such as test-time ensembles, heuristic blending, or consistency losses, help reduce artifacts but significantly increase inference cost.

PRO~\cite{kwon2025onelook} jointly processes grouped overlapping patches and introduces patch-consistency supervision with bias-free masking, enabling single-pass inference per patch and reducing boundary artifacts. Although this improves local continuity, its ability to enforce global geometric coherence remains constrained by the limited receptive field of patch groups.

\begin{figure*}[!ht]
    \centering
    \includegraphics[width=1\linewidth]{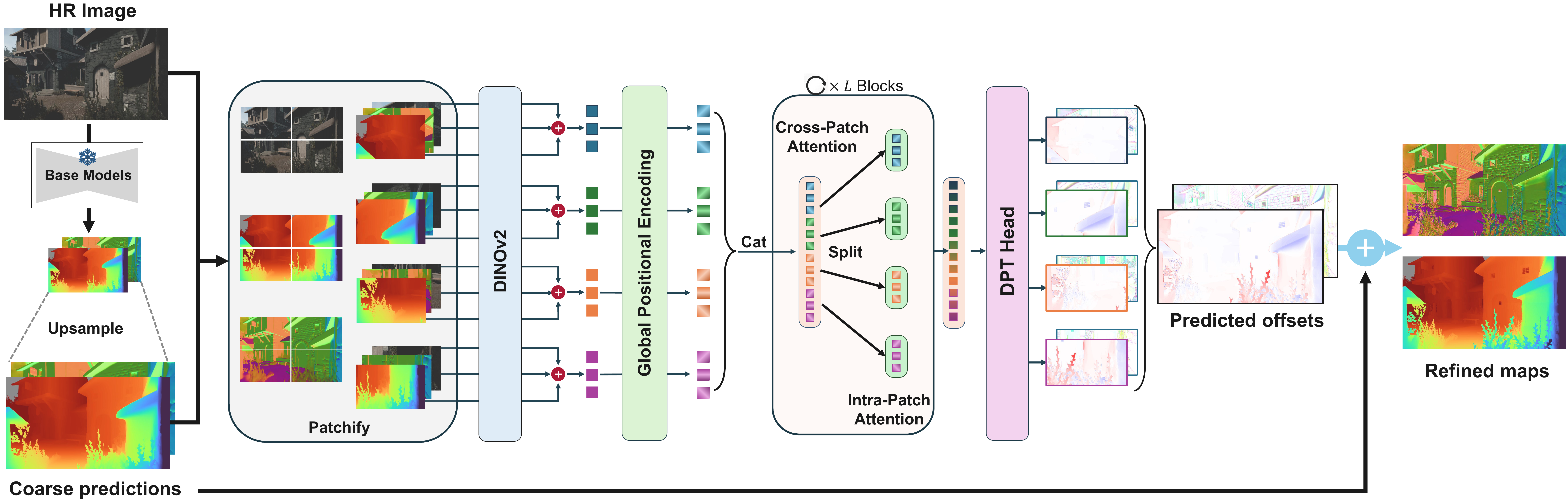}
    \hspace{-10pt}
    \caption{
    \textbf{Multi-patch transformer pipeline for high-resolution geometry refinement.}
    A coarse depth/normal prediction is first obtained and upsampled, after which the image and coarse outputs are patchified and encoded with DINOv2 and global positional embeddings. All patch tokens are jointly processed through transformer blocks with intra- and cross-patch attention, enabling global geometric reasoning. A DPT head predicts offsets that are fused with the coarse input to produce a globally consistent, detail-preserving high-resolution output.
    }
    \vspace{-10pt}

    \label{fig:pipeline}
\end{figure*}

%-------------------------------------------------------------------------
\subsection{Joint Estimation of Depth and Surface Normal}

Depth and surface normal provide complementary geometric cues: depth represents global scene structure, while normals encode local surface orientation. Their joint estimation enhances geometric consistency and benefits downstream tasks such as 3D reconstruction and relighting. Early supervised approaches~\cite{qi2018geonet, li2020geonetpp, bae2021estimating} enforced explicit depth-normal consistency terms or iterative mutual refinement, but performance was limited by small datasets with sparse or noisy normal annotations.

Large-scale zero-shot approaches have revitalized joint estimation by leveraging stronger geometric priors. GeoWizard~\cite{fu2024geowizard} employs diffusion-based priors to simultaneously infer depth and normals from a single image, yielding coherent geometry at the cost of high computational overhead. Metric3D v2~\cite{hu2025metric3dv2} proposes a geometry foundation model that jointly estimates metric depth and normals using a canonical camera transformation and a learned depth–normal optimization module. By learning normal cues from large-scale depth data, it achieves efficient training and strong generalization across domains.

Overall, recent advances move toward unified architectures that leverage geometric coupling to achieve globally consistent and detail-preserving predictions. However, most existing models still operate at low resolutions or rely on patch-wise refinement without global reasoning, motivating methods that can jointly estimate depth and normals with seamless continuity at high resolutions.

\section{Method}

\begin{figure*}[!ht]
    \centering
    \includegraphics[width=.8\linewidth]{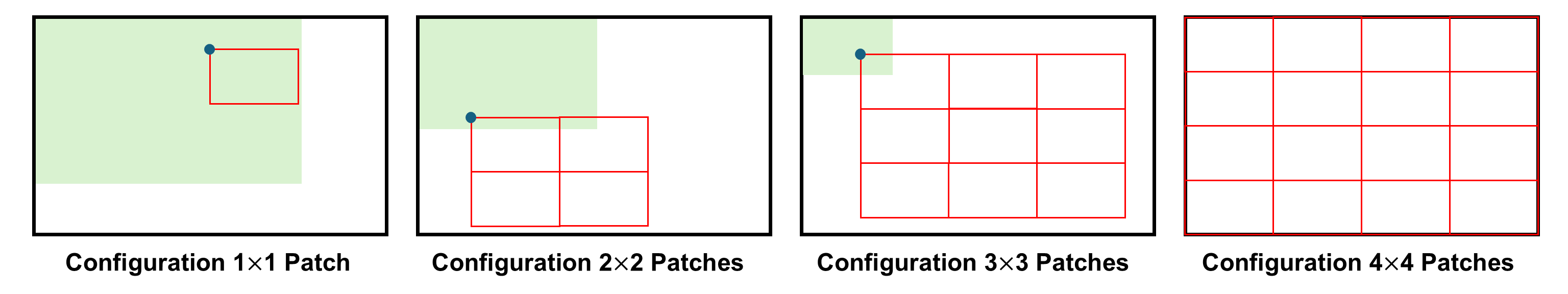}
    \vspace{-10pt}
    \caption{Illustration of the four GridMix patch sampling strategy configurations used in the training strategy: (from left to right) a single randomly sampled patch, a $2 \times 2$ grid of patches, a $3\times3$ grid of patches, and a $4\times4$ grid covering the entire image. The red grids represent the patch arrangements, while the green regions indicate the valid areas for randomly selecting the top-left corner of the grid, ensuring the entire grid remains fully within the image boundaries.}
    \label{fig:patches}
    \vspace{-10pt}
\end{figure*}

Our proposed geometry refiner \textbf{URGT} takes a single high-resolution RGB image as input and jointly predicts its corresponding high-resolution depth and surface normal maps.
Following~\cite{li2024patchfusion,li2024patchrefiner}, we adapt a multi-patch processing strategy, enabling efficient handling of high-resolution imagery, where the image is divided into patches, and coarse geometry estimates are incorporated to refine predictions.
To jointly process multi-patch inputs, we leverage the strengths of VGGT's unified transformer-based design.
Unlike VGGT, which is designed for multi-view image processing, we focus on a single-view setting where the model reasons both within and across patches to recover detailed 3D attributes and enforce cross-patch geometric consistency.

\subsection{Formulation}
The input consists of a single high-resolution RGB image 
$\Img \in \mathbb{R}^{3 \times \HImg \times \WImg}$ and coarse estimates for depth $\DMaps^{coarse} \in \mathbb{R}^{\HCoarse \times \WCoarse}$ and surface normal $\SNormal^{coarse} \in \mathbb{R}^{3\times\HCoarse \times \WCoarse}$. In our implementation, the coarse depth map is generated using Depth-Anything v2~\cite{yang2025depthanythingv2}, while the coarse surface normal is obtained by running Metric3D v2~\cite{hu2025metric3dv2}. As these models output lower-resolution results compared to the original image, we apply bilinear upsampling to match their resolutions with that of 
\Img. These coarse predictions are used as conditional inputs to guide the network. We denote $k$-th RGB patch using $\Patch_k \in \mathbb{R}^{3 \times \HPatch \times \WPatch}$ and its aligned coarse depth crop $\DMaps^{coarse}_k$ and normal crop $\SNormal^{coarse}_k$. Each of these inputs is then independently encoded using the DINOv2~\cite{oquab2023dinov2} to obtain visual tokens $\Token{\Patch_k}$, depth tokens $\Token{\DMaps^{coarse}_k}$, and normal tokens $\Token{\SNormal^{coarse}_k}$. To construct geometry-aware representations, we sum all encoded tokens to form a unified representation:
\begin{equation}
    \Token{joint}^k = \Token{\Patch_k} + \Token{\DMaps^{coarse}_k} + \Token{\SNormal^{coarse}_k}
    \label{eq:joint_token}
\end{equation}

These fused tokens are then concatenated across all patches to form a unified input sequence, enabling the model to integrate visual and geometric information seamlessly. This sequence is fed into our model, which consisting of $\NBlocks$ blocks that combine intra-patch attention (for intra-patch reasoning, focusing on local features within each patch) with cross-patch attention (for cross-patch interactions, promoting geometric consistency and continuity across patch boundaries).
Finally, lightweight prediction heads, inspired by DPT-style architectures~\cite{ranftl2021vision}, are applied to the transformer's output, generating offsets relative to the coarse estimates. For the depth estimation, the head produces an offset that is added to the coarse depth map to yield the refined depth map $\DMaps^{refined}_k = \DMaps^{coarse}_k + \Offset{Depth}_k$. Likewise, for the surface normal prediction, the head generates an offset to produce the refined surface normals $\SNormal^{refined}_k = \SNormal^{coarse}_k + \Offset{Normal}_k$, leveraging the fused tokens to deliver high-fidelity estimates that correct and enhance the initial coarse predictions. By combining these outputs, we obtain the high-resolution prediction of the geometry assets.

\subsection{Globally Consistent Multi-Patch Attention}

\subsubsection{Global Positional Encoding}
\label{sec:global_pe}

To maintain spatial correlation between patches, each patch $J_k$ is indexed by a {global spatial coordinate} $(x_{\patchidx}, y_{\patchidx})$, defined as the top-left pixel position of $J_k$ in the source high-resolution image~$I$.  
Within each patch, let $\vect{p}_i = (u_i, v_i)$ denote the local coordinate of the $i$-th token, where $(u_i, v_i)$ specifies its position relative to the top-left corner of the patch.  
To unify all tokens under a common reference frame, we assign each token a corresponding {global spatial coordinate} $\vect{p}_i^{g}$ by offsetting its local position by the patch’s global origin:
\begin{equation}
  \vect{p}_i^{g}
  = (u_i + x_{\patchidx},\, v_i + y_{\patchidx}),
  \label{eq:global_coord}
\end{equation}
which maps every token back to its precise location in the original high-resolution image. This global indexing ensures that tokens describing the same physical region are consistently aligned across different patches and crops.

\subsubsection{Intra- and Cross-Patch Attention}

Let $\Token{i}^c$ denote the $i$-th token extracted from patch $J_c$, and let $\vect{q}_i^c,\vect{k}_i^c,\vect{v}_i^c$ be the corresponding query, key, and value vectors obtained by linear projection.
We apply RoPE~\cite{rope} using the global coordinate $\vect{p}_i^{g}$ of the token:
\begin{equation}
  \tilde{\vect{q}}_i^c = \RoPE(\vect{q}_i^c,\vect{p}_i^{g}),\quad
  \tilde{\vect{k}}_i^c = \RoPE(\vect{k}_i^c,\vect{p}_i^{g}),
  \label{eq:rope_global_tokens}
\end{equation}
while the value vectors remain unchanged, $\tilde{\vect{v}}_i^k = \vect{v}_i^k$.

\vspace{-8pt}
\paragraph{Intra-patch attention.}
To capture fine-grained local structure, we first restrict self-attention to tokens within the same patch.
Let $\tilde{Q}_c,\tilde{K}_c,\tilde{V}_c$ denote the matrices obtained by stacking $\tilde{\vect{q}}_i^c,\tilde{\vect{k}}_i^c,\tilde{\vect{v}}_i^c$ over all tokens in patch $c$.
An intra-patch attention layer computes:
\begin{equation}
  \mathrm{Attn}_{\text{intra}}(\tilde{Q}_c,\tilde{K}_c,\tilde{V}_c)
  = \mathrm{softmax}\!\left(\frac{\tilde{Q}_c \tilde{K}_c^{\top}}{\sqrt{d}}\right)\tilde{V}_c,
  \label{eq:attn_intra}
\end{equation}
where $d$ is the feature dimension.
This operation focuses on local context within $J_c$, refining details and boundaries using information confined to each patch.
\vspace{-8pt}
\paragraph{Cross-patch attention.}
To propagate information globally, we perform cross-patch attention over the concatenated token sequence from all patches.
Let $\tilde{Q},\tilde{K},\tilde{V}$ denote the matrices collecting the RoPE-encoded $\tilde{Q}^c,\tilde{K}^c,\tilde{V}^c$ from every patch.
A cross-patch attention layer is then given by:
\begin{equation}
  \mathrm{Attn}_{\text{cross}}(\tilde{Q},\tilde{K},\tilde{V})
  = \mathrm{softmax}\!\left(\frac{\tilde{Q} \tilde{K}^{\top}}{\sqrt{d}}\right)\tilde{V},
  \label{eq:attn_cross}
\end{equation}
so that each token can attend not only to tokens within its own patch but also to tokens in all other patches.

By alternating intra-patch and cross-patch attention blocks via \Equation\ref{eq:rope_global_tokens}, the network first consolidates local detail within each patch and then exchanges information across patches at the full-image scale.

\begin{figure*}[t]
  \centering
  \setlength{\tabcolsep}{1pt}
  \renewcommand{\arraystretch}{1.0}
  \begin{tabular}{cccc}
    % -------- Row 1 --------
    \begin{tikzpicture}
      \node[anchor=south west, inner sep=0] (img) {\includegraphics[width=0.24\textwidth]{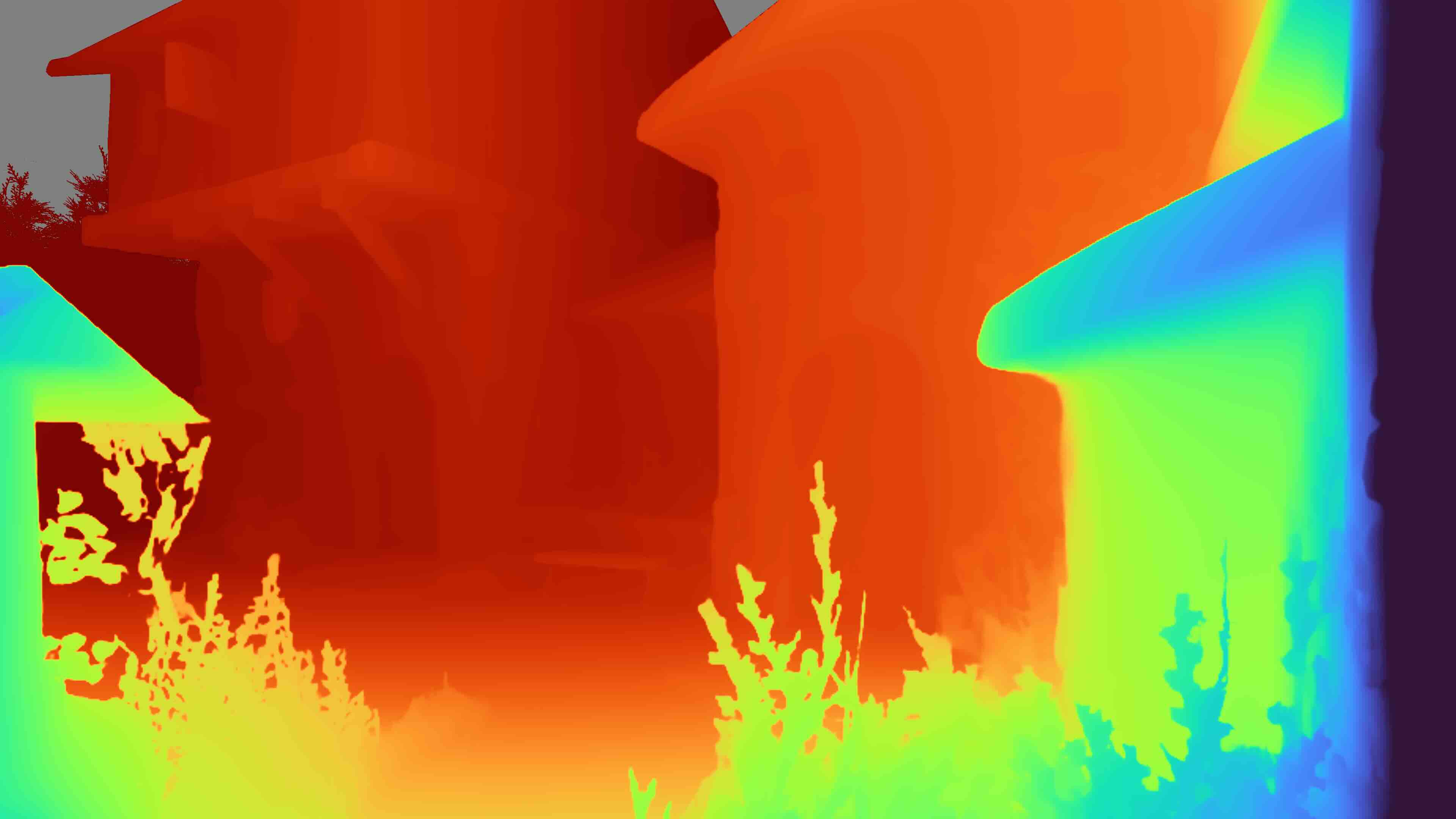}};
      \begin{scope}[x={(img.south east)}, y={(img.north west)}]
        \draw[red, line width=1pt] (0,0) rectangle (0.60,0.60);
      \end{scope}
    \end{tikzpicture} &
    \begin{tikzpicture}
      \node[anchor=south west, inner sep=0] (img) {\includegraphics[width=0.24\textwidth]{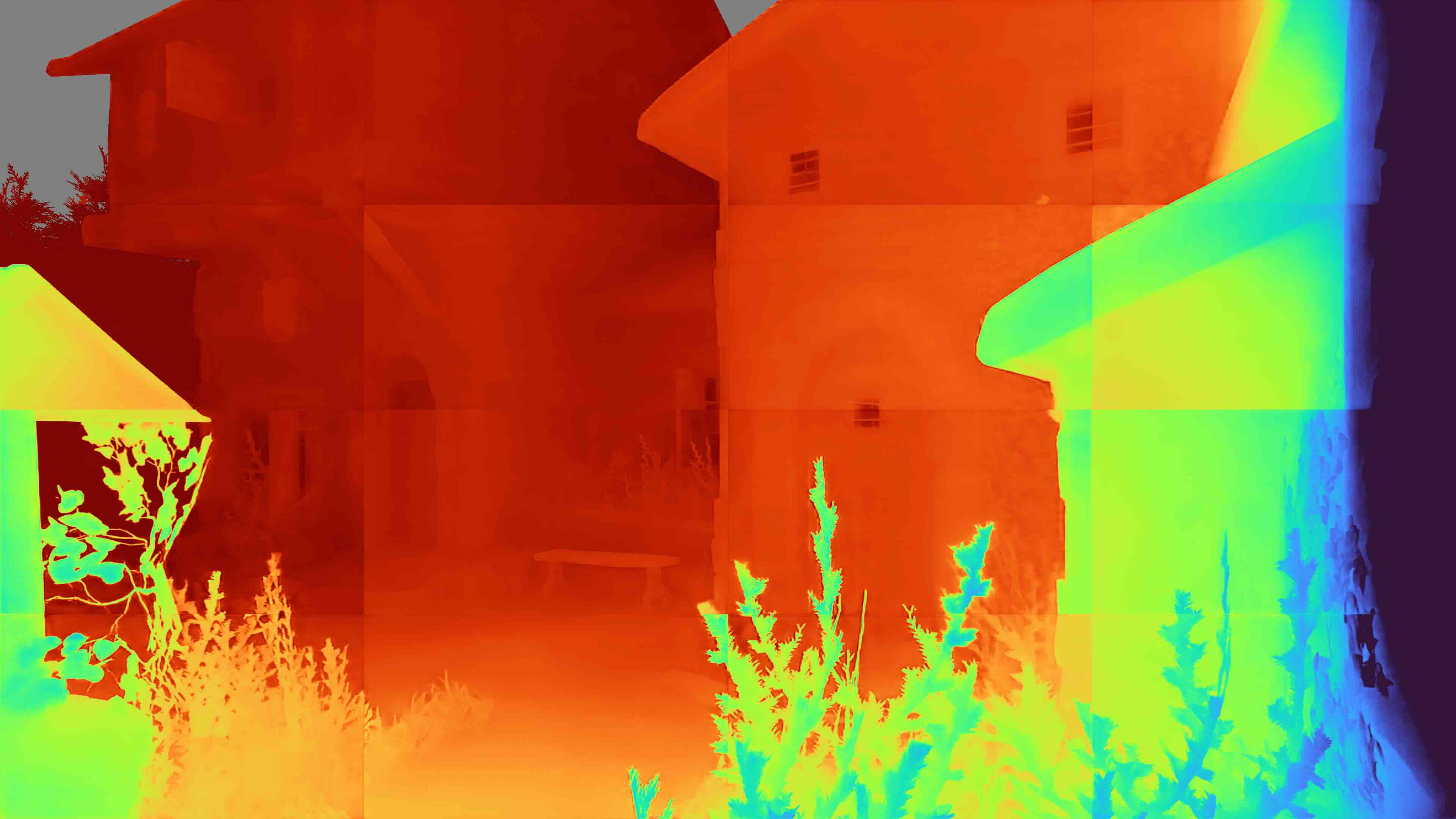}};
      \begin{scope}[x={(img.south east)}, y={(img.north west)}]
        \draw[red, line width=1pt] (0,0) rectangle (0.60,0.60);
      \end{scope}
    \end{tikzpicture} &
    \begin{tikzpicture}
      \node[anchor=south west, inner sep=0] (img) {\includegraphics[width=0.24\textwidth]{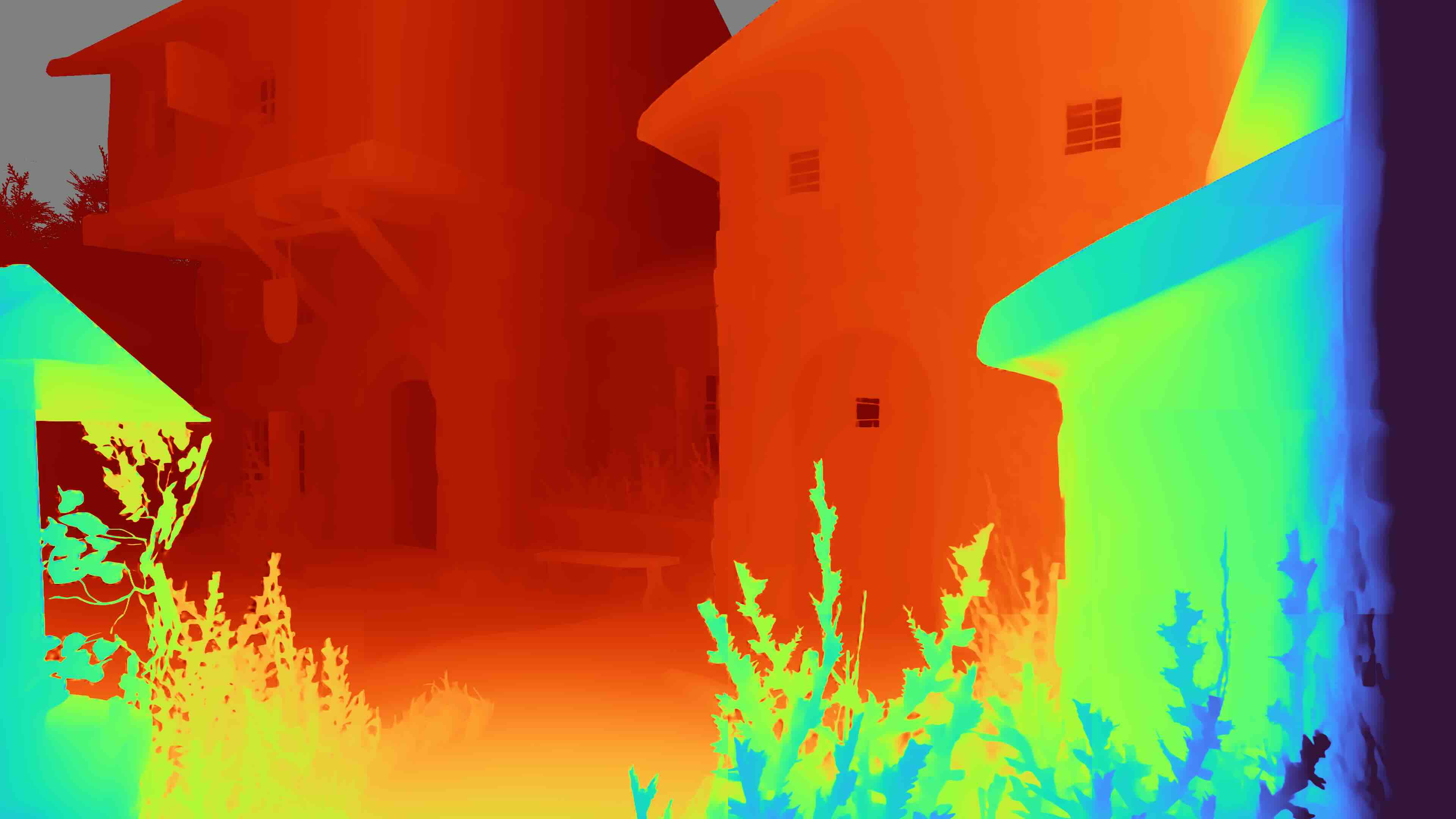}};
      \begin{scope}[x={(img.south east)}, y={(img.north west)}]
        \draw[red, line width=1pt] (0,0) rectangle (0.60,0.60);
      \end{scope}
    \end{tikzpicture} &
    \begin{tikzpicture}
      \node[anchor=south west, inner sep=0] (img) {\includegraphics[width=0.24\textwidth]{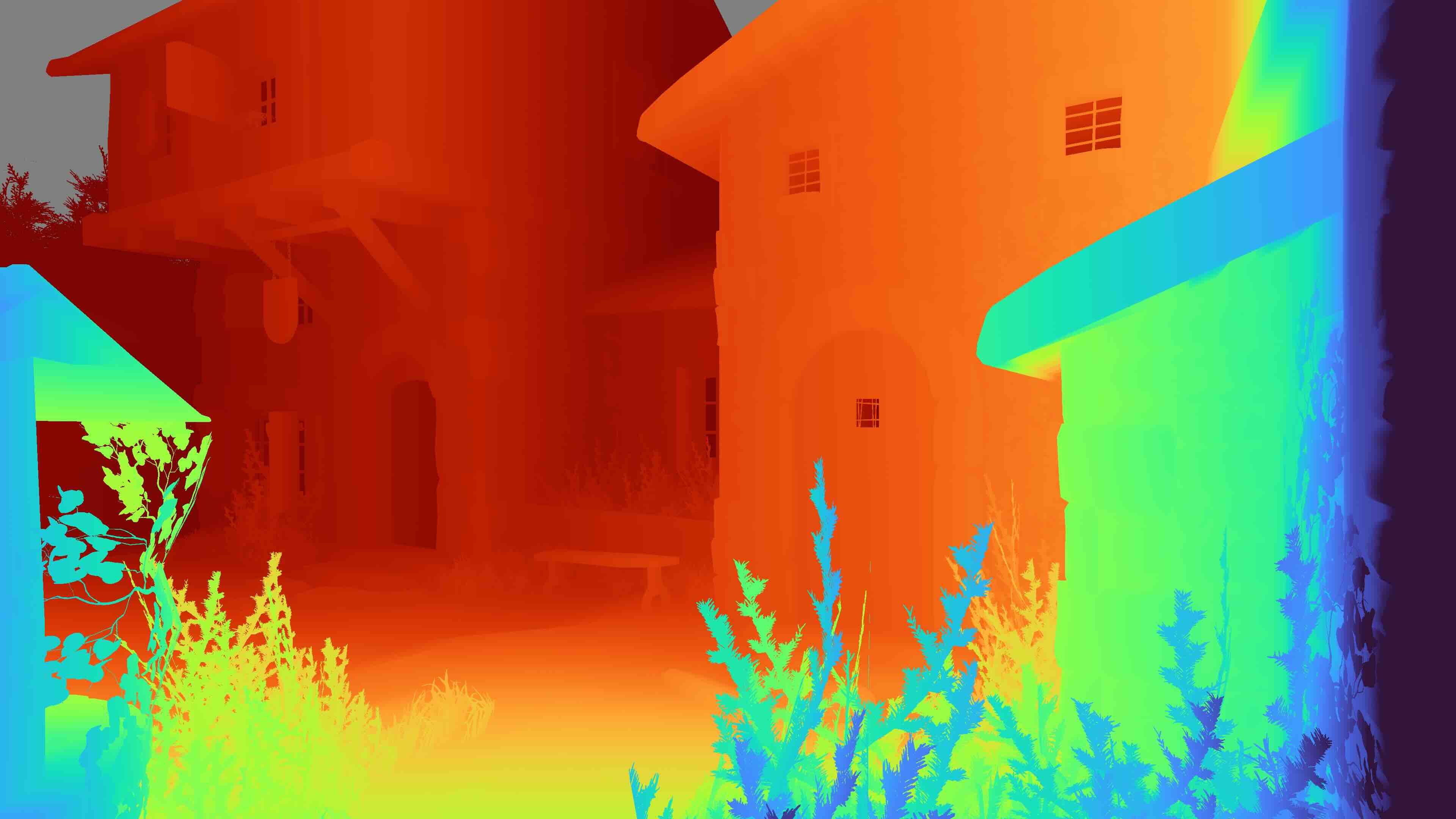}};
      \begin{scope}[x={(img.south east)}, y={(img.north west)}]
        \draw[red, line width=1pt] (0,0) rectangle (0.60,0.60);
      \end{scope}
    \end{tikzpicture} \\[-2pt]
    % -------- Row 2 --------
    \includegraphics[width=0.24\textwidth]{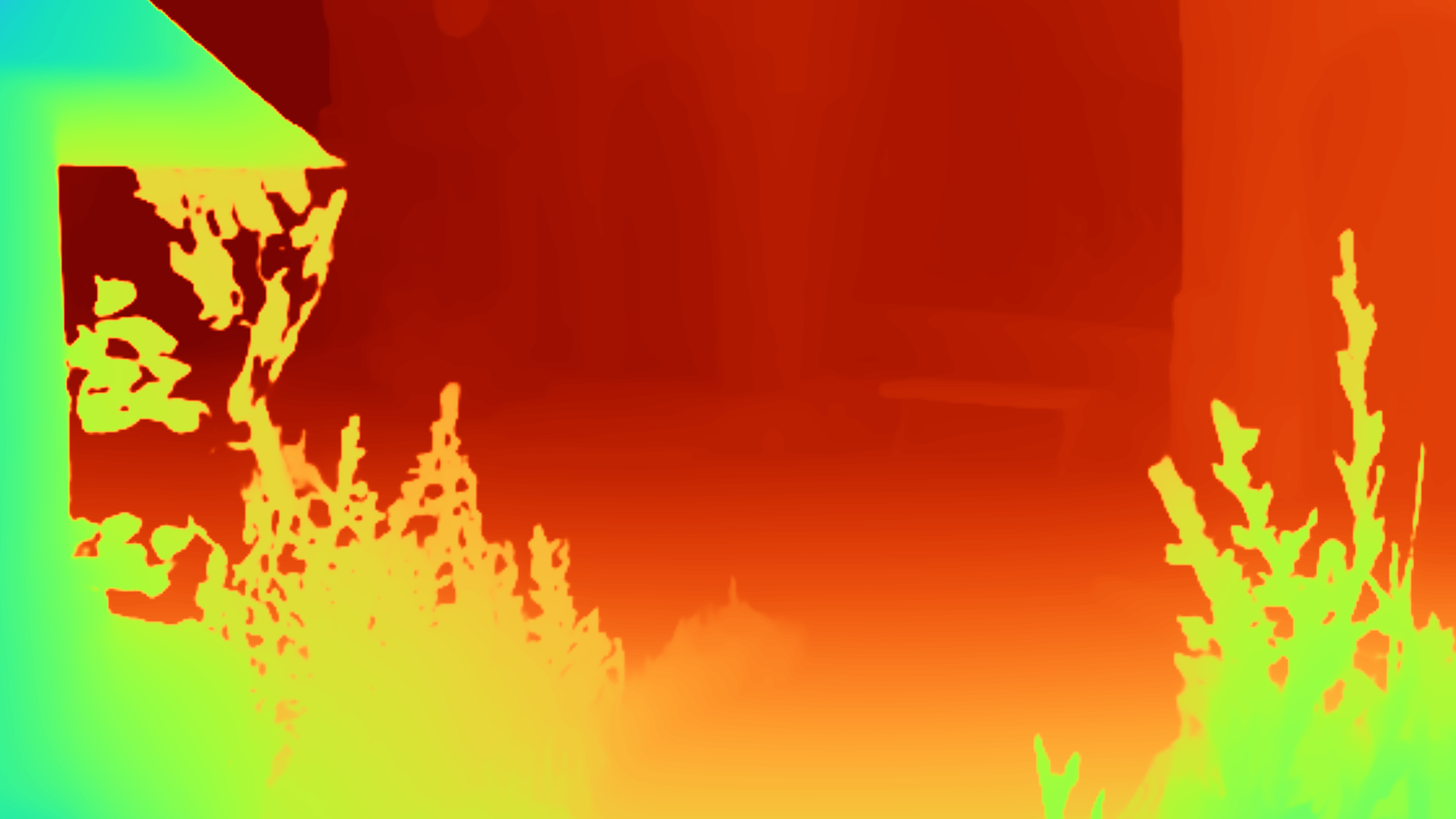} &
    \includegraphics[width=0.24\textwidth]{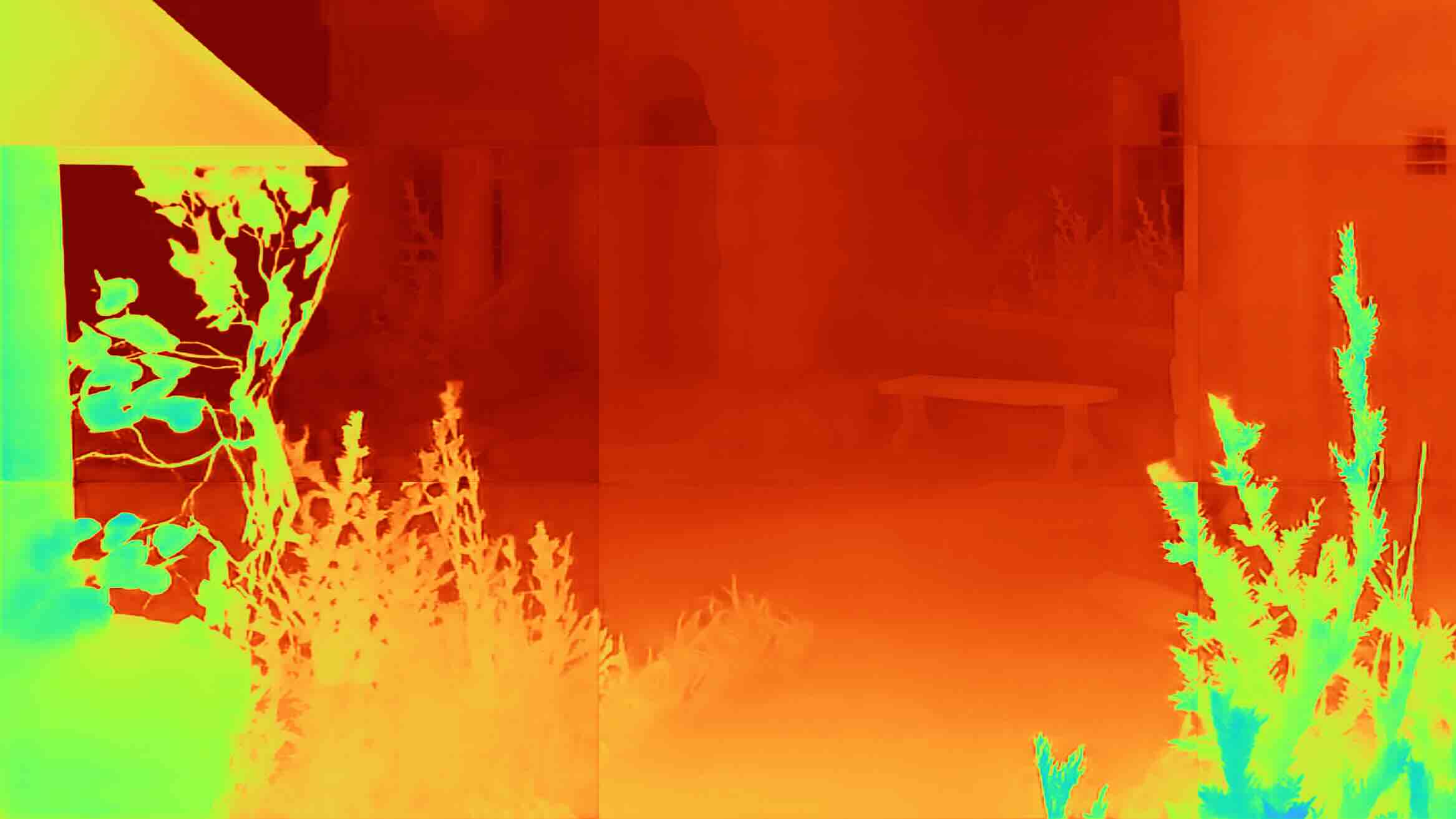} &
    \includegraphics[width=0.24\textwidth]{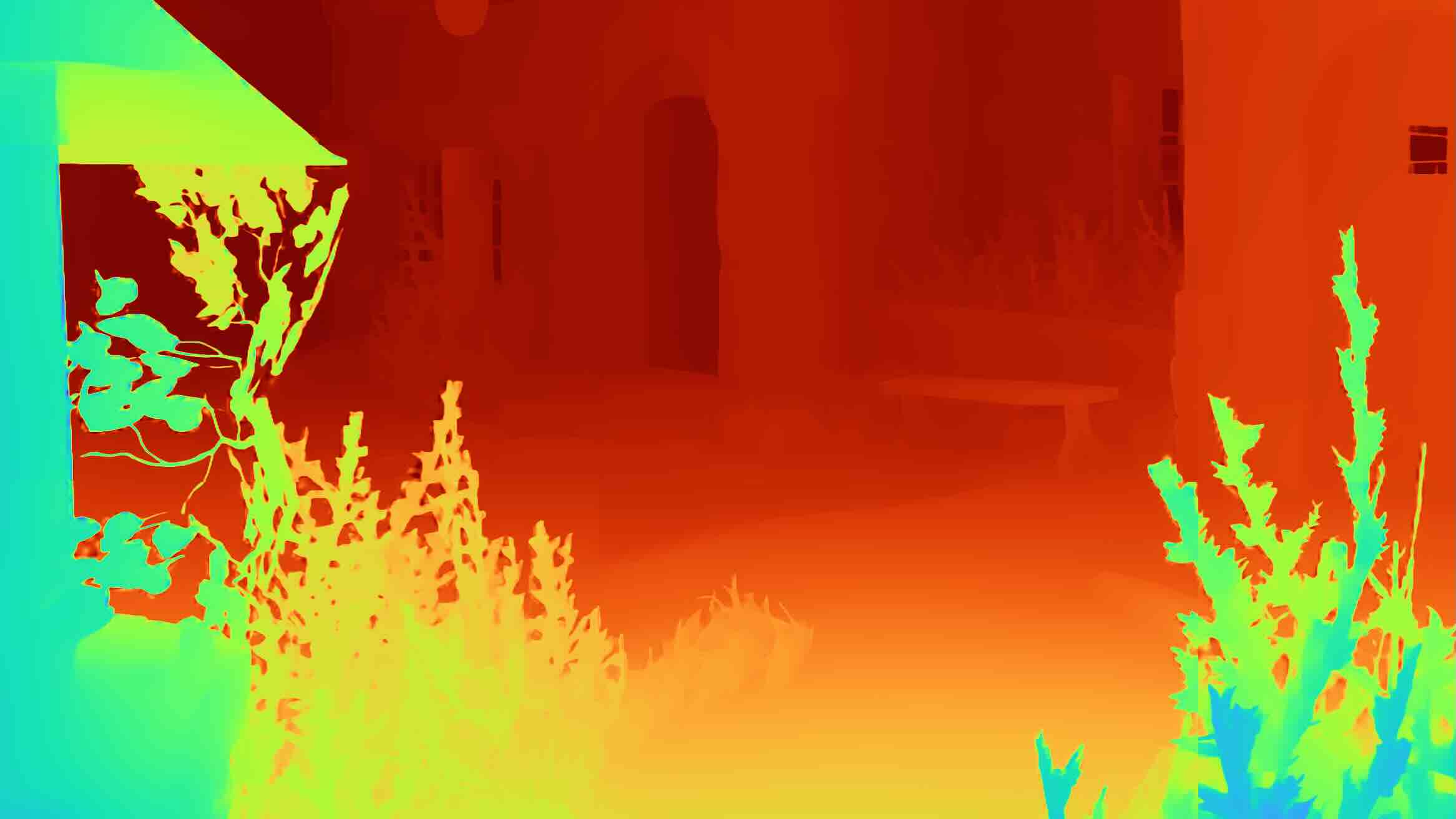} &
    \begin{tikzpicture}
      \node[anchor=south west, inner sep=0] (img) {\includegraphics[width=0.24\textwidth]{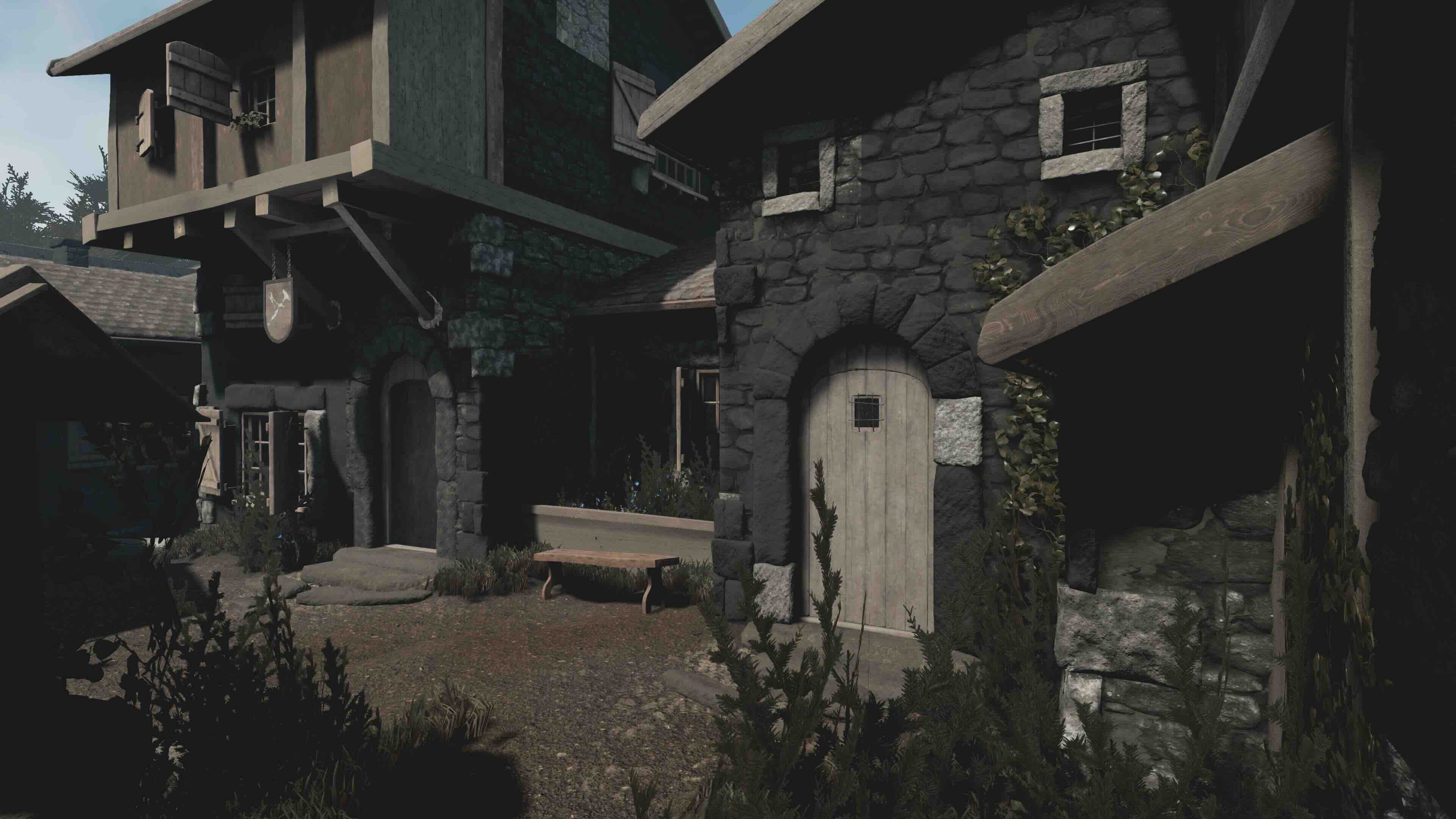}};
      \begin{scope}[x={(img.south east)}, y={(img.north west)}]
        \draw[red, line width=1pt] (0,0) rectangle (0.60,0.60);
      \end{scope}
    \end{tikzpicture} \\ 
    % -------- Row 3 --------
    \begin{tikzpicture}
      \node[anchor=south west, inner sep=0] (img) {\includegraphics[width=0.24\textwidth]{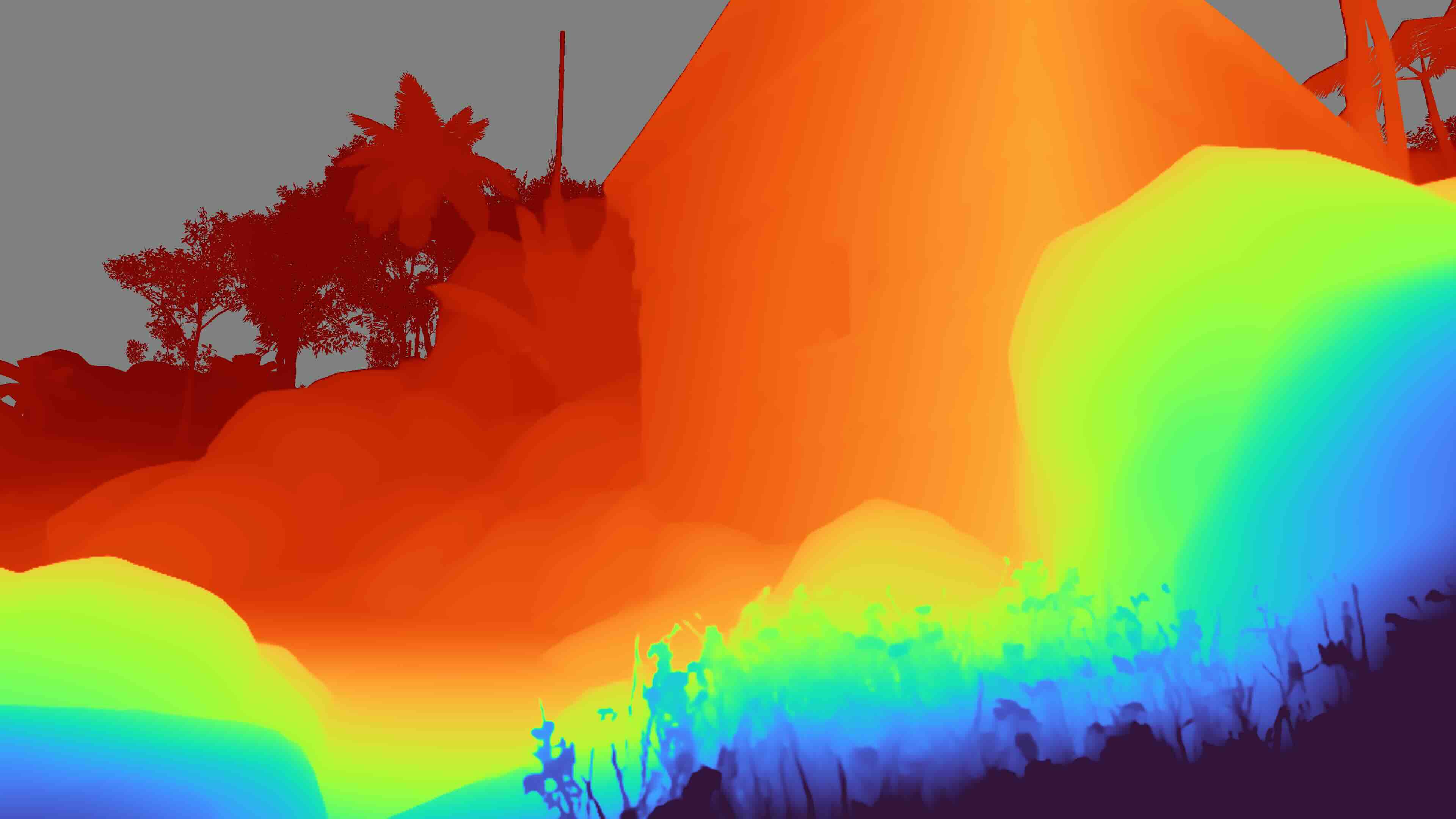}};
      \begin{scope}[x={(img.south east)}, y={(img.north west)}]
        \draw[yellow, line width=1pt] (0.36,0) rectangle (1,0.64);
      \end{scope}
    \end{tikzpicture} &
    \begin{tikzpicture}
      \node[anchor=south west, inner sep=0] (img) {\includegraphics[width=0.24\textwidth]{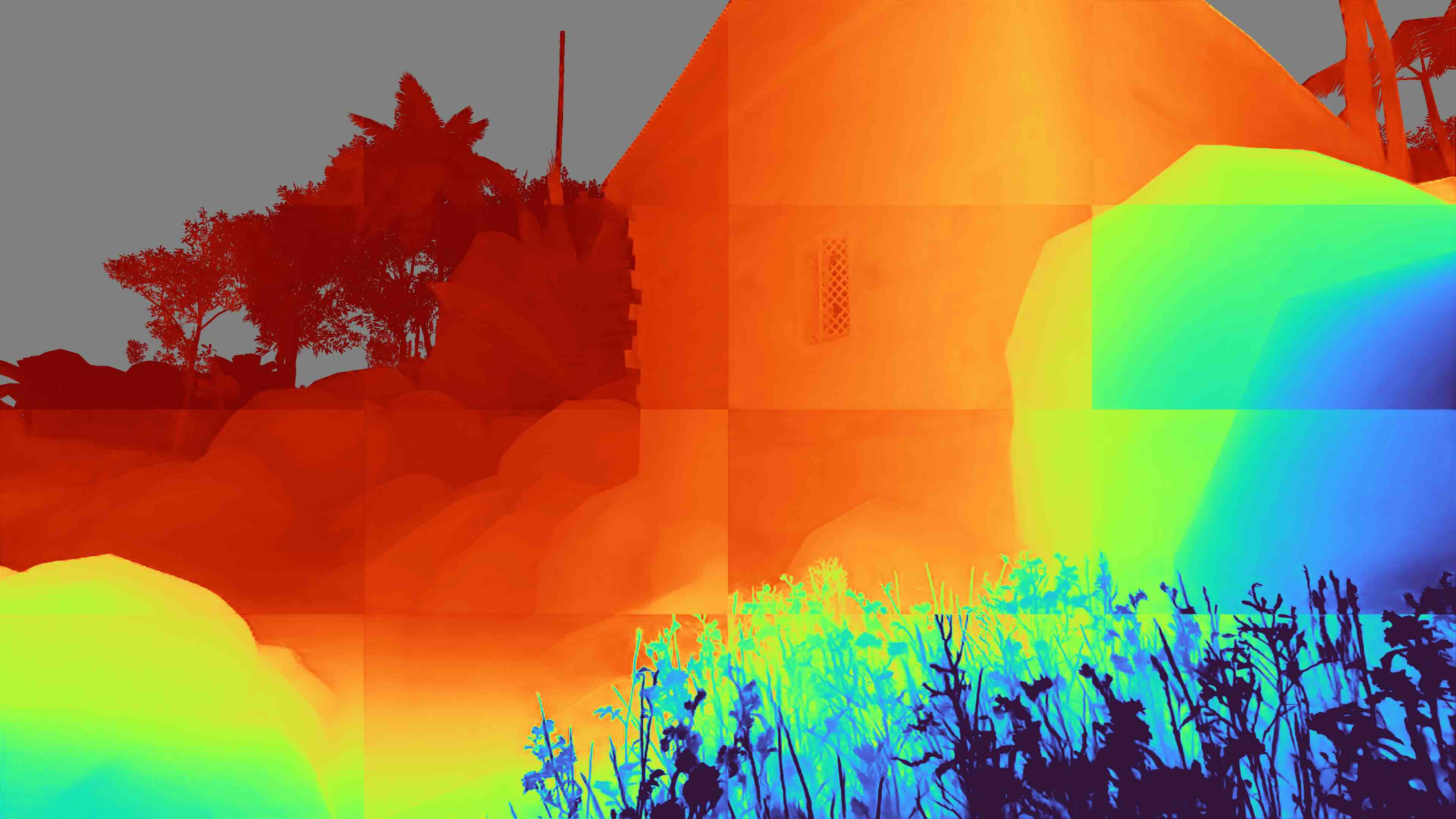}};
      \begin{scope}[x={(img.south east)}, y={(img.north west)}]
        \draw[yellow, line width=1pt] (0.36,0) rectangle (1,0.64);
      \end{scope}
    \end{tikzpicture} &
    \begin{tikzpicture}
      \node[anchor=south west, inner sep=0] (img) {\includegraphics[width=0.24\textwidth]{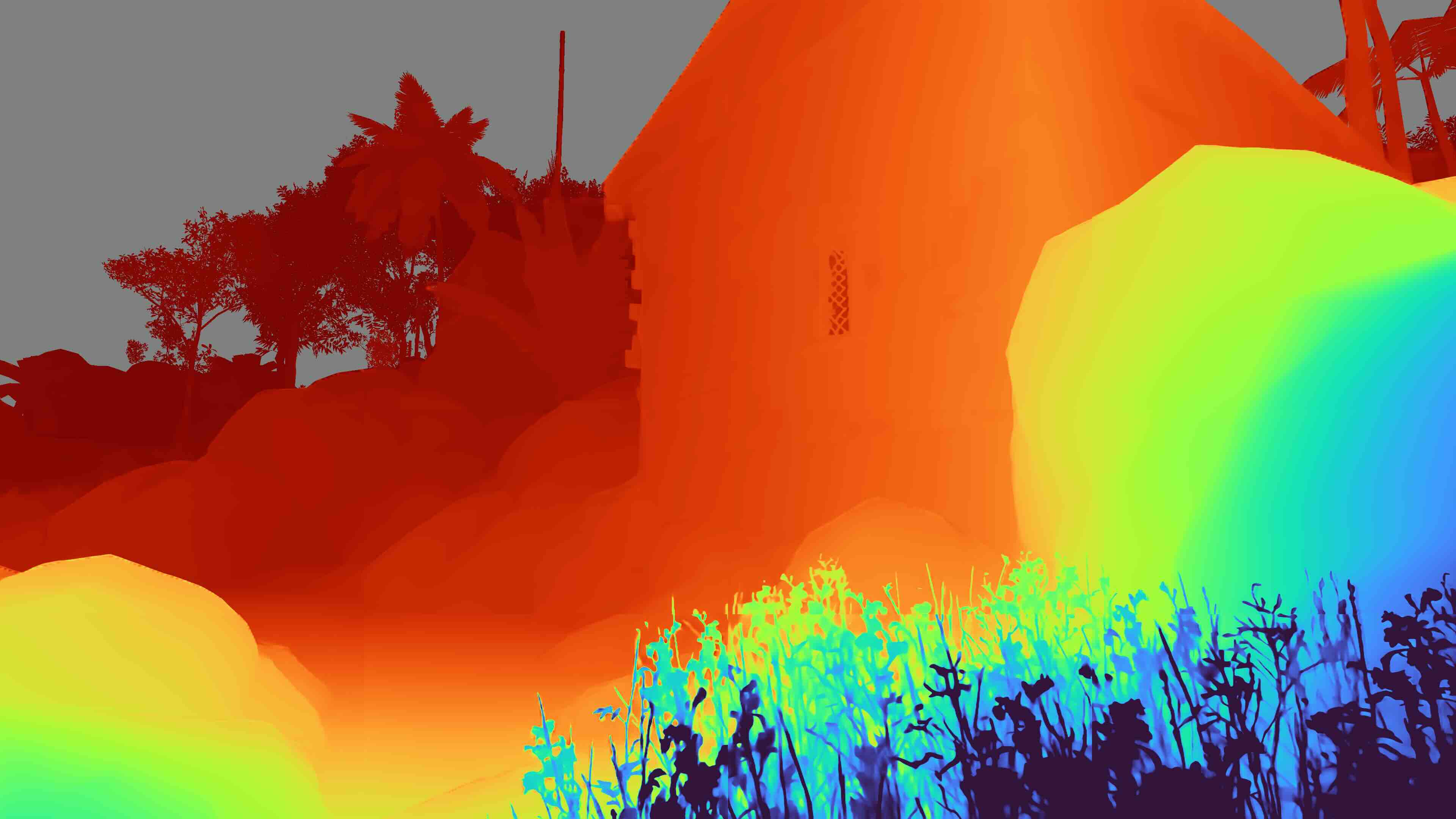}};
      \begin{scope}[x={(img.south east)}, y={(img.north west)}]
        \draw[yellow, line width=1pt] (0.36,0) rectangle (1,0.64);
      \end{scope}
    \end{tikzpicture} &
    \begin{tikzpicture}
      \node[anchor=south west, inner sep=0] (img) {\includegraphics[width=0.24\textwidth]{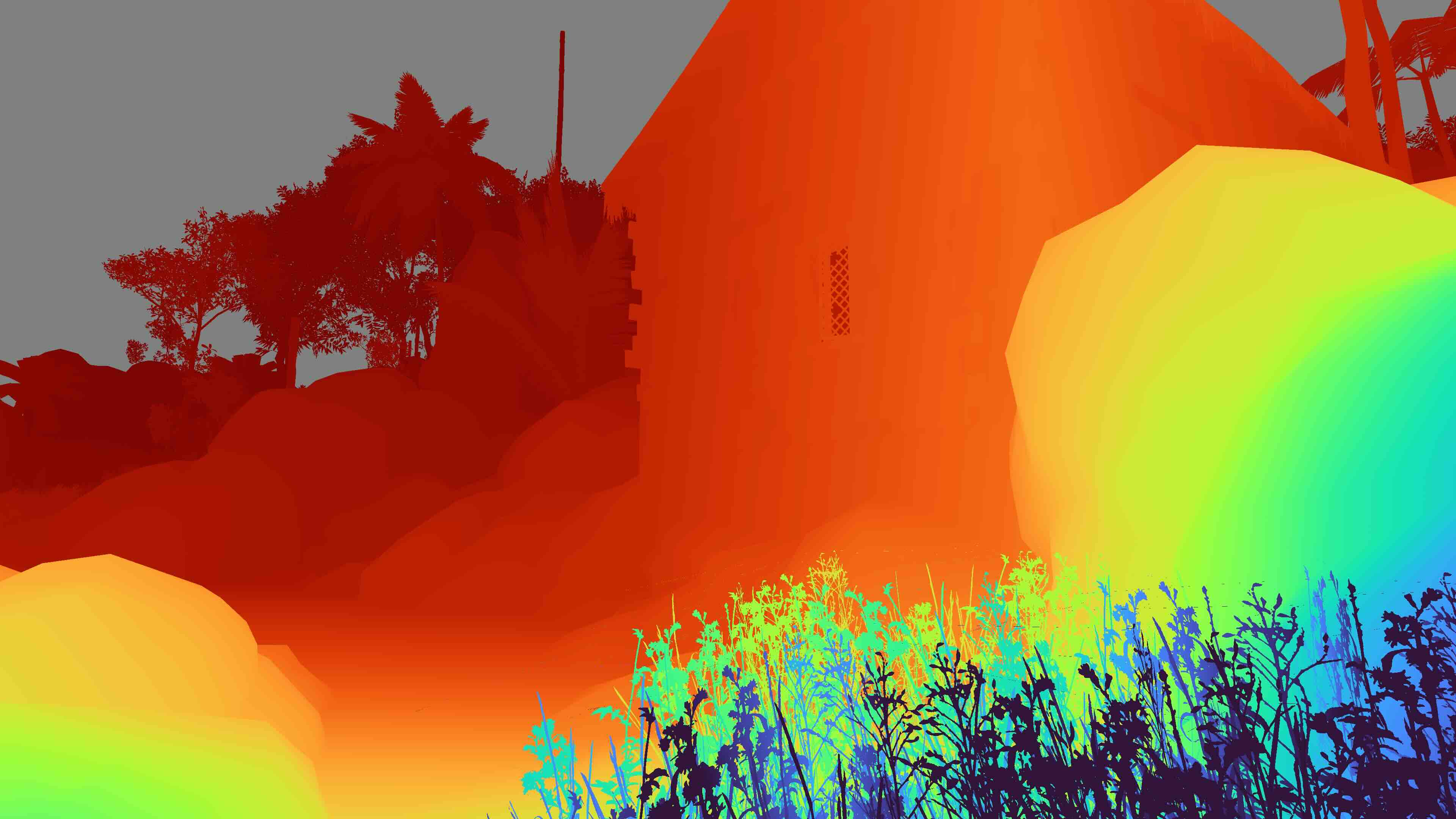}};
      \begin{scope}[x={(img.south east)}, y={(img.north west)}]
        \draw[yellow, line width=1pt] (0.36,0) rectangle (1,0.64);
      \end{scope}
    \end{tikzpicture} \\[-2pt]
    % -------- Row 4 --------
    \includegraphics[width=0.24\textwidth]{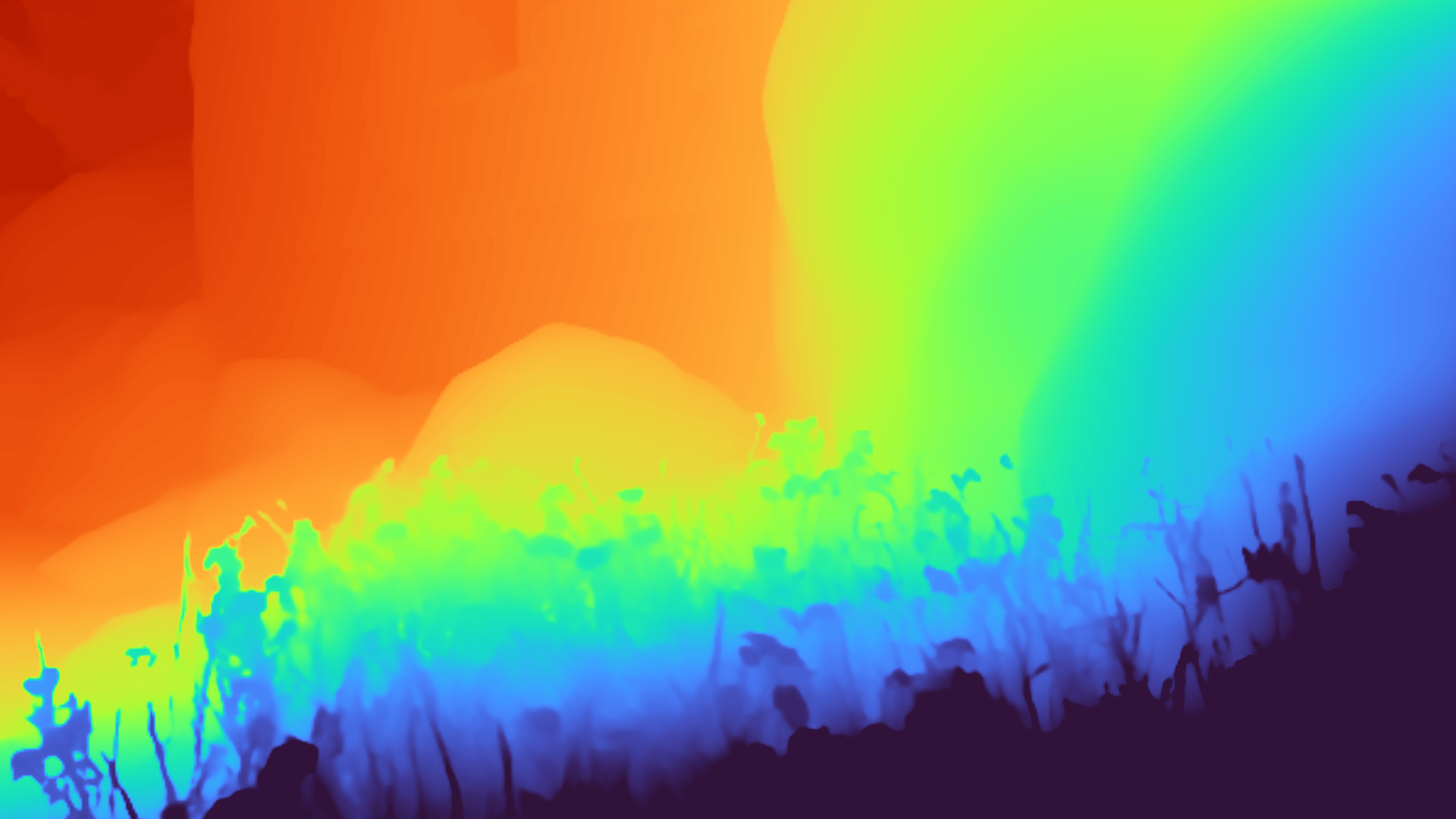} &
    \includegraphics[width=0.24\textwidth]{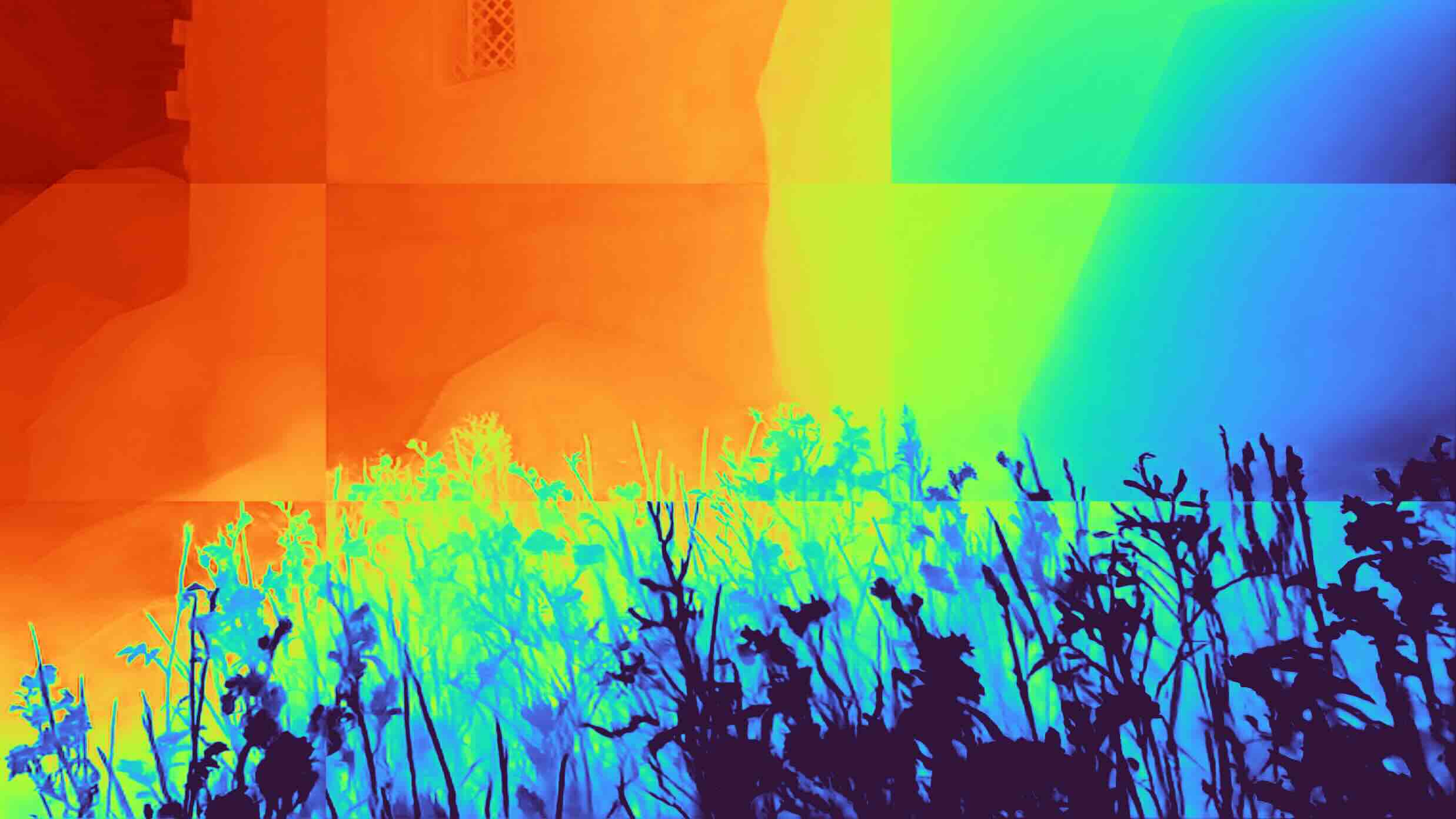} &
    \includegraphics[width=0.24\textwidth]{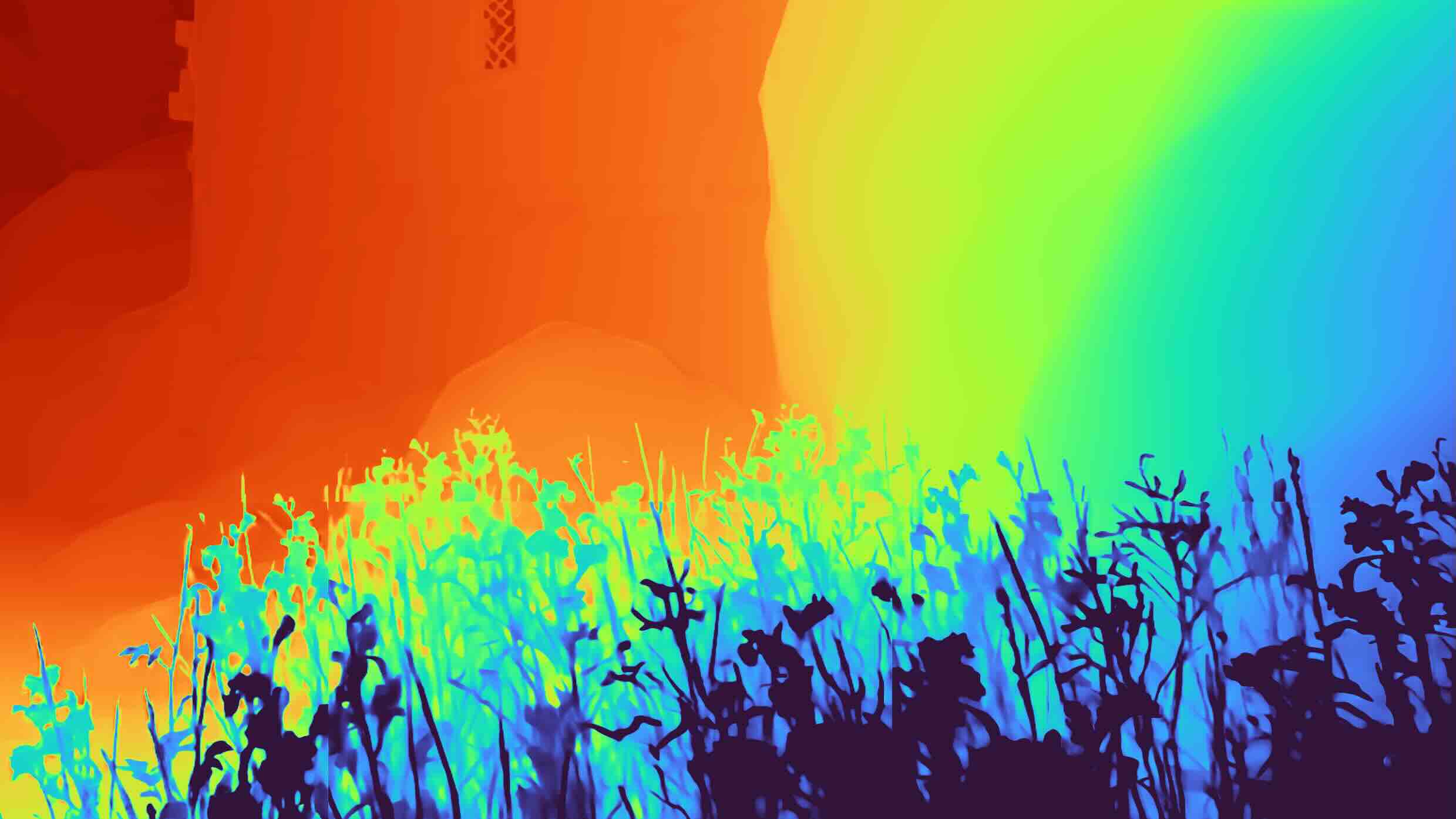} &
    \begin{tikzpicture}
      \node[anchor=south west, inner sep=0] (img) {\includegraphics[width=0.24\textwidth]{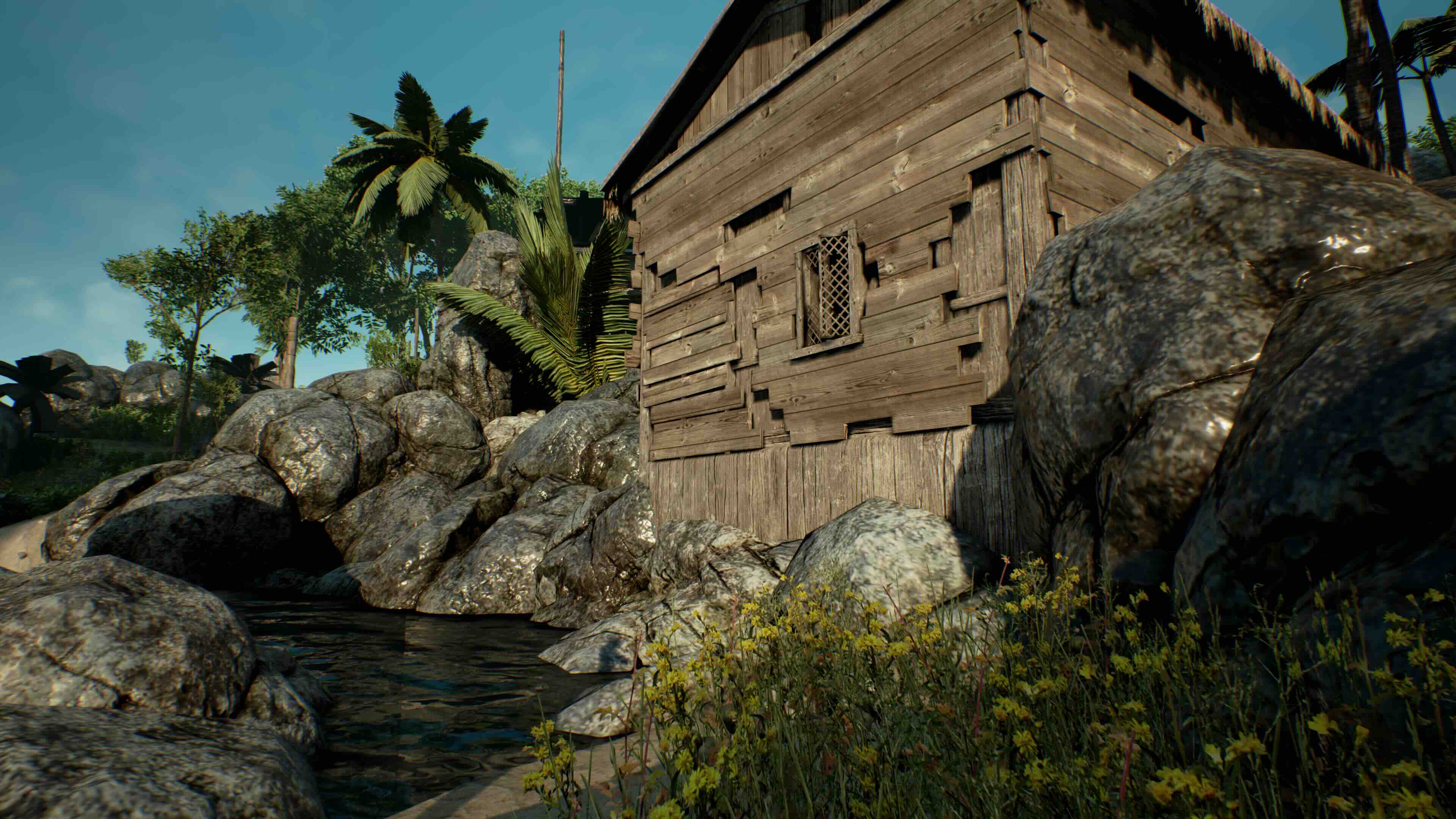}};
      \begin{scope}[x={(img.south east)}, y={(img.north west)}]
        \draw[yellow, line width=1pt] (0.36,0) rectangle (1,0.64);
      \end{scope}
    \end{tikzpicture} \\[3pt]
    % -------- Column labels --------
    \textbf{\small Depth Anything V2}~\cite{yang2025depthanythingv2} &
    \textbf{\small PatchRefiner}~\cite{li2024patchrefiner} &
    \textbf{\small Ours} &
    \textbf{\small GT, Input}
  \end{tabular}
  \hspace{-20pt}
    \caption{
    \textbf{Qualitative comparison on the UnrealStereo4K\cite{u4k} dataset.}
    Red rectangles indicate the bounding boxes of zoomed-in regions.
    The second and fourth rows show close-up (zoom-in) views corresponding to the first and third rows, respectively.
    The rightmost column presents the RGB input and ground-truth depth maps.
    Compared to previous methods, our predictions exhibit smoother depth continuity and sharper geometry boundaries.
    }

  \label{fig:u4k}
  \hspace{-20pt}
  
\end{figure*}

\subsection{GridMix Patch Sampling}
We employ a GridMix patch sampling strategy that incorporates different grid configurations for training our models effectively. This approach utilizes patches of a fixed size but sampled in various spatial configurations. Specifically, for an input RGB image $\Img \in \mathbb{R}^{3 \times \HImg \times \WImg}$, we extract at each iteration $\NPatch \times \NPatch$ patches of fixed size $\Patch_k \in \mathbb{R}^{3 \times \frac{\HImg}{4} \times \frac{\WImg}{4}}, k=1 \dots \NPatch^2$, for $\NPatch \in \{1,2,3,4\}$. Thus, the four configurations that we employ to generate the patches are: 1) if $M=1$, a single patch randomly sampled from the image, 2) if $M\in \{2,3\}$, a $M \times M$ grid of patches will be randomly sampled from the image, and 3) if $M=4$, a fixed $4 \times 4$ grid of patches that covers the entire image.
Figure~\ref{fig:patches} illustrates the GridMix patch sampling strategy by visualizing the four distinct configurations applied to an input image, including the valid regions (green areas) of the top-left point for random sampling to ensure full grid containment.
In our training procedure, at each iteration, one of the four configurations is selected according to a predefined probability distribution:
\begin{equation}
\label{eq:config_probability}
\Prob{M}_{M=1..4} = \vProb{M}, \quad \sum_{M=1}^{4} \vProb{M} = 1 
\end{equation}

where $\big(\vProb{k}\big)_{k=1\dots4}$ are hyperparameters to control the probability of choosing each of the configurations. This probabilistic selection promotes diversity in how the image is partitioned, enabling the model to encounter various sampling strategies while keeping all patches at a fixed size.
As demonstrated in the experiment section, the GridMix patch sampling strategy functions as an effective data augmentation technique by offering multiple ways to divide and sample the image.
Our experiments show that this strategy leads to significant improvements in high-resolution depth estimation compared to training with only a fixed grid.

\begin{table*}[ht]
\caption{
\textbf{Quantitative comparison on the UnrealStereo4K~\cite{u4k} dataset.}
Top: depth estimation results (AbsRel, $\delta_1$, RMSE, CE, PDBE). 
Bottom: surface normal estimation results (Mean/Median/RMSE angular errors and threshold accuracies).
}
\label{tab:u4k}
\centering

% ---------------- DEPTH TABLE ----------------
\vspace{-10pt}
\vspace{0pt}

\scalebox{0.8}{
\begin{tabular}{@{}L{3.4cm}*{7}{|C{1.9cm}}@{}}
\toprule
\textbf{Method {\footnotesize (Depth)}} &
\textbf{Infer Time} $\downarrow$ &
\textbf{AbsRel} $\downarrow$ &
$\boldsymbol{\delta_1}$ $\uparrow$ &
\textbf{RMSE} $\downarrow$ &
\textbf{CE} $\downarrow$ &
$\boldsymbol{\epsilon_{\mathrm{PDBE}}^{\mathrm{acc}}}$ $\downarrow$ &
$\boldsymbol{\epsilon_{\mathrm{PDBE}}^{\mathrm{compl}}}$ $\downarrow$ \\
\midrule
Depth-Anything v2~\cite{yang2025depthanythingv2} & --   & 0.0812 & 0.924 & 2.86 & --      & 2.46 & 189.0 \\
PatchRefiner$_{p=16}$~\cite{li2024patchrefiner}  & 1.02s & 0.0633 & 0.950 & 2.28 & 0.0753  & 1.98 & \textbf{55.66} \\
PatchRefiner$_{p=49}$~\cite{li2024patchrefiner} & 4.12s & 0.0582 & 0.956 & 2.17 & 0.0715  & 1.50 & 169.81 \\
PRO~\cite{kwon2025onelook}           & 1.88 & 0.0771 & 0.927 & 2.73 & 0.0549 & 2.74 & 417.7 \\
\textbf{Ours\hspace{1em} {\footnotesize \color{gray} Separate}}                                    & \textbf{0.94s} & 0.0295 & 0.982 & 1.38 & 0.0418 & 1.73 & 72.8 \\
\textbf{Ours\hspace{1em} {\footnotesize \color{gray} Joint}}                              & 0.97s & \textbf{0.0291} & \textbf{0.983} & \textbf{1.31} & \textbf{0.0415} & \textbf{1.40} & 64.21 \\
\bottomrule
\end{tabular}
}
\scalebox{0.8}{
\begin{tabular}{@{}L{3.4cm}*{7}{|C{1.9cm}}@{}}
\toprule
\textbf{Method {\footnotesize (Normal)}} &
\textbf{Infer Time} $\downarrow$ &
\textbf{Median} $\downarrow$ &
\textbf{Mean} $\downarrow$ &
\textbf{RMSE} $\downarrow$ &
$5^{\circ}$ $\uparrow$ &
$11.25^{\circ}$ $\uparrow$ &
$30^{\circ}$ $\uparrow$ \\
\midrule
Metric3D v2~\cite{hu2025metric3dv2} & \textbf{0.64s} & 33.15 & 23.36 & 13.90 & 11.74 & 44.96 & 79.77 \\
\textbf{Ours \hspace{1em} {\footnotesize \color{gray} Joint}} & 0.97s & \textbf{28.83} & \textbf{18.51} & \textbf{9.60} & \textbf{29.37} & \textbf{59.43} & \textbf{85.06} \\
\bottomrule
\end{tabular}
}

\end{table*}

\begin{figure*}[t]
  \centering
  \setlength{\tabcolsep}{1pt}
  \renewcommand{\arraystretch}{1.0}
  \begin{tabular}{cccc}
    % -------- Row 1 --------

    \begin{overpic}[width=0.24\linewidth,height=0.135\linewidth]{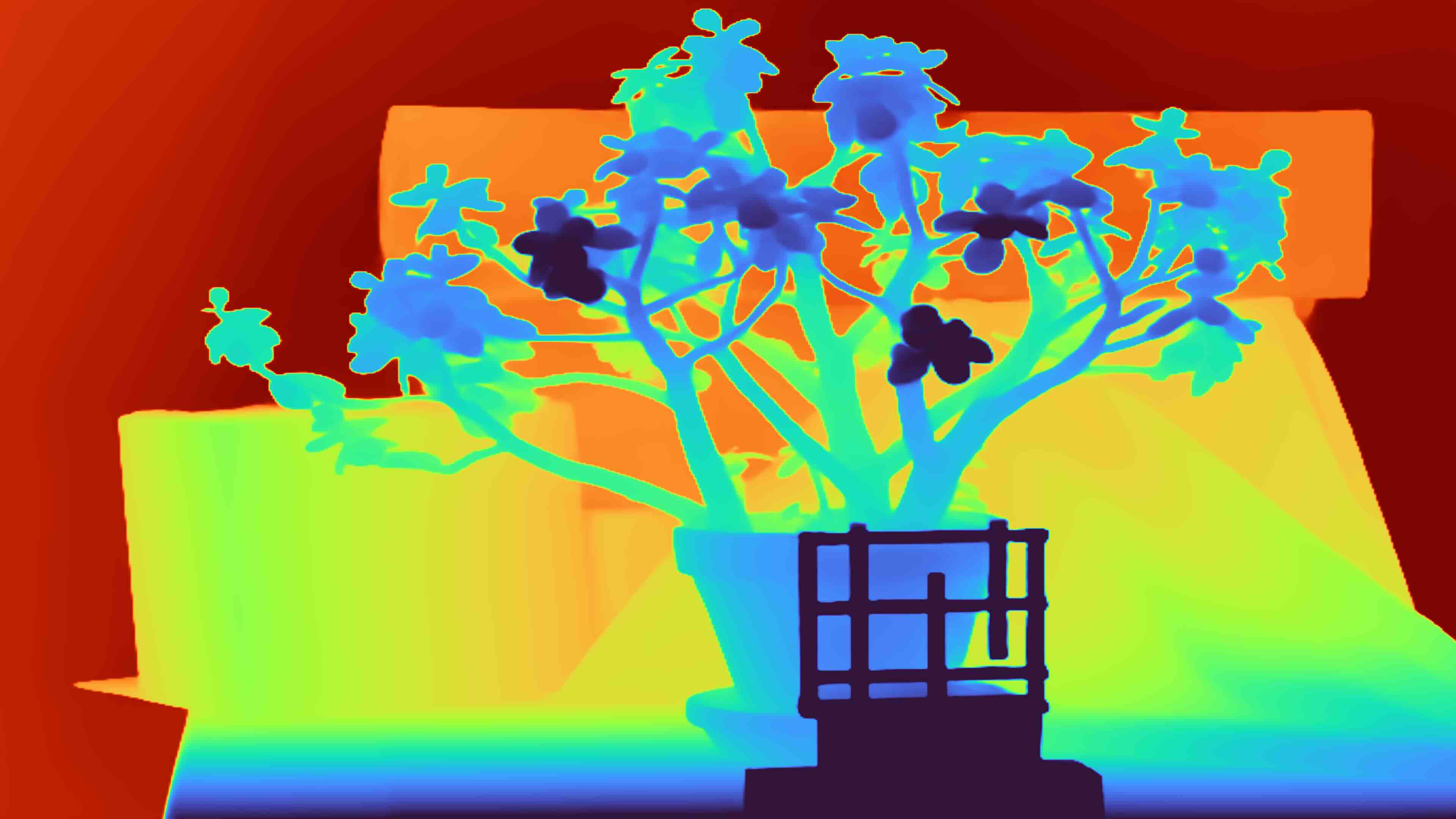}
        \put(57,17.5){\makebox[0pt]{\adjincludegraphics[height=0.1\linewidth,trim={{.3\width} {.2\height} {.3\width} {.3\height}},clip, cfbox=black 2pt 0cm]{sec/fig/middlebury_sample/08_dav2.jpg}}}
    \end{overpic}
    &
    \begin{overpic}[width=0.24\linewidth,height=0.135\linewidth]{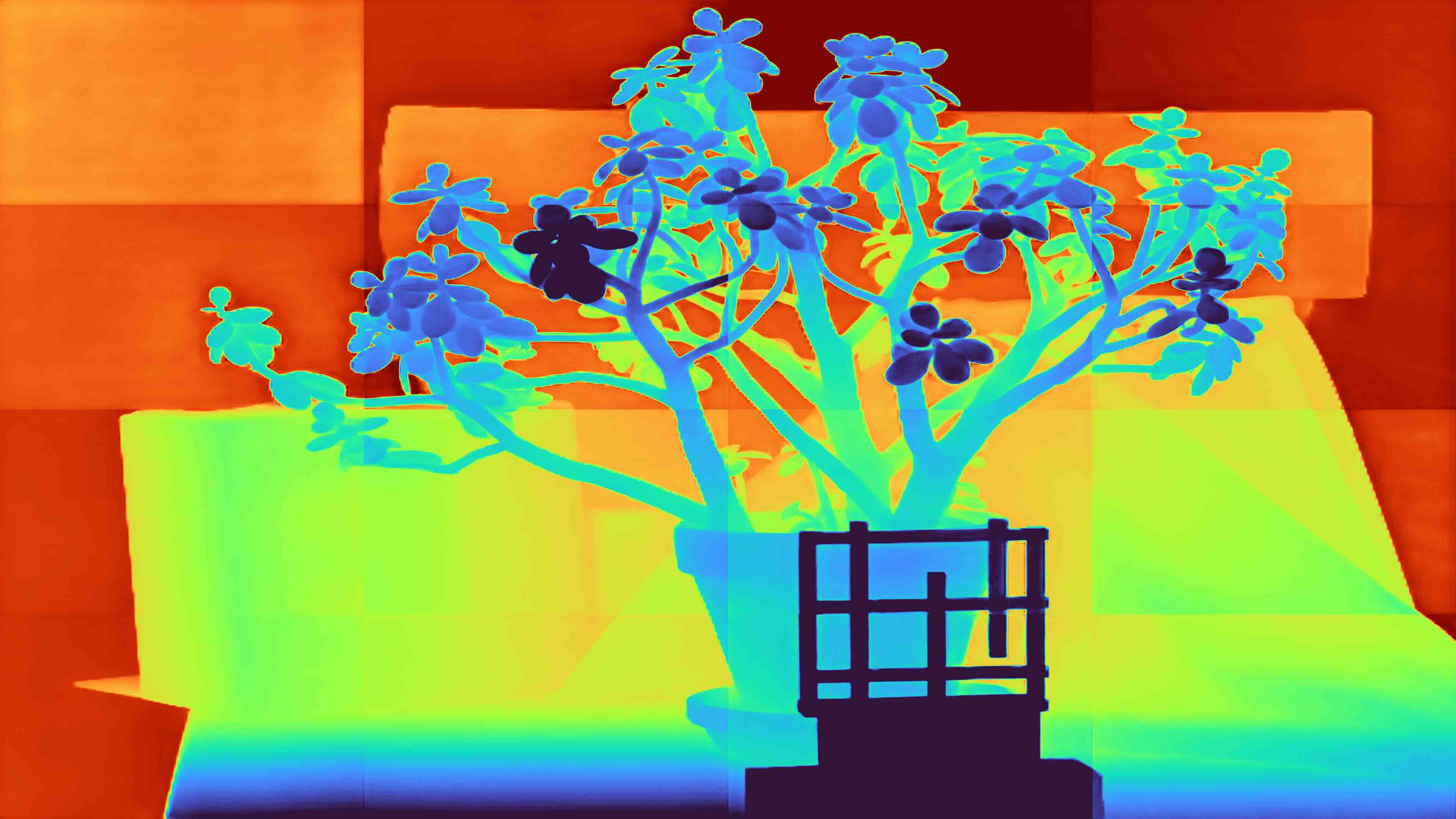}
        \put(57,17.5){\makebox[0pt]{\adjincludegraphics[height=0.1\linewidth,trim={{.3\width} {.2\height} {.3\width} {.3\height}},clip, cfbox=black 2pt 0cm]{sec/fig/middlebury_sample/08_pr.jpg}}}
    \end{overpic}
    &
    \begin{overpic}[width=0.24\linewidth,height=0.135\linewidth]{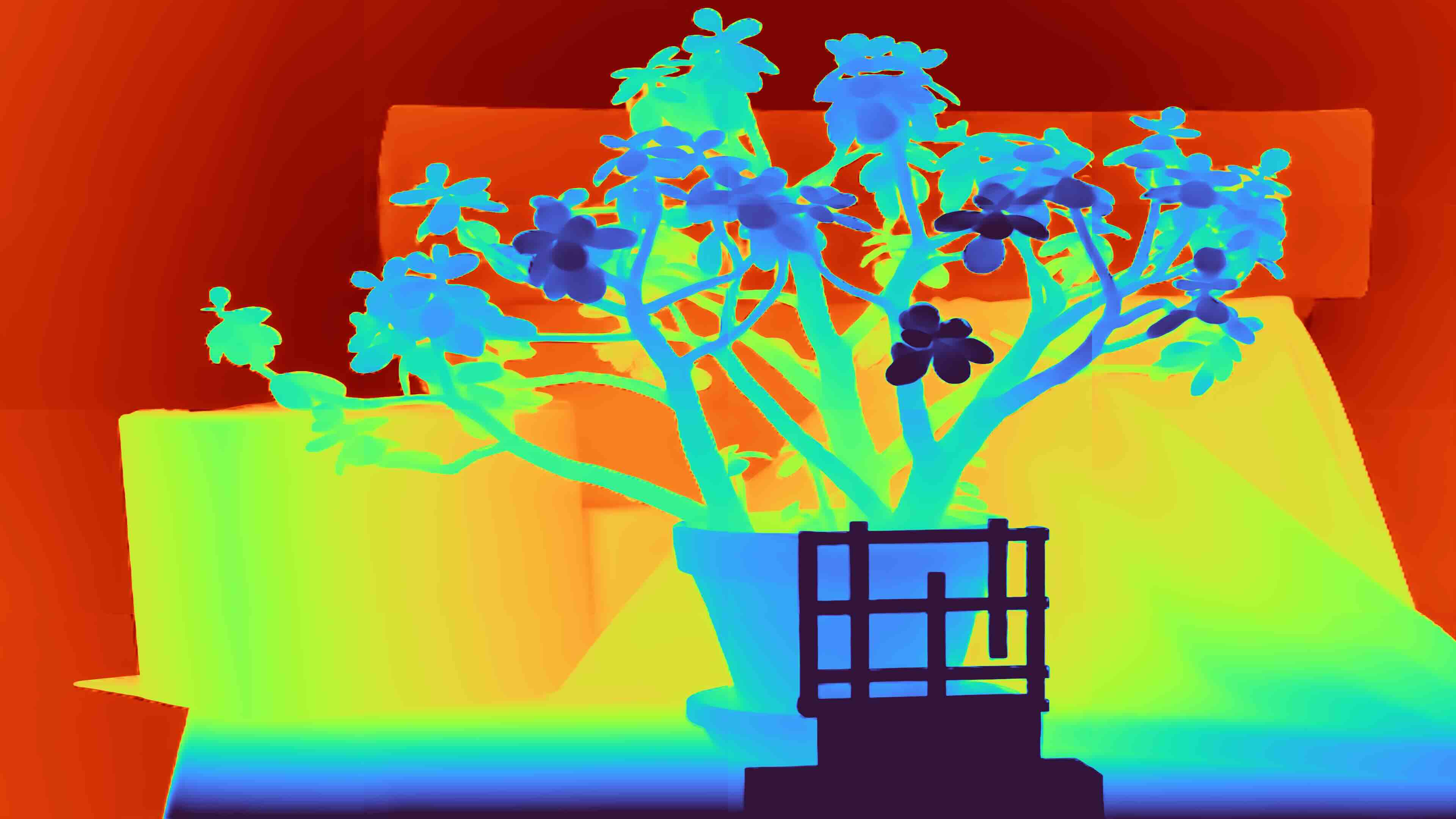}
        \put(57,17.5){\makebox[0pt]{\adjincludegraphics[height=0.1\linewidth,trim={{.3\width} {.2\height} {.3\width} {.3\height}},clip, cfbox=black 2pt 0cm]{sec/fig/middlebury_sample/08_our.jpg}}}
    \end{overpic}
    &
    \includegraphics[width=0.24\textwidth]{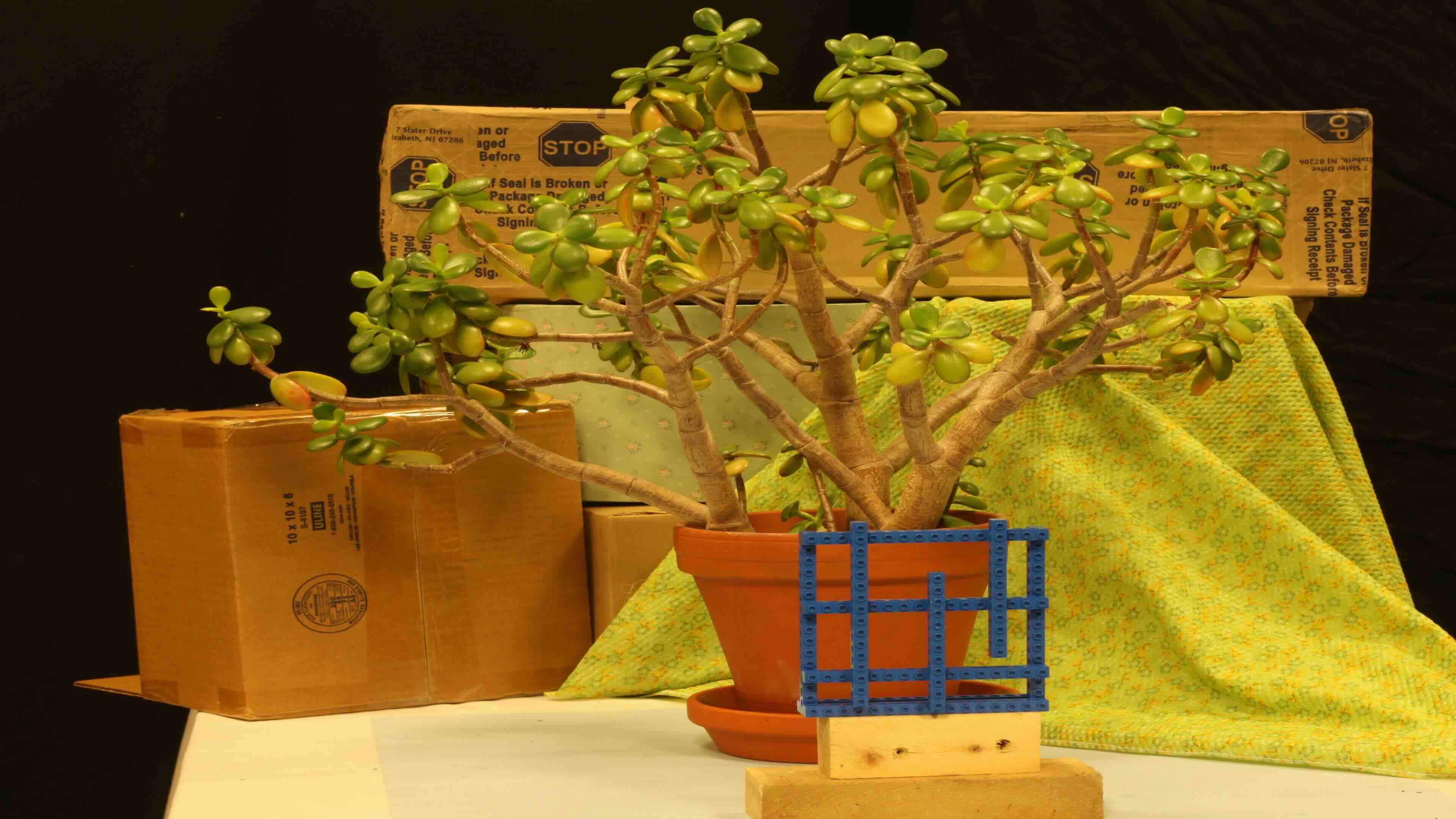} \\[-2pt]
    % -------- Row 2 --------
    \begin{overpic}[width=0.24\linewidth,height=0.135\linewidth]{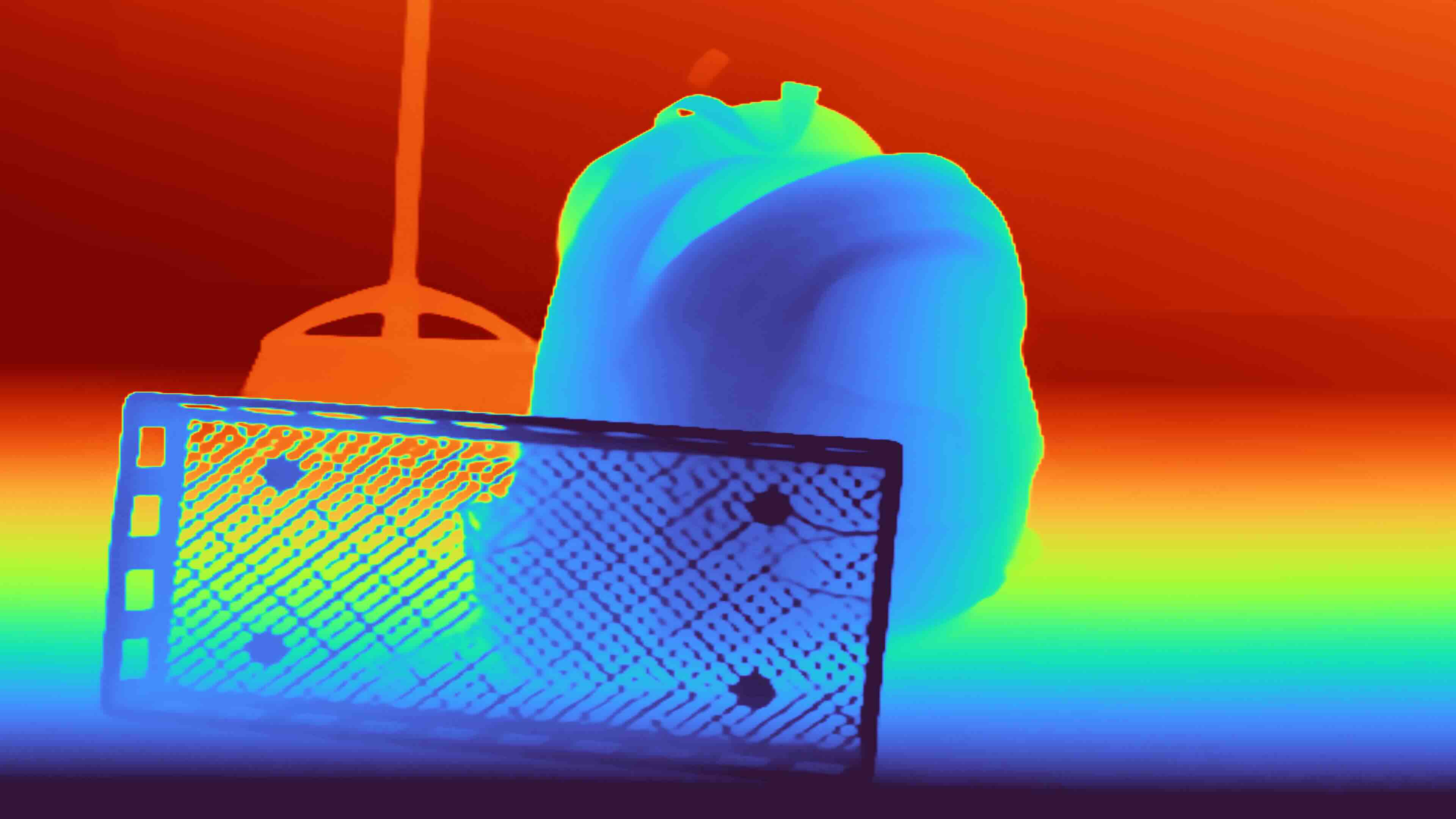}
        \put(64,5){\makebox[0pt]{\adjincludegraphics[height=0.1\linewidth,trim={{.05\width} {.1\height} {.6\width} {.5\height}},clip, cfbox=black 2pt 0cm]{sec/fig/middlebury_sample/18_dav2.jpg}}}
    \end{overpic}
    &
    \begin{overpic}[width=0.24\linewidth,height=0.135\linewidth]{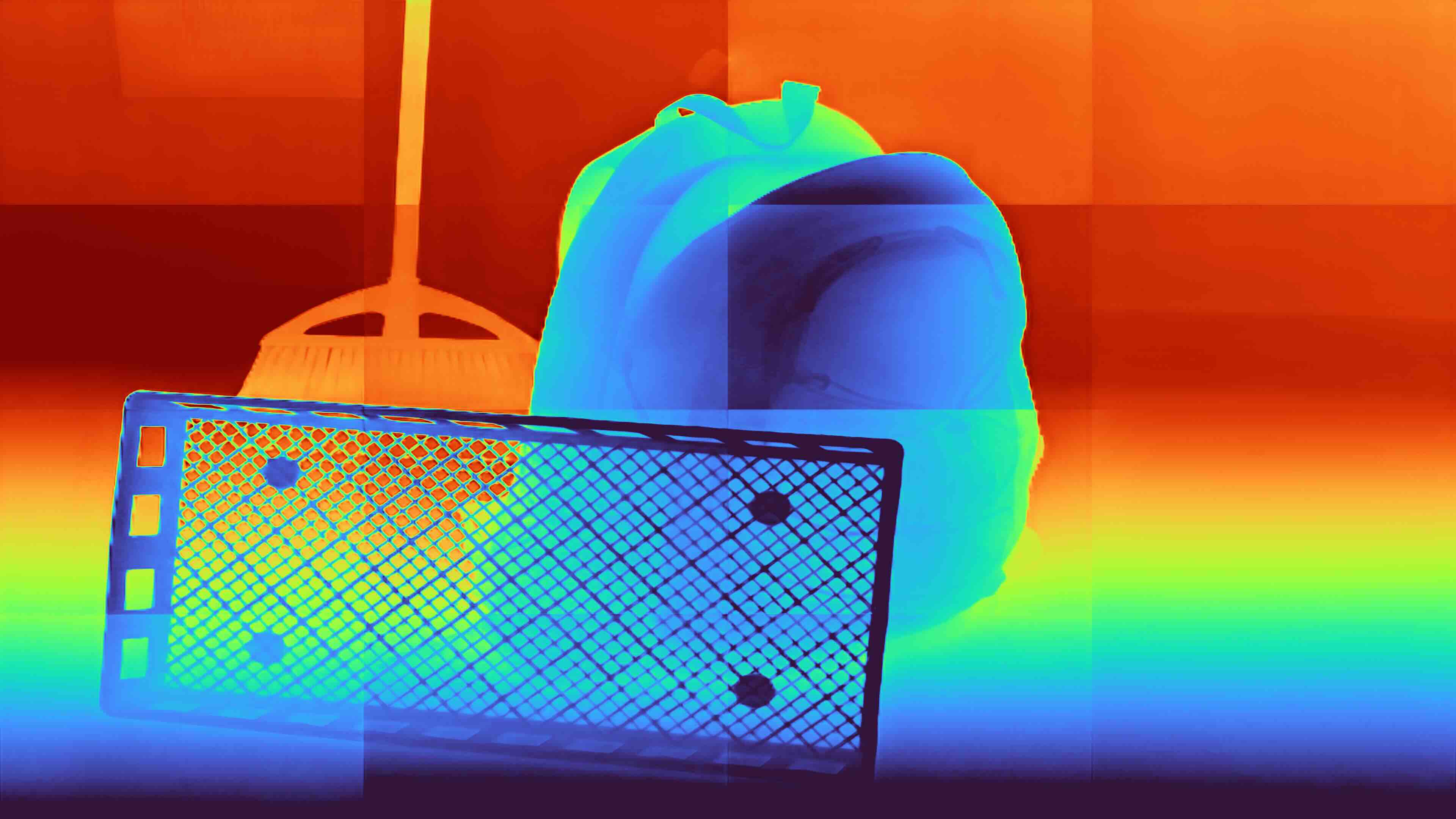}
        \put(64,5){\makebox[0pt]{\adjincludegraphics[height=0.1\linewidth,trim={{.05\width} {.1\height} {.6\width} {.5\height}},clip, cfbox=black 2pt 0cm]{sec/fig/middlebury_sample/18_pr.jpg}}}
    \end{overpic}
    &
    \begin{overpic}[width=0.24\linewidth,height=0.135\linewidth]{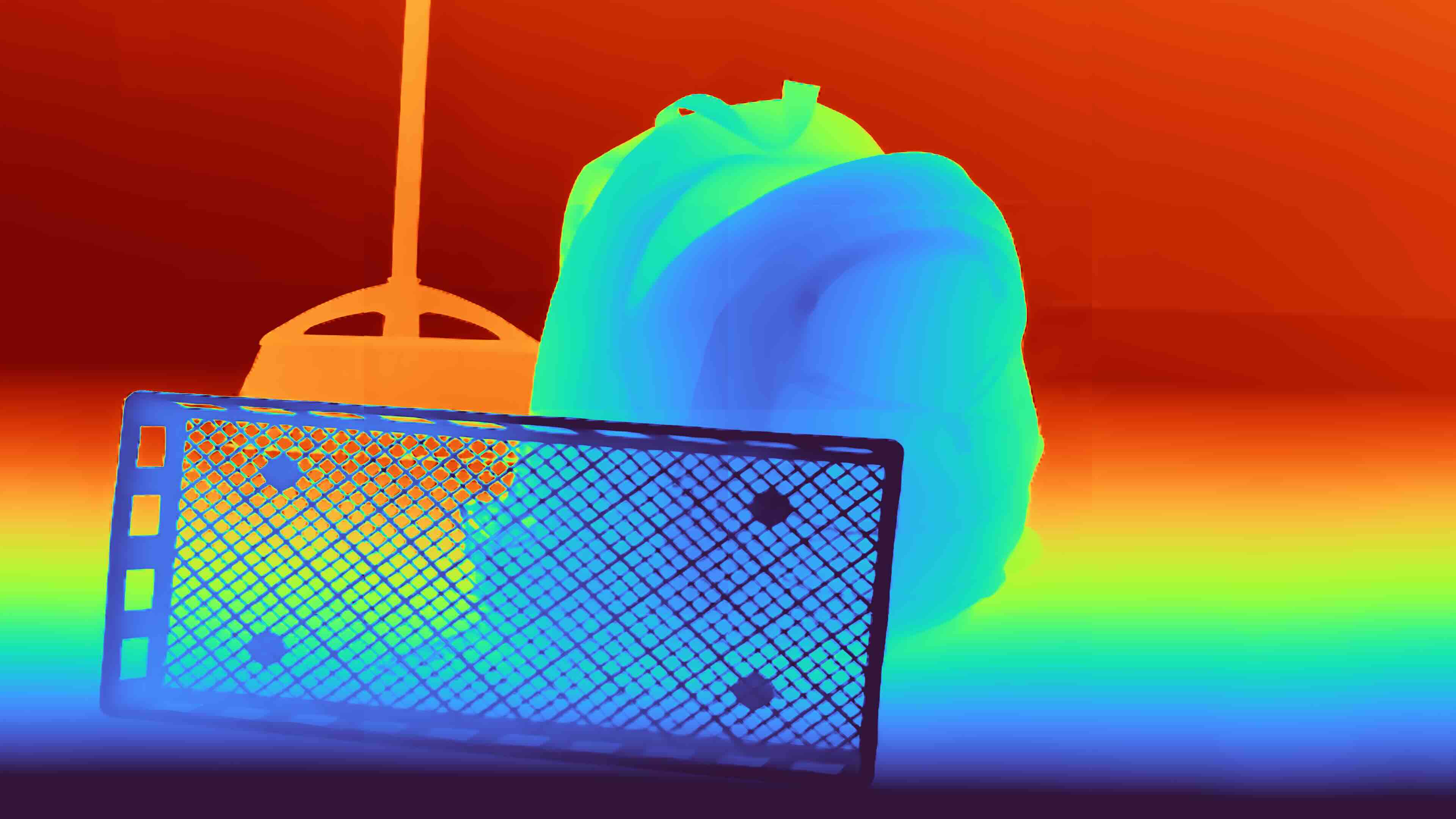}
        \put(64,5){\makebox[0pt]{\adjincludegraphics[height=0.1\linewidth,trim={{.05\width} {.1\height} {.6\width} {.5\height}},clip, cfbox=black 2pt 0cm]{sec/fig/middlebury_sample/18_our.jpg}}}
    \end{overpic}
    &
    \includegraphics[width=0.24\textwidth]{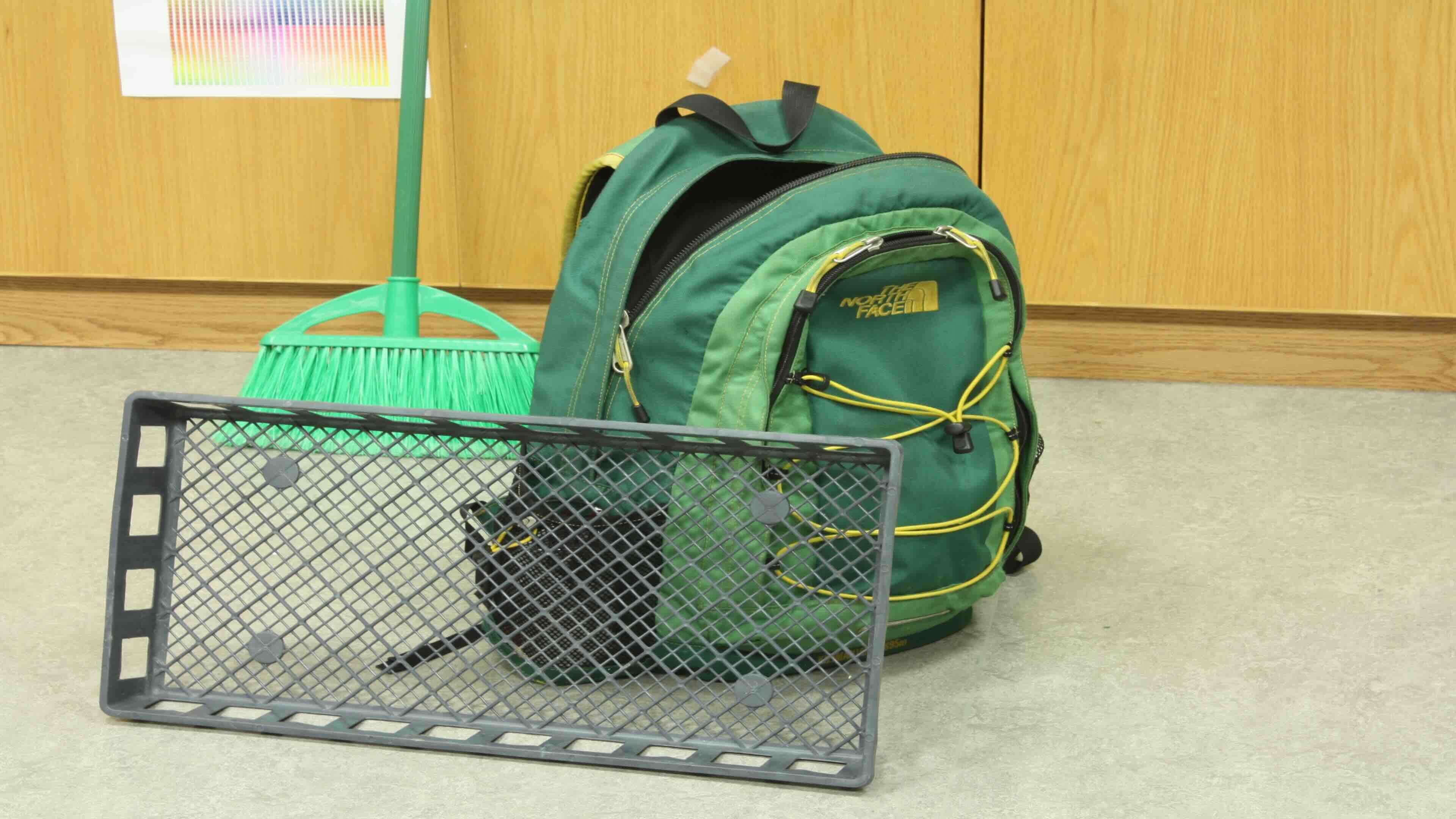} \\[-2pt]
    % -------- Column labels --------
    \textbf{\small Depth Anything V2} &
    \textbf{\small PatchRefiner} &
    \textbf{\small Ours} &
    \textbf{\small Input}
  \end{tabular}
  \hspace{-10pt}
  \caption{
  \textbf{Qualitative comparison on the Zero-Shot depth estimation task.}
  Each row shows samples from different scenes. 
  The rightmost column presents the RGB input.
  Compared to previous models, our zero-shot predictions demonstrate improved geometric consistency and sharper depth boundaries across diverse domains.
  }
  \label{fig:zeroshot_depth}
\end{figure*}

\subsection{Geometrically Consistent Supervision}
\label{sec:normal_supervision}
We supervise both depth and normals in a unified manner so that the two predictions are constrained by the same underlying geometry.
\vspace{-5pt}
\paragraph{Depth loss.}
Let $\DMaps^{refined}$ denote the refined depth prediction and $\DMaps^{gt}$ the ground-truth depth.
We combine a point-wise reconstruction term with a gradient term to encourage both accurate values and sharp depth predictions.
In compact form, the depth loss is written as
\begin{equation}
  \begin{aligned}
    \Loss{depth}
    &= \Loss{MSE}\big(\DMaps^{refined}, \DMaps^{gt}\big) \\
    &\quad + \lambda_{\text{grad}}\,\Loss{grad}\big(\DMaps^{refined}, \DMaps^{gt}\big),
  \end{aligned}
\end{equation}
where $\Loss{MSE}$ denotes a pixel-wise mean squared error and $\Loss{grad}$ measures horizontal and vertical depth gradients, with $\lambda_{\text{grad}}$ balancing the two terms.
\vspace{-5pt}
\paragraph{Normal loss.}
To enable geometrically consistent supervision, we derive a pseudo-normal field $\SNormal^{pseudo}$ from $\DMaps^{gt}$ using a standard least-squares method on local depth neighborhoods~\cite{qi2018geonet}. 
Let $\SNormal^{refined}$ denote the predicted normal map produced by our network.
We define the normal supervision term in an abstract form as
\begin{equation}
  \begin{aligned}
    \mathcal{L}_{\text{normal}}
    &= \Loss{angle}\big(\SNormal^{refined}, \SNormal^{pseudo}\big) \\
    &\quad + \lambda_{\text{mse}}\,\Loss{MSE}\big(\SNormal^{refined}, \SNormal^{pseudo}\big),
  \end{aligned}
  \label{eq:normal_loss}
\end{equation}
where $\Loss{angle}$ penalizes orientation differences and $\Loss{MSE}$ enforces per-pixel alignment, with $\lambda_{\text{mse}}$ controlling their relative weight.

The final training objective combines the depth and normal losses:
\begin{equation}
  \begin{aligned}
    \Loss{total}
    &= \lambda_{\text{depth}}\,\Loss{depth}\big(\DMaps^{refined}, \DMaps^{gt}\big) \\
    &\quad + \lambda_{\text{normal}}\,\Loss{normal}\big(\SNormal^{refined}, \SNormal^{pseudo}\big),
  \end{aligned}
\end{equation}
with $\lambda_{\text{depth}}$ and $\lambda_{\text{normal}}$ controlling the relative contribution of each term.
Since $\SNormal^{pseudo}$ is fully determined by the ground-truth depth, both depth and normal heads are ultimately constrained by the same underlying 3D geometry, which encourages their predictions to remain geometrically consistent. The model supports separate training modes to predict depth alone by disabling the normal loss, or vice versa by disabling the depth loss.

\vspace{-10pt}

\section{Experiments}

\begin{figure*}[t]
  \centering
  \setlength{\tabcolsep}{1pt}
  \renewcommand{\arraystretch}{1.0}
  \begin{tabular}{cccc}
    % -------- Row 1 --------
    \includegraphics[width=0.24\textwidth]{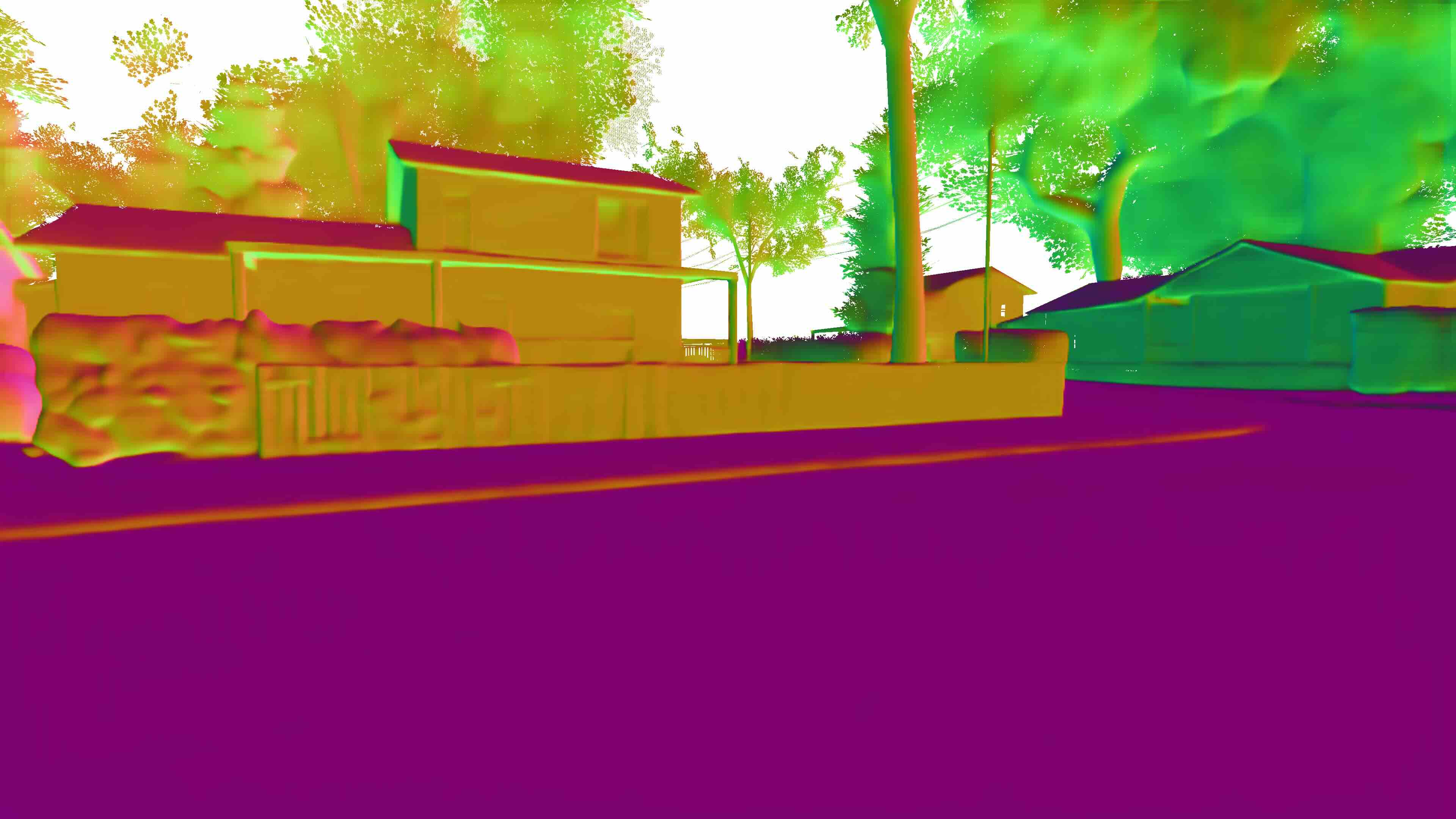} &
    \includegraphics[width=0.24\textwidth]{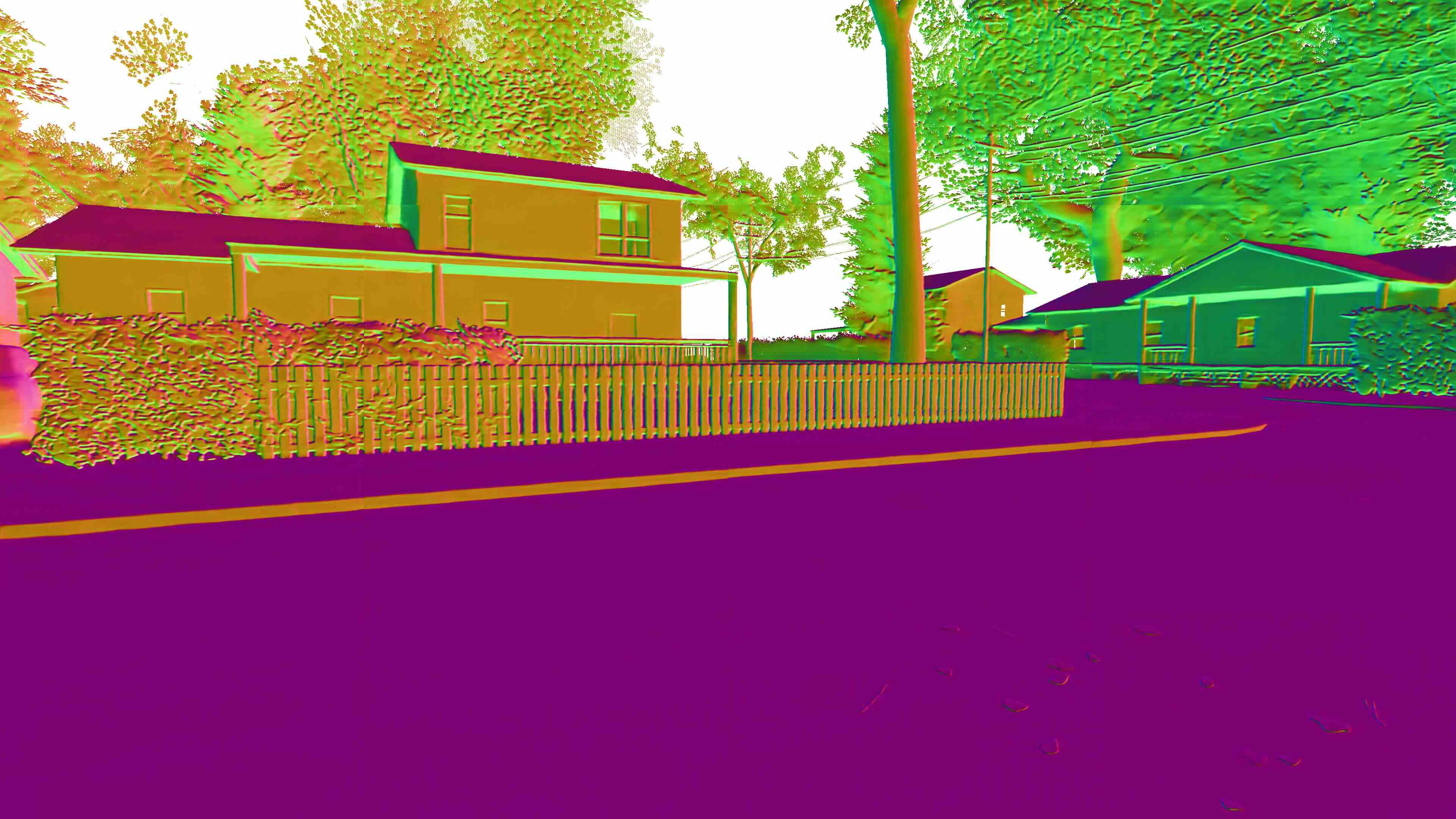} &
    \includegraphics[width=0.24\textwidth]{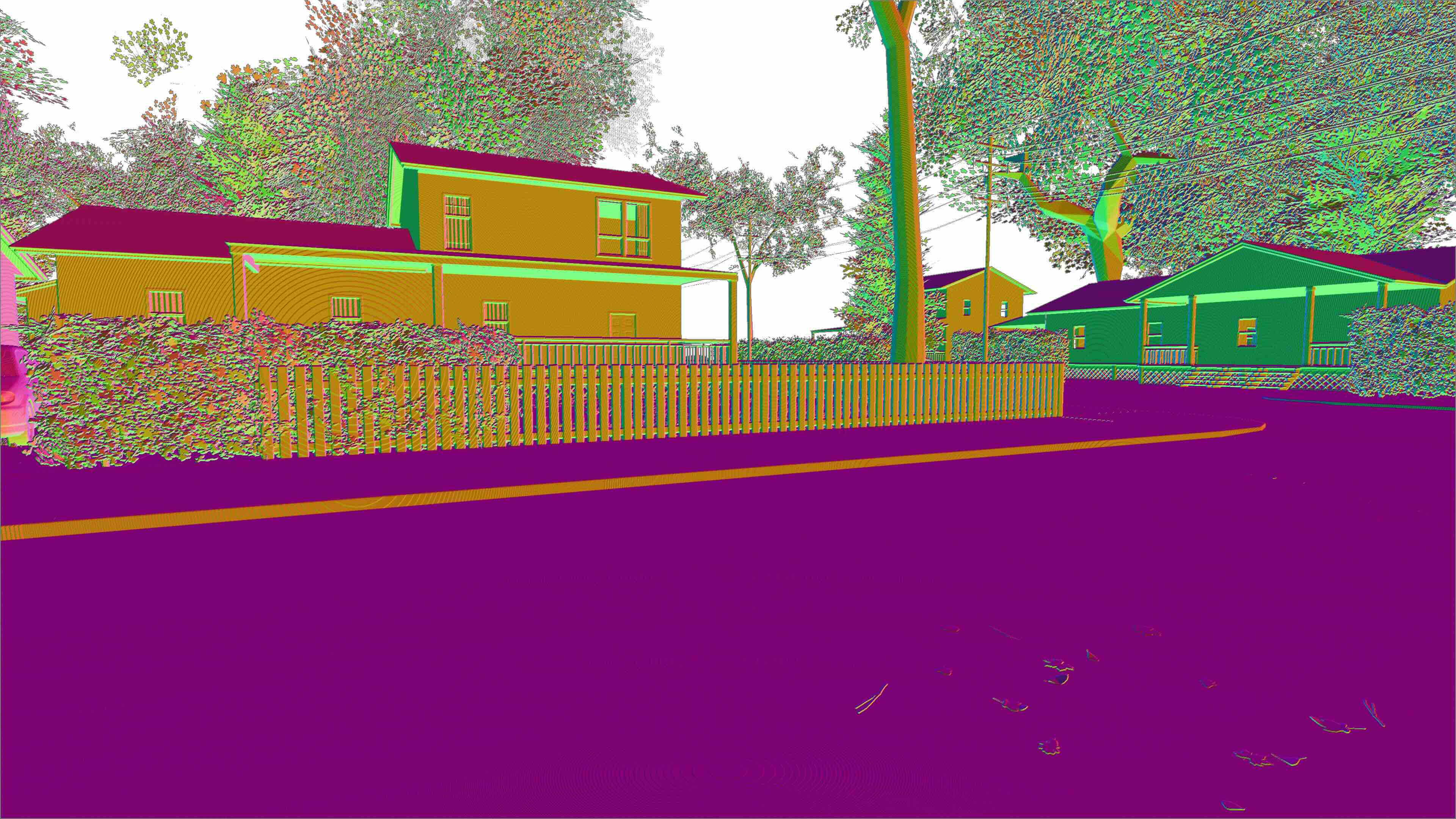} &
    \includegraphics[width=0.24\textwidth]{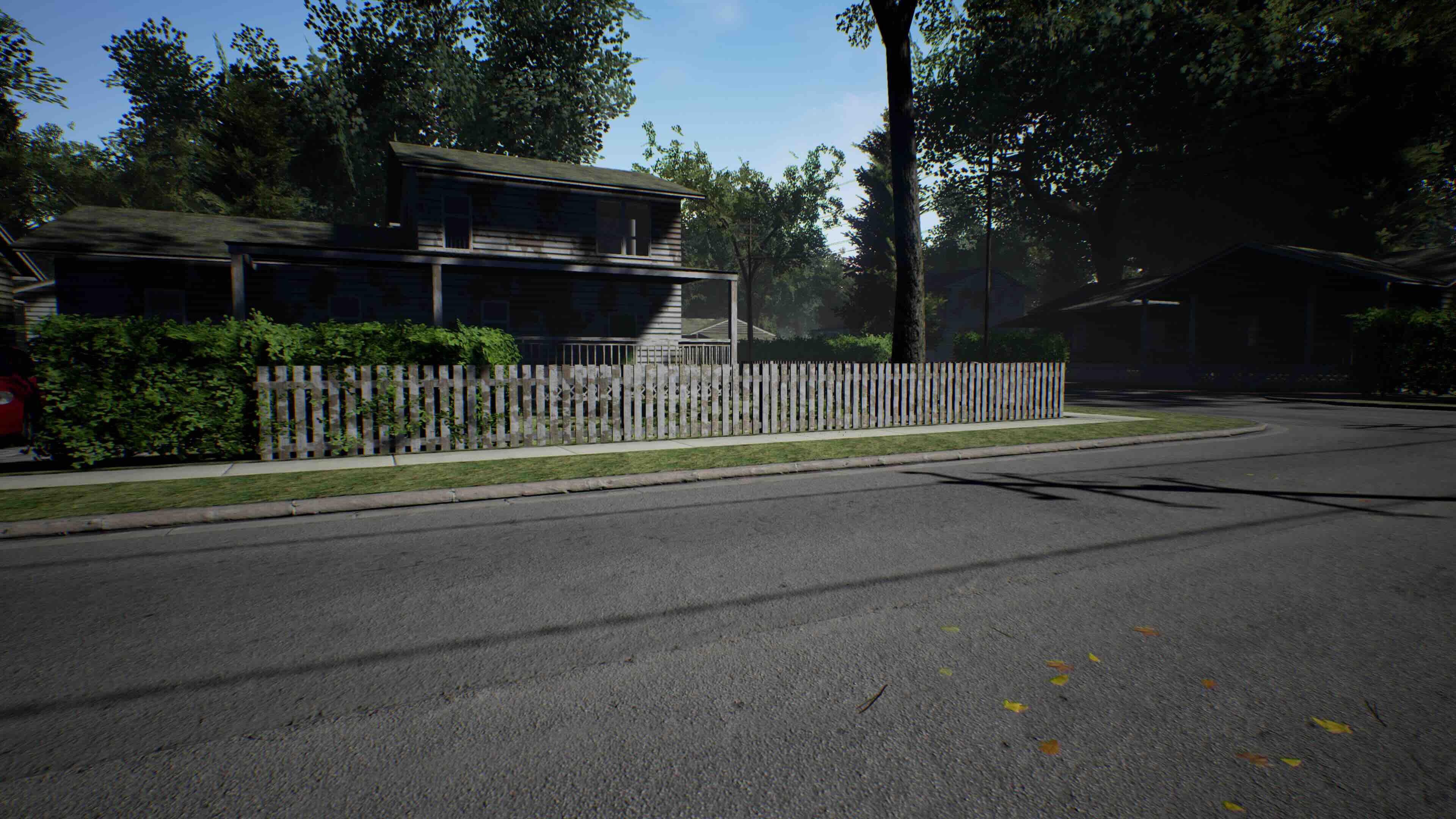} \\[-2pt]
    % -------- Row 2 --------
    \includegraphics[width=0.24\textwidth]{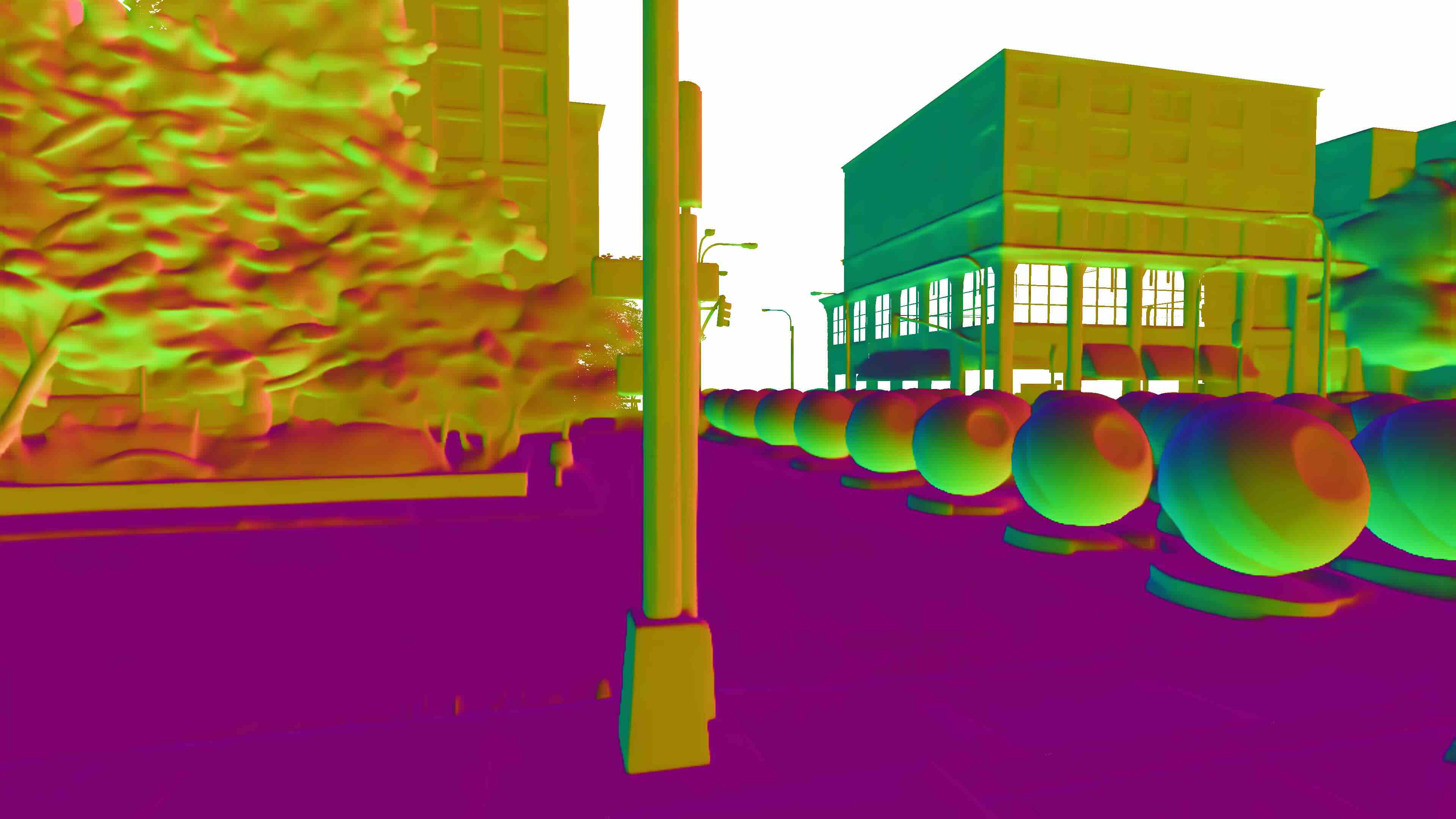} &
    \includegraphics[width=0.24\textwidth]{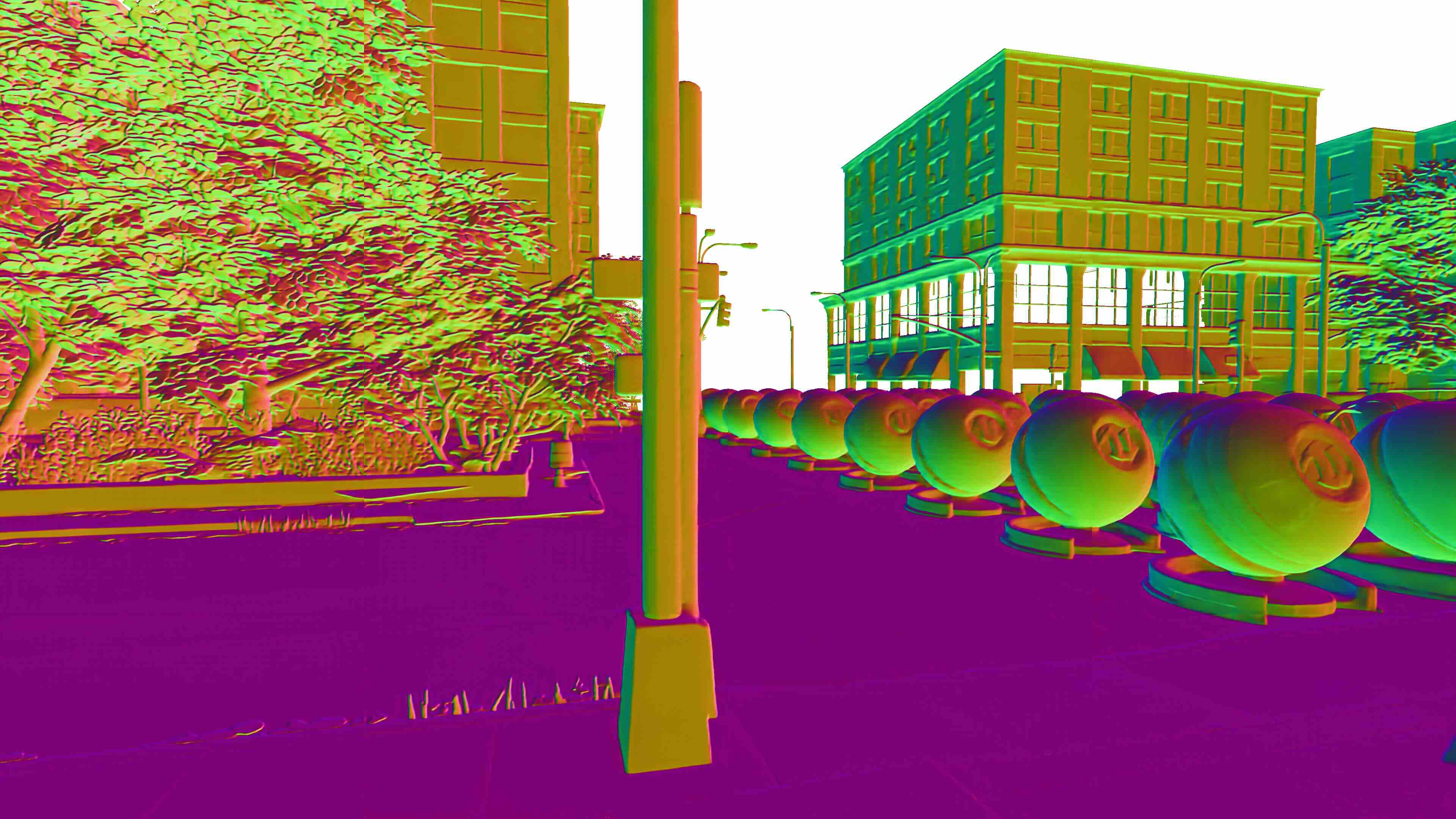} &
    \includegraphics[width=0.24\textwidth]{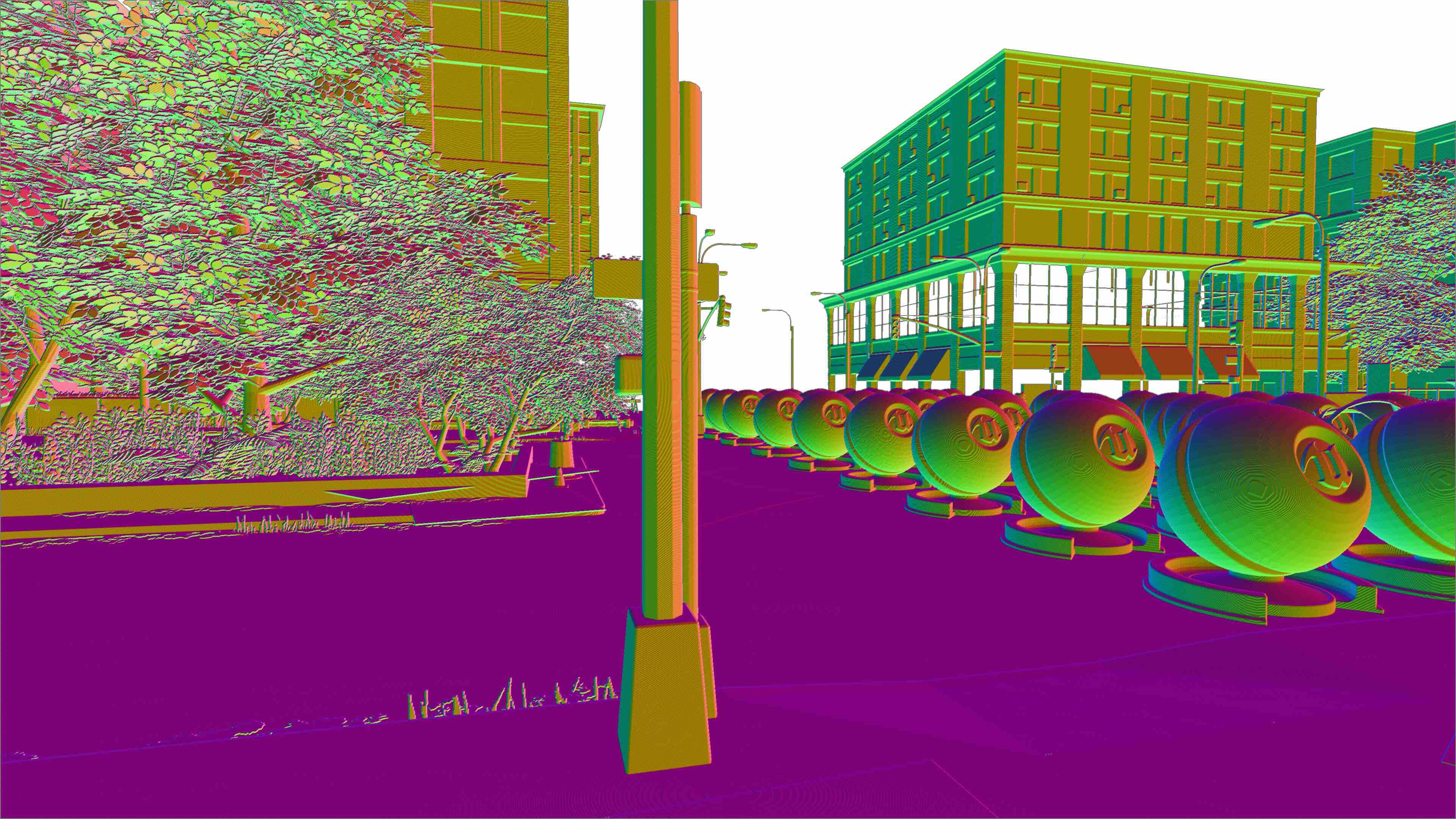} &
    \includegraphics[width=0.24\textwidth]{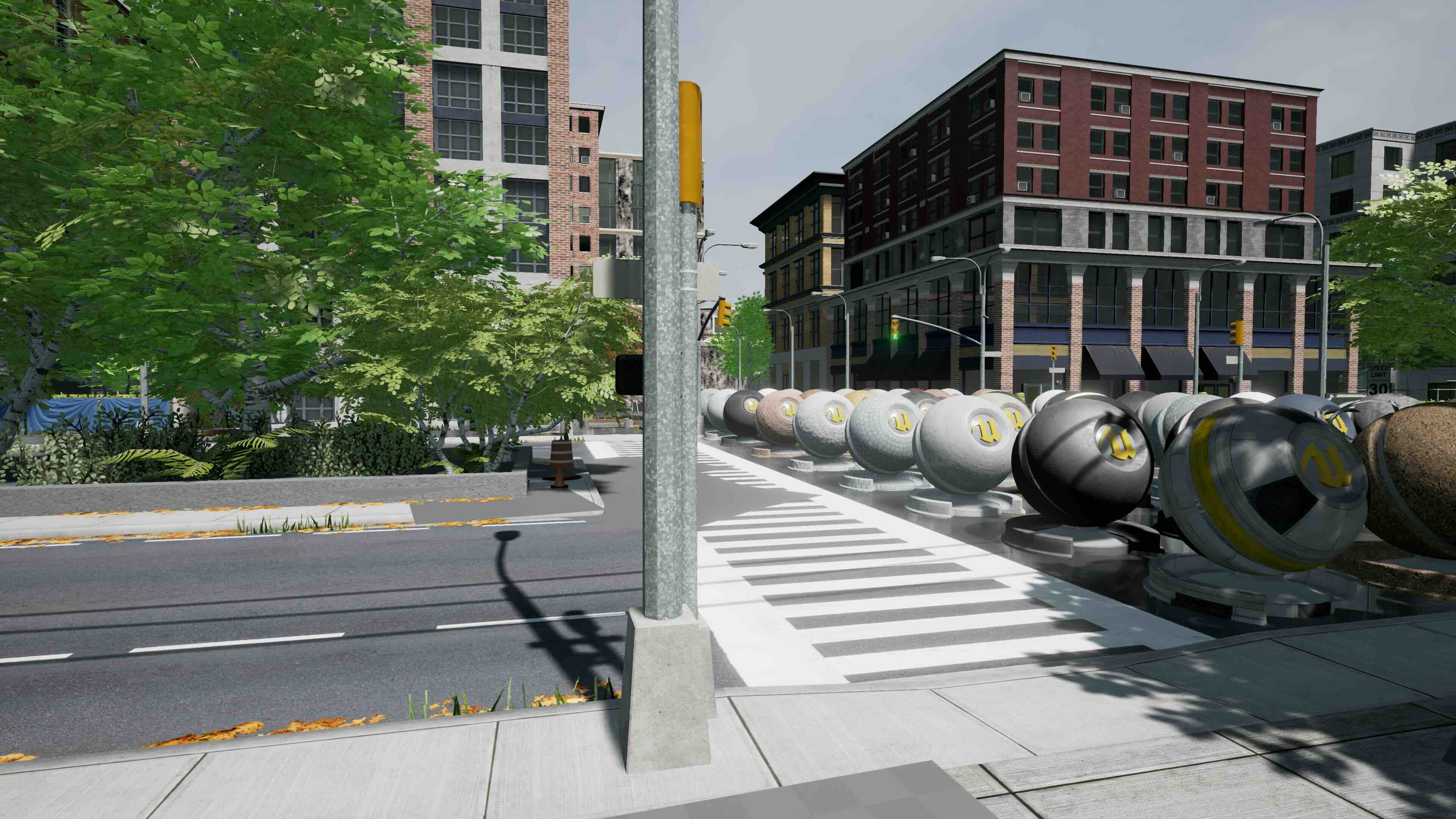} \\[-2pt]

    % -------- Column labels --------
    \textbf{\small Metric3D V2}~\cite{hu2025metric3dv2} &
    \textbf{\small Ours} &
    \textbf{\small GT Normal} &
    \textbf{\small RGB Input} 

  \end{tabular}
  \vspace{-10pt}
  \caption{
  \textbf{Qualitative comparison of surface normal estimation on the UnrealStereo4K~\cite{u4k} dataset.}
  Our method produces more accurate and spatially consistent surface normals compared to Metric3D~V2.
  }
  \label{fig:u4k_normal}
  \vspace{-10pt}
\end{figure*}

\subsection{Experimental Setup}
\noindent \textbf{Datasets.}
We conduct experiments on UnrealStereo4K~\cite{u4k}, Booster~\cite{zamaramirez2022booster}, ETH3D~\cite{ETH3D}, and Middlebury 2014~\cite{middlebury2014}; see the supplementary material (Sec.~\ref{supp:implement}) for more details.

\noindent \textbf{Evaluation metrics.}
For depth estimation, we adopt standard metrics including Absolute Relative Error (AbsRel), Root Mean Squared Error (RMSE), and the accuracy threshold $\delta_1$. 
To assess the spatial consistency of depth predictions across patches, we additionally employ the Consistency Error (CE)~\cite{li2024patchfusion}. Following SharpDepth~\cite{pham2024sharpdepthsharpeningmetricdepth}, we further evaluate boundary sharpness using the Pseudo Depth Boundary Error (PDBE). 
Since annotated depth edges are unavailable for synthetic datasets, we extract edges from both predicted and ground-truth depth maps using the Canny operator, and compute PDBE accuracy $\epsilon_{\mathrm{PDBE}}^{\mathrm{acc}}$ and completeness $\epsilon_{\mathrm{PDBE}}^{\mathrm{compl}}$. For surface normal estimation, we report the mean, median, and root mean squared (RMS) angular error, along with the percentage of pixels whose angular error falls below $5^\circ$, $11.25^\circ$, and $30^\circ$.

\noindent \textbf{Implementation details.}
All implementation details are provided in the supplementary material (Sec.~\ref{supp:implement}).

\begin{table}[t]
\vspace{-4pt}
\centering
\caption{
\textbf{Zero-shot depth estimation benchmark across multiple real-world datasets.}
We evaluate depth accuracy on Booster~\cite{zamaramirez2022booster}, ETH3D~\cite{ETH3D}, and Middlebury 2014~\cite{middlebury2014}. 
DAV2 and PR are short for DepthAnythingV2 and PatchRefiner, respectively.
Best results are shown in \textbf{bold}.
}

\vspace{-5pt}

\label{depth:benchmark}
\scalebox{0.75}{

\newcommand{\best}[1]{\textcolor{red}{\mathbf{#1}}}
\newcommand{\secondbest}[1]{\textcolor{orange}{#1}}

\begin{tabular}{L{2.1cm} | C{1cm}C{1cm} | C{1cm}C{1cm} | C{1cm}C{1cm} }
\toprule
\multirow{2}{*}{\textbf{Methods}} 
& \multicolumn{2}{c|}{\textbf{Booster}} 
& \multicolumn{2}{c|}{\textbf{ETH3D}} 
& \multicolumn{2}{c}{\textbf{Middle2014}} 
\\
& AbsRel$\downarrow$ & $\delta_1\uparrow$ 
& AbsRel$\downarrow$ & $\delta_1\uparrow$ 
& AbsRel$\downarrow$ & $\delta_1\uparrow$ 
\\
\midrule
DAV2~\cite{yang2025depthanythingv2}   
& 0.0274 & 0.996
& 0.0507 & 0.978 
& 0.0307 & 0.994 
\\
PR$_{p=16}$~\cite{li2024patchrefiner}  
& 0.0404 & 0.988 
& 0.0608 & 0.968 
& 0.0380 & 0.987     
\\
PR$_{p=49}$~\cite{li2024patchrefiner}  
& 0.0382 & 0.989 
& 0.0577 & 0.972 
& 0.0376 & 0.987    
\\
PRO~\cite{kwon2025onelook} 
& 0.0287 & \textbf{0.997}
& 0.0448 & 0.981
& 0.0287 & \textbf{0.996}
\\
\textbf{Ours\hspace{0.3em} {\footnotesize \color{gray} Separate}}
& 0.0269 & 0.996
& 0.0446 & 0.981
& 0.0291 & \textbf{0.996}
\\
\textbf{Ours\hspace{0.3em} {\footnotesize \color{gray} Joint}}
& \textbf{0.0248} & \textbf{0.997}
& \textbf{0.0434} & \textbf{0.983} 
& \textbf{0.0280} & \textbf{0.996}  
\\
\bottomrule
\end{tabular}
}
\vspace{-10pt}
\end{table}

\subsection{Results}

\noindent\textbf{Depth Estimation.}
Table~\ref{tab:u4k} summarizes the quantitative results on the UnrealStereo4K dataset. 
Our separate depth model achieves state-of-the-art performance across all metrics, reaching \textbf{0.0295} AbsRel, 
\textbf{0.982} $\delta_1$, and \textbf{1.38} RMSE at only \textbf{0.94\,s} per 4K image. 
Joint depth–normal model further achieves accuracy to \textbf{0.0291} AbsRel, \textbf{0.983} $\delta_1$, and \textbf{1.31} RMSE. 
Compared with PatchRefiner~\cite{li2024patchrefiner}, our approach reduces AbsRel by over \textbf{49\%} and RMSE by more than \textbf{35\%}, 
highlighting the effectiveness of multi-patch reasoning and cross-patch attention in capturing both global structure and fine local geometry. 
Our model also achieves the lowest Consistency Error (CE, \textbf{0.0415}) and improved PDBE sharpness 
($\epsilon_{\mathrm{PDBE}}^{\mathrm{acc}}{=}1.40$, $\epsilon_{\mathrm{PDBE}}^{\mathrm{compl}}{=}64.21$), producing smoother boundaries and crisper depth edges. 
Qualitative examples in Fig.~\ref{fig:u4k} further show that our method preserves thin structures and avoids the block-wise artifacts common in patch-based refinement pipelines.

\begin{table}[t]
  \vspace{-4pt}
  \centering
  \small
  \caption{Ablation on GridMix Patch Sampling Strategy.}
  \vspace{-10pt}
  \label{tab:training_grid}
  \resizebox{\columnwidth}{!}{
  \begin{tabular}{l|cccc}
    \toprule
    \begin{tabular}{c}
    \textbf{Configurations} \\
    \textbf{($\vProb{1},\vProb{2},\vProb{3},\vProb{4}$)}
    \end{tabular}
    & AbsRel $\downarrow$ & $\delta_1$ $\uparrow$ & RMSE $\downarrow$ & CE $\downarrow$ \\
    \midrule
    Depth-Anything v2                 & 0.0812 & 0.924 & 2.86 & --      \\
    (1,0,0,0)             & 0.0500 & 0.963 & 2.01 & 0.0648 \\
    (0.5,0.5,0,0)         & 0.0473 & 0.966 & 1.79 & 0.0436 \\
    (0.1,0.2,0.3,0.4)     & \textbf{0.0295} & \textbf{0.982} & \textbf{1.35} & \textbf{0.0418} \\
    (0,0,0,1)             & 0.0321 & 0.980 & 1.42 & 0.0635 \\
    \bottomrule
  \end{tabular}
  }
\vspace{-10pt}
\end{table}

\noindent\textbf{Zero-shot Evaluation.}
We additionally assess zero-shot generalization on Booster, ETH3D, and Middlebury 2014 (Table~\ref{depth:benchmark}). 
Our method delivers strong and consistent performance across all benchmarks, outperforming prior refinement approaches. 
On Booster, the joint model achieves the lowest AbsRel (\textbf{0.0248}) while matching PRO on $\delta_1$ (\textbf{0.997}). 
On ETH3D and Middlebury 2014, our model also outperforms PRO, obtaining the best AbsRel (\textbf{0.0434} vs.\ \textbf{0.0448} on ETH3D, \textbf{0.0280} vs.\ \textbf{0.0287} on Middlebury) and $\delta_1$ (\textbf{0.983} vs.\ \textbf{0.981}, \textbf{0.996} vs.\ \textbf{0.996}), while consistently improving over both PatchRefiner variants and the separate variant. Joint training offers mutual benefits, exploiting depth-normal coupling to achieve better refinement. These results demonstrate that the proposed multi-patch transformer retains robust zero-shot generalization and produces reliable geometry across diverse real-world scenes.

\noindent\textbf{Surface Normal Estimation.}
We also evaluate the model's ability to predict dense surface normals (Table~\ref{tab:u4k}, bottom). 
Under the joint training setting, our method achieves substantial improvements over Metric3D~v2~\cite{hu2025metric3dv2}, reducing the mean angular error from \textbf{23.36} to \textbf{18.27} and RMS from \textbf{13.90} to \textbf{9.42}. 
Our model also yields markedly higher threshold accuracies, achieving \textbf{29.88\%} within $5^\circ$, \textbf{59.79\%} within $11.25^\circ$, and \textbf{85.43\%} within $30^\circ$. 
These results confirm that jointly modeling depth and normals within the multi-patch transformer framework enhances geometric consistency and produces sharper, more stable high-frequency surface details.

\noindent \textbf{Extension to Any Resolution.} To demonstrate the scalability of our approach to estimate high-resolution depth maps and surface normals at arbitrary input resolutions, we run an experiment on an 8K in-the-wild image as illustrated in the Teaser figure. Leveraging the adaptive multi-patch framework, our method seamlessly handles varying scales without compromising accuracy or necessitating resolution-specific training, enabling robust performance across diverse real-world scenarios. This is evident in the enhanced preservation of fine details, such as thin structures (e.g., the metallic box) and high-frequency textures (e.g., plant in the first plane), where our outputs surpass those of zero-shot baselines, while maintaining global geometric consistency.

\begin{figure}[t]
  \centering
  \setlength{\tabcolsep}{2pt}
  \renewcommand{\arraystretch}{1.0}
  \begin{tabular}{cc}
    \includegraphics[width=0.47\columnwidth]{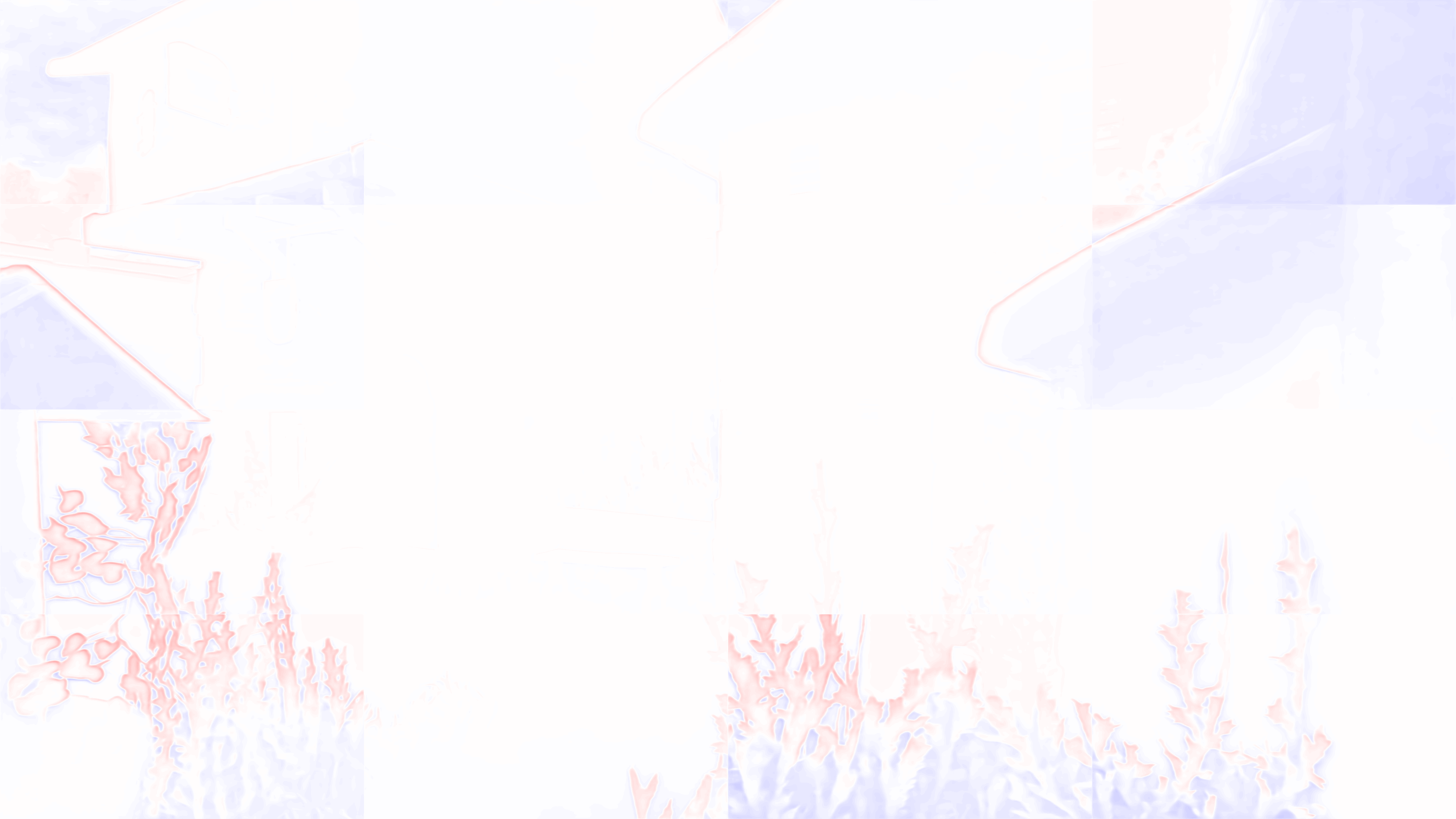} &
    \includegraphics[width=0.47\columnwidth]{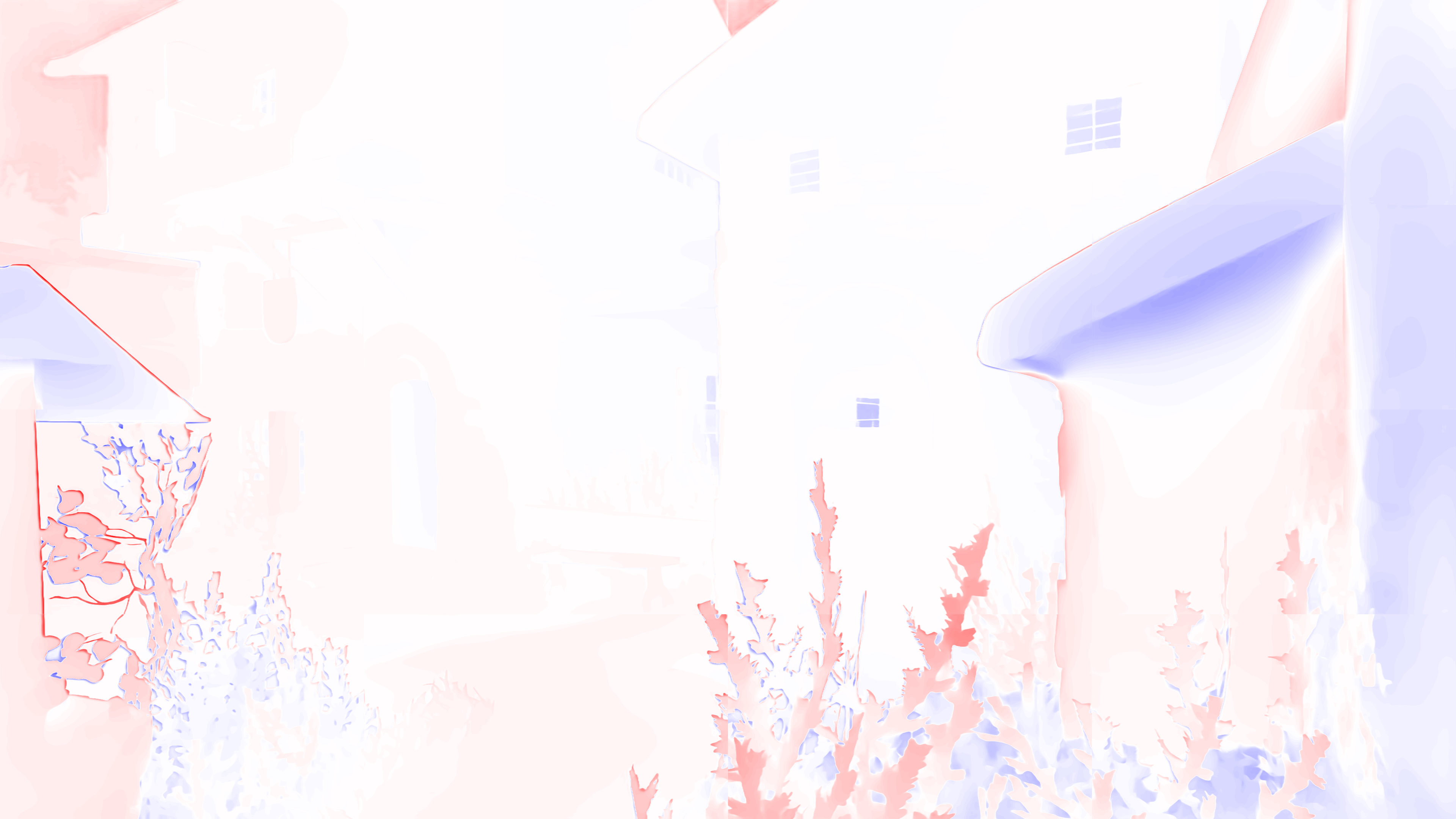} \\
    \textbf{\small (a) Offset w/ Local RoPE} &
    \textbf{\small (b) Offset w/ Global RoPE}
  \end{tabular}
  \vspace{-10pt}
  \caption{
    \textbf{Ablation study on Global Positional Encoding.}
    Comparison of offset predictions under different positional encoding strategies.
    The \textbf{Global RoPE} enhances cross-patch geometric alignment and yields smoother transitions across patch boundaries compared to the local variant.
  }
  \label{fig:ablation_pos}
\end{figure}

\begin{table}[t]
  \setlength{\tabcolsep}{3pt} 
  \caption{Ablation study on Global Postional Encoding.}
  \vspace{-10pt}
  \label{tab:ablation_pos}
  \centering
  \resizebox{\columnwidth}{!}{
  \begin{tabular}{l|cccc}
    \toprule
    \textbf{Position Embedding} & AbsRel $\downarrow$ & $\delta_1$ $\uparrow$ & RMSE $\downarrow$ & CE $\downarrow$ \\
    \midrule
    Local RoPE   & 0.0343 & 0.976 & 1.60 & 0.2830 \\
    Global RoPE & \textbf{0.0321} & \textbf{0.980} & \textbf{1.42} & \textbf{0.0635} \\
    \bottomrule
  \end{tabular}
  }
\end{table}

\subsection{Ablation Studies}

\noindent \textbf{GridMix Patch Sampling Strategy.}
Table~\ref{tab:training_grid} shows that using a single grid configuration (e.g., $1{\times}1$ or $4{\times}4$) limits performance and generalization. 
Mixing multiple grid sizes yields clear improvements, with the best results obtained when sampling with probabilities $(0.1, 0.2, 0.3, 0.4)$, achieving \textbf{0.0295} AbsRel and \textbf{0.982} $\delta_1$. 
This setting also reduces RMSE to \textbf{1.35} and maintains low CE ($0.0418$), unlike the fixed $4{\times}4$ grid which suffers from poor consistency. 
These results indicate that probabilistic grid sampling effectively enhances inter-patch coherence while preserving fine geometric details.
More details are provided in the supplementary material (Sec.~\ref{supp:ablation}).

\noindent \textbf{Global Positional Encoding.}
We compare local and global RoPE variants in Table~\ref{tab:ablation_pos} and Fig.~\ref{fig:ablation_pos}. 
Global RoPE not only improves numerical accuracy (AbsRel from 0.0343 to \textbf{0.0321}, RMSE from 1.60 to \textbf{1.42}) 
but also yields a significantly lower CE (0.2830 → \textbf{0.0635}), indicating much stronger cross-patch geometric consistency. 
As shown in Fig.~\ref{fig:ablation_pos}, the offsets predicted with Global RoPE exhibit smoother transitions across patch boundaries 
and fewer discontinuities, whereas the local variant shows visible misalignment artifacts. 
These results demonstrate that incorporating global positional context helps the transformer reason over long-range spatial relationships, 
leading to more coherent geometry reconstruction.

\begin{figure}[t]
  \centering
  \setlength{\tabcolsep}{2pt}
  \begin{tabular}{cc}
    % -------- Row 1: Predictions --------
    \includegraphics[width=0.49\columnwidth]{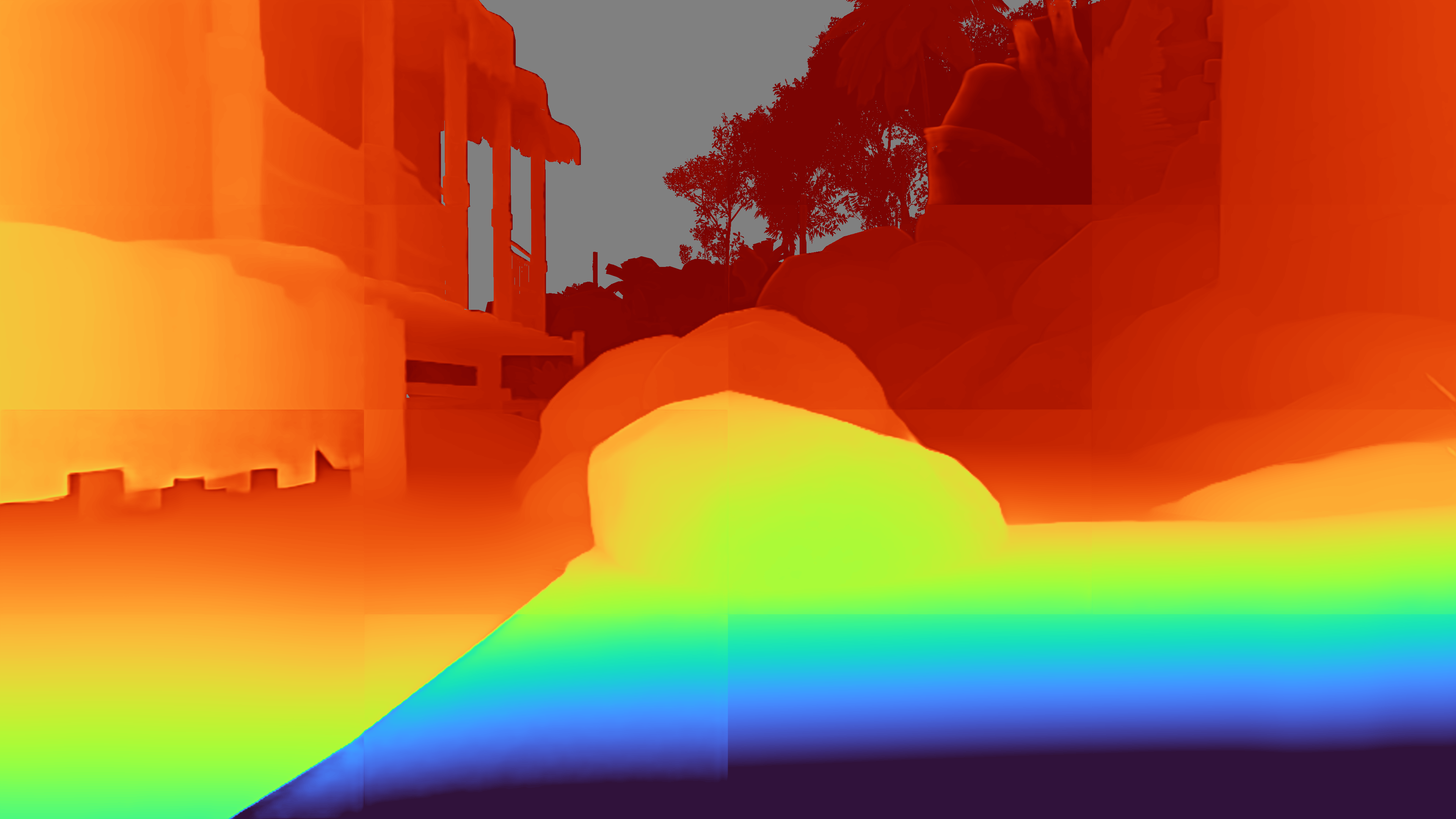} &
    \includegraphics[width=0.49\columnwidth]{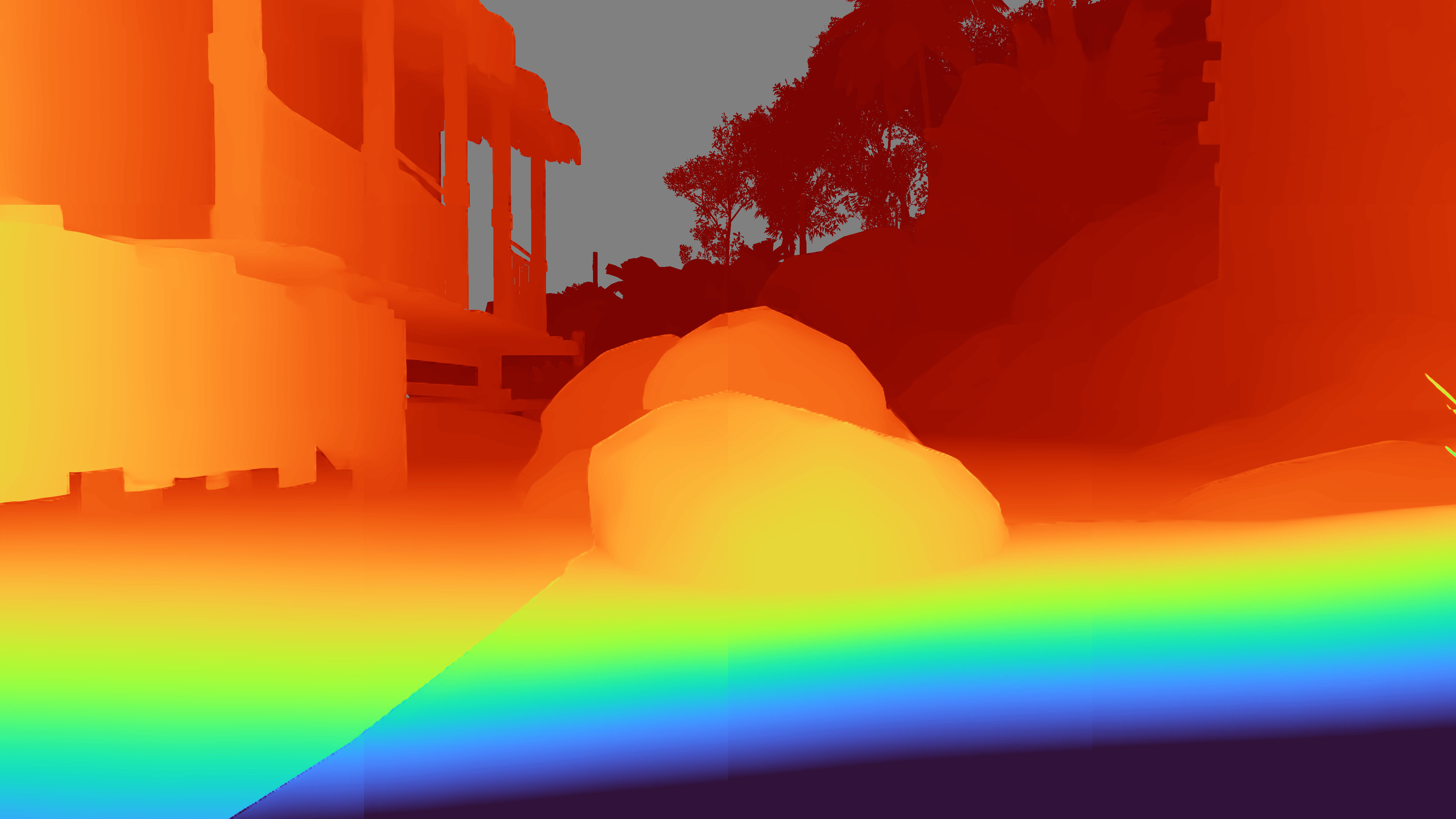} \\[2pt]
    {\scriptsize \textbf{(a) Pred w/o Cross-Patch Attention}} &
    {\scriptsize \textbf{(b) Pred w/ Cross-Patch Attention}} \\[4pt]

    % -------- Row 2: Offset Maps --------
    \includegraphics[width=0.49\columnwidth]{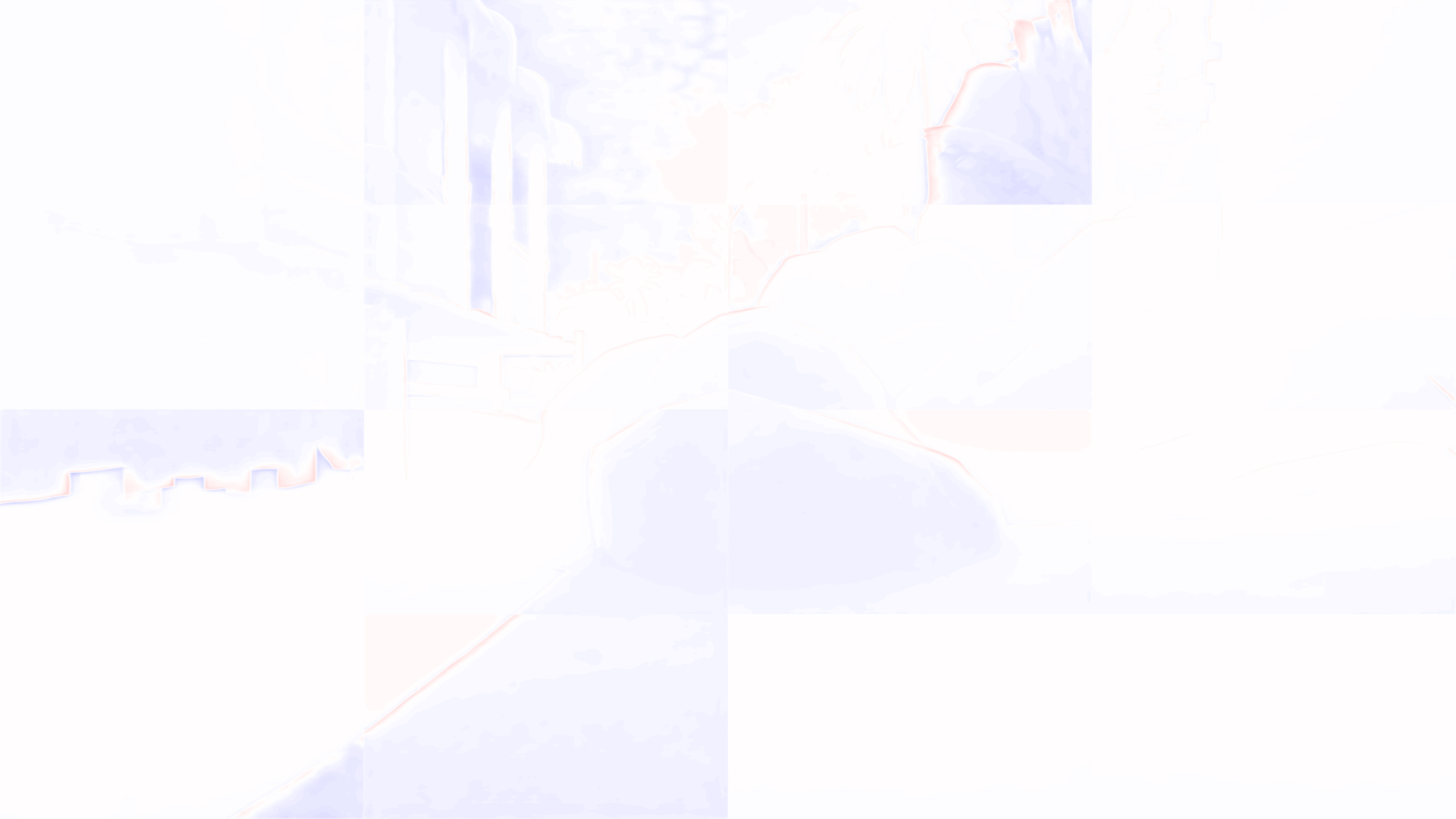} &
    \includegraphics[width=0.49\columnwidth]{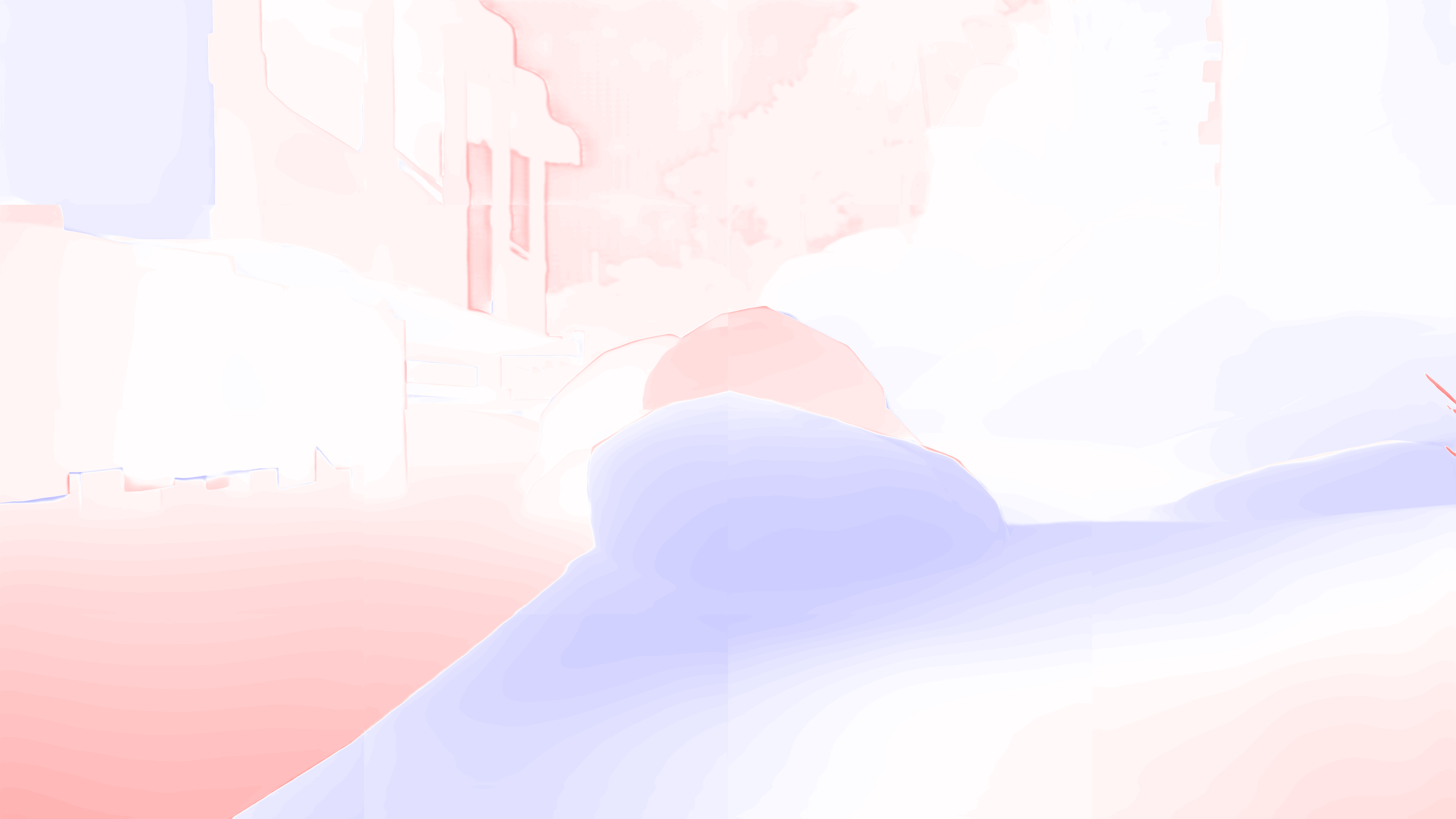} \\[2pt]
    {\scriptsize \textbf{(c) Offset w/o Cross-Patch Attention}} &
    {\scriptsize \textbf{(d) Offset w/ Cross-Patch Attention}}
  \end{tabular}

  \caption{
    \textbf{Ablation on Cross-Patch Attention.}
    Visual comparison of both depth predictions and offset maps with and without cross-patch attention.
    Cross-patch attention significantly improves inter-patch geometric coherence, yielding smoother transitions and more stable boundary structures across patch regions.
  }
  \label{fig:ablation_crosspatch}
\end{figure}

\noindent \textbf{Cross-Patch Attention.}
We evaluate the effect of cross-patch attention, which enables long-range interaction among spatially separated patches. 
As shown in Table~\ref{tab:ablation_crosspatch} and Fig.~\ref{fig:ablation_crosspatch}, disabling this module results in higher errors 
(AbsRel 0.0678 → \textbf{0.0500}, RMSE 2.51 → \textbf{2.01}) and visible boundary discontinuities between neighboring regions. 
The predicted offset maps further highlight these artifacts: without cross-patch attention, patch seams remain strongly pronounced, 
whereas our full model produces smooth transitions and coherent structures. 
These results confirm that global token communication is essential for maintaining inter-patch geometric consistency 
and generating seamless high-resolution depth maps.

\begin{table}[t]
  \vspace{-4pt}  
  \centering
  \small
  \caption{Ablation on Cross-Patch Attention.}
  \vspace{-10pt}
  \label{tab:ablation_crosspatch}
  \resizebox{\columnwidth}{!}{
  \begin{tabular}{l|ccc}
    \toprule
    \textbf{Attention Setting} & AbsRel $\downarrow$ & $\delta_1$ $\uparrow$ & RMSE $\downarrow$ \\
    \midrule
    w/o Cross-Patch Attention & 0.0678 & 0.945 & 2.51 \\
    w/ Cross-Patch Attention & \textbf{0.0500} & \textbf{0.963} & \textbf{2.01} \\
    \bottomrule
  \end{tabular}
  }
\vspace{-4pt}  
\end{table}

\vspace{-2pt}
\section{Conclusion}
We presented a unified transformer framework that adapts VGGT from multi-view 3D reconstruction to high-resolution single-image geometry prediction. By reformulating the input as a set of spatially distributed patches with coarse geometric priors, our model effectively combines local detail preservation with global structural reasoning. The proposed GridMix patch sampling strategy further enhances spatial robustness and inter-patch consistency. Extensive experiments demonstrate clear advantages over prior high-resolution refinement approaches, achieving sharper boundaries, smoother transitions, and improved geometric coherence. Our results highlight the potential of global-context transformers for scalable, high-fidelity 3D geometry estimation across arbitrary resolutions.

\section*{Acknowledgment}
The research reported in this publication was supported by funding from King Abdullah University of Science and Technology (KAUST) – Center of Excellence for Generative AI, under award number 5940 and a gift from Google.

{
    \small
    \bibliographystyle{ieeenat_fullname}
    \bibliography{main}
}

\clearpage

% Column 1: 
\newcommand{\zoomA}[1]{%
  \begin{overpic}[width=0.30\textwidth]{#1}%
    \put(65,20){%
      \makebox[0pt]{%
        \adjincludegraphics[
          height=0.10\textwidth,
          trim={{.40\width} {.30\height} {.40\width} {.40\height}},
          clip,
          cfbox=red 1pt 0cm
        ]{#1}%
      }%
    }%
  \end{overpic}%
}

% Column 2: 
\newcommand{\zoomB}[1]{%
  \begin{overpic}[width=0.30\textwidth]{#1}%
    \put(35,20){%
      \makebox[0pt]{%
        \adjincludegraphics[
          height=0.10\textwidth,
          trim={{.15\width} {.35\height} {.65\width} {.35\height}},
          clip,
          cfbox=red 1pt 0cm
        ]{#1}%
      }%
    }%
  \end{overpic}%
}

% Column 3: 
\newcommand{\zoomC}[1]{%
  \begin{overpic}[width=0.30\textwidth]{#1}%
    \put(35,20){%
      \makebox[0pt]{%
        \adjincludegraphics[
          height=0.10\textwidth,
          trim={{.15\width} {.35\height} {.55\width} {.35\height}},
          clip,
          cfbox=red 1pt 0cm
        ]{#1}%
      }%
    }%
  \end{overpic}%
}

\setcounter{page}{1}
\maketitlesupplementary
\appendix

In this supplementary material, we provide additional implementation and dataset details (including training hyper-parameters, evaluation protocols, and metric definitions), extended ablation studies on data augmentation and GridMix sampling, more qualitative visualizations at multiple resolutions (2K--8K) and out-of-domain images, and a discussion of current limitations and directions for future work. We also refer readers to our website (\url{https://dreamaker-mrc.github.io/Any-Resolution-Any-Geometry}), which provides an interactive comparison of our predictions.

\section{Implementation Details}
\label{supp:implement}

\subsection{Training Details}
We adopt Depth-Anything V2 (DA2)~\cite{yang2025depthanythingv2} and Metric3D V2~\cite{hu2025metric3dv2} as the coarse prediction backbones. 
Training is conducted on 7{,}592 samples from the UnrealStereo4K~\cite{u4k} dataset for 80K iterations using 4 NVIDIA A100 GPUs with a batch size of 1. 
We use the AdamW optimizer with a learning rate of $1\times10^{-5}$, weight decay of $1\times10^{-6}$, and apply gradient norm clipping with a maximum $\ell_2$ norm of 35 to stabilize training; gradient checkpointing is enabled to reduce memory consumption.

During training, we operate on fixed-size patches with a resolution of $540 \times 960$.
Our multi-task objective combines a depth loss $\mathcal{L}_{\text{depth}}$ with a normal loss $\mathcal{L}_{\text{normal}}$, and the final loss is given by
\begin{equation}
  \begin{aligned}
    \Loss{total}
    &= \lambda_{\text{depth}}\,\Loss{depth}\big(\DMaps^{refined}, \DMaps^{gt}\big) \\
    &\quad + \lambda_{\text{normal}}\,\Loss{normal}\big(\SNormal^{refined}, \SNormal^{pseudo}\big),
  \end{aligned}
\end{equation}
where we set $\lambda_{\text{depth}} = 1$, $\lambda_{\text{normal}} = 0.01$ to balance the contribution of the normal supervision. 

\subsection{Datasets}
\noindent \textbf{UnrealStereo4K.} The UnrealStereo4K dataset~\cite{u4k} provides stereo image pairs at 4K resolution (2160$\times$3840), each with dense and boundary-preserving ground-truth annotations. 
All scenes are rendered in Unreal Engine using the UnrealCV plugin across eight virtual environments, offering diverse geometry, materials, and lighting conditions. 
Following~\cite{li2024patchfusion, li2024patchrefiner}, we first remove mislabeled samples using the Structural Similarity Index (SSIM)~\cite{SSIM}. Following PRO~\cite{kwon2025onelook}, we employ the same Bias-Free Mask during training to ensure a fair comparison.
The final depth ground truth is then computed from the provided disparity maps using the calibrated camera baseline and focal length.

\noindent \textbf{Middlebury.} The Middlebury 2014 dataset~\cite{middlebury2014} provides high-resolution indoor scenes with accurate ground-truth disparity and depth annotations. 
Following common practice, we select 23 stereo pairs with valid ground truth and convert the provided disparity maps into depth maps using the calibrated camera parameters.

\noindent \textbf{Booster.} The Booster dataset~\cite{zamaramirez2022booster} contains high-resolution indoor images (3008$\times$4112) featuring challenging specular, reflective, and transparent surfaces. We use the whole training set with GT for evaluation, resulting in 228 images.

\noindent \textbf{ETH3D.} The ETH3D high-resolution dataset~\cite{ETH3D} includes both indoor and outdoor scenes (6048$\times$4032) with accurate ground-truth depth maps captured using LiDAR sensors.

\subsection{Evaluation Details}
\paragraph{PatchRefiner.} We retrained PatchRefiner~\cite{li2024patchrefiner} on top of DA2 for a fair comparison. We evaluate under two configurations: (i) $p = 16$, which uses the same number of patches as our method, and (ii) $p = 49$, where additional patches are employed for test-time ensembling. Here, $p$ denotes the number of patches used to reassemble the final depth map during inference.

\paragraph{Consistency Error (CE).}When computing the consistency error (CE) between neighboring patches, we evaluate the discrepancy only within a $270$-pixel-wide overlapping region at the patch boundaries.

\paragraph{Pseudo Depth Boundary Error (PDBE).}
Following SharpDepth~\cite{pham2024sharpdepthsharpeningmetricdepth}, we evaluate depth boundary quality using the Pseudo Depth Boundary Error, decomposed into an accuracy term $\epsilon_{\mathrm{PDBE}}^{\mathrm{acc}}$ and a completeness term $\epsilon_{\mathrm{PDBE}}^{\mathrm{compl}}$. 
Given a predicted depth map $D^{refined}$ and ground-truth depth $D^{gt}$, we first normalize each map to $[0,1]$ and apply a Canny edge detector to obtain depth edges $E^{refined}_{\text{depth}}$ and $E^{gt}_{\text{depth}}$. 
In addition, we convert depth to disparity, normalize them, and run Canny again to extract disparity edges $E^{refined}_{\text{disp}}$ and $E^{gt}_{\text{disp}}$. 
The final ground-truth and predicted edge maps are then defined as the union of depth and disparity edges, i.e., $E^{refined} = E^{refined}_{\text{depth}} \lor E^{refined}_{\text{disp}}$ and $E^{gt} = E^{gt}_{\text{depth}} \lor E^{gt}_{\text{disp}}$.

We compute Euclidean distance transforms $T^{refined}$ and $T^{gt}$ on the complements of $E^{refined}$ and $E^{gt}$, respectively, truncated to a local neighborhood of 10 pixels. 
The PDBE accuracy $\epsilon_{\mathrm{PDBE}}^{\mathrm{acc}}$ measures how close each predicted edge is to the nearest ground-truth edge:
\begin{equation}
\epsilon_{\mathrm{PDBE}}^{\mathrm{acc}} = 
\frac{\sum_{x} T^{refined}(x)\, E^{gt}(x)}{\sum_{x} E^{gt}(x)} ,
\end{equation}
while the PDBE completeness $\epsilon_{\mathrm{PDBE}}^{\mathrm{compl}}$ quantifies how well ground-truth edges are recovered by the prediction:
\begin{equation}
\epsilon_{\mathrm{PDBE}}^{\mathrm{compl}} = 
\frac{\sum_{x} T^{gt}(x)\, E^{refined}(x)}{\sum_{x} E^{refined}(x)} .
\end{equation}
In both cases, lower values indicate sharper and better aligned depth boundaries.

\section{More Ablation Study}
\label{supp:ablation}
To empirically determine the optimal configuration for our proposed GridMix, we conduct a series of experiments focusing on different patch sampling probabilities $(p_1, p_2, p_3, p_4)$. As shown in Table~\ref{tab:training_grid_more}, the configuration $(0.1, 0.2, 0.3, 0.4)$ consistently outperforms other variants across all evaluated metrics. Specifically, it achieves the best performance not only in depth accuracy---indicated by AbsRel~$\downarrow$, $\delta_1$~$\uparrow$, and RMSE~$\downarrow$---but also in the consistency metric CE~$\downarrow$. These results indicate that probabilistic grid sampling effectively enhances inter-patch coherence while preserving fine geometric detail, leading to a more robust representation compared to the Depth-Anything v2 baseline.
\begin{table}[t]
  \centering
  \small
  \caption{Ablation on GridMix Patch Sampling Strategy.}
  \label{tab:training_grid_more}
  \resizebox{\columnwidth}{!}{
  \begin{tabular}{l|cccc}
    \toprule
    \begin{tabular}{c}
    \textbf{Configurations} \\
    \textbf{($\vProb{1},\vProb{2},\vProb{3},\vProb{4}$)}
    \end{tabular}
    & AbsRel $\downarrow$ & $\delta_1$ $\uparrow$ & RMSE $\downarrow$ & CE $\downarrow$ \\
    \midrule
    Depth-Anything v2                 & 0.0812 & 0.924 & 2.86 & --      \\
    (1,0,0,0)             & 0.0500 & 0.963 & 2.01 & 0.0648 \\
    (0.5,0.5,0,0)         & 0.0473 & 0.966 & 1.79 & 0.0436 \\
    (0,1,0,0)         & 0.0405 & 0.973 & 1.72 & 0.0447 \\
    (0.5,0,0.5,0)         & 0.0365 & 0.975 & 1.58 & 0.0440 \\
    (0,0,1,0)         & 0.0350 & 0.976 & 1.57 & 0.0443 \\
    (0.4,0.3,0.3,0)         & 0.0343 & 0.977 & 1.52 & 0.0457 \\
    (0.1,0.2,0.3,0.4)     & \textbf{0.0295} & \textbf{0.982} & \textbf{1.35} & \textbf{0.0418} \\
    (0,0,0.5,0.5)             & 0.0311 & 0.981 & 1.38 & 0.0435 \\
    (0,0,0,1)             & 0.0321 & 0.980 & 1.42 & 0.0635 \\
    \bottomrule
  \end{tabular}
  }
\end{table}

\section{Extension to Any Resolution}
Our patch-based formulation naturally supports arbitrary input resolutions at test time. 
Given a high-resolution image, we keep both the patch size and the transformer backbone fixed, and only scale the patch grid to cover the full image domain. 
This enables our model to handle 2K, 4K, 8K, and even higher resolutions without any resolution-specific retraining, while preserving local fine details and maintaining global geometric consistency across all patches.

As shown in Fig.~\ref{fig:any_res}, we apply the same model to in-the-wild images at 2K, 4K, and 8K resolutions. 
Across these settings, our approach consistently sharpens thin structures, refines object boundaries, and produces smoother, more coherent normal fields compared to the coarse predictions from Depth-Anything V2 and Metric3D V2. 
In addition, Fig.~\ref{fig:8k_solo} illustrates an 8K manga-style image that lies far outside the training domain. 
Even under this highly stylized, out-of-domain scenario, the model still recovers geometrically plausible depth and normals, highlighting the strong generalization ability of our multi-patch transformer for high-resolution single-image geometry estimation.

\begin{figure}[h]
  \centering
  \setlength{\tabcolsep}{2pt}
  \renewcommand{\arraystretch}{1.0}
  \begin{tabular}{ccc}
    \small RGB &
    \small Depth (ours) &
    \small Normal (ours) \\
    \includegraphics[width=0.3\columnwidth]{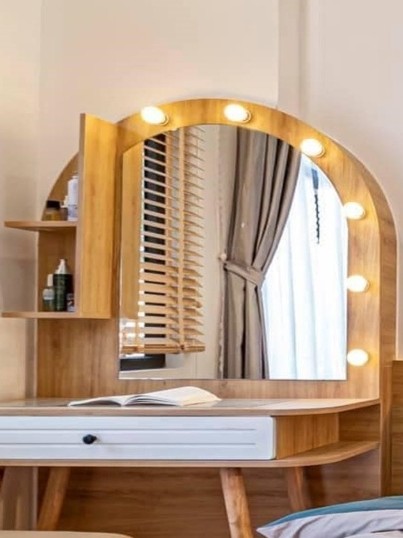} &
    \includegraphics[width=0.3\columnwidth]{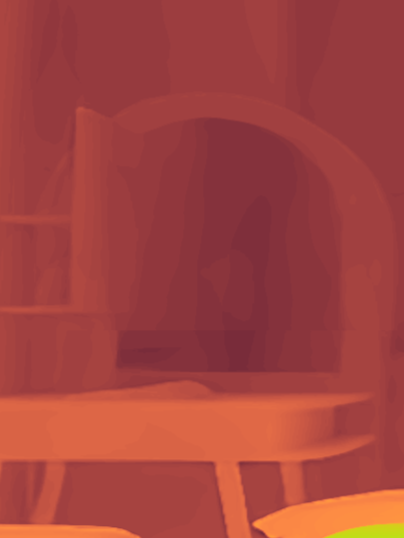} &
    \includegraphics[width=0.3\columnwidth]{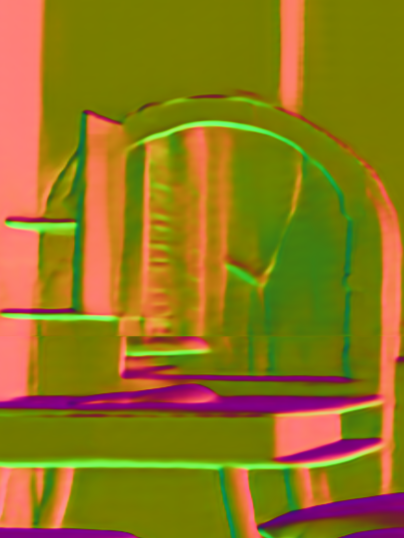} \\
  \end{tabular}
  \caption{
    \textbf{Failure case on reflective surfaces.}
    Our model struggles to handle strongly reflective objects such as mirrors:
    the mirror region is incorrectly interpreted as a continuation of the surrounding geometry.
  }
  \label{fig:limitation_mirror}
\end{figure}

\begin{figure*}[h]
  \centering
  \setlength{\tabcolsep}{1pt}
  \renewcommand{\arraystretch}{1.0}
  \begin{tabular}{l@{\hspace{2pt}}ccc}
    % ---------------- Row 1: RGB ----------------
    \rotatebox{90}{\hspace{2.5em} \small RGB} &
    \includegraphics[width=0.28\textwidth]{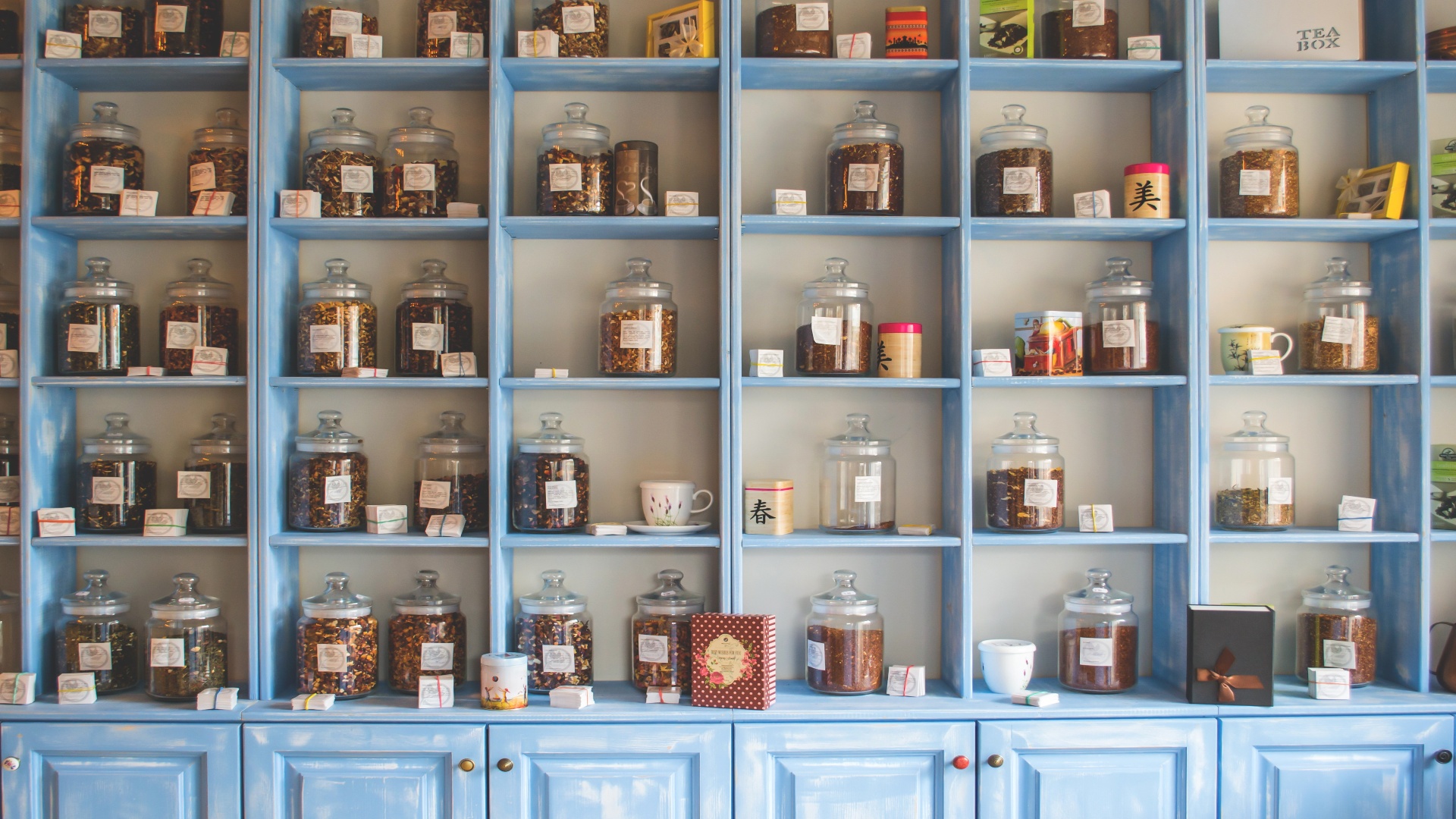} &
    \includegraphics[width=0.28\textwidth]{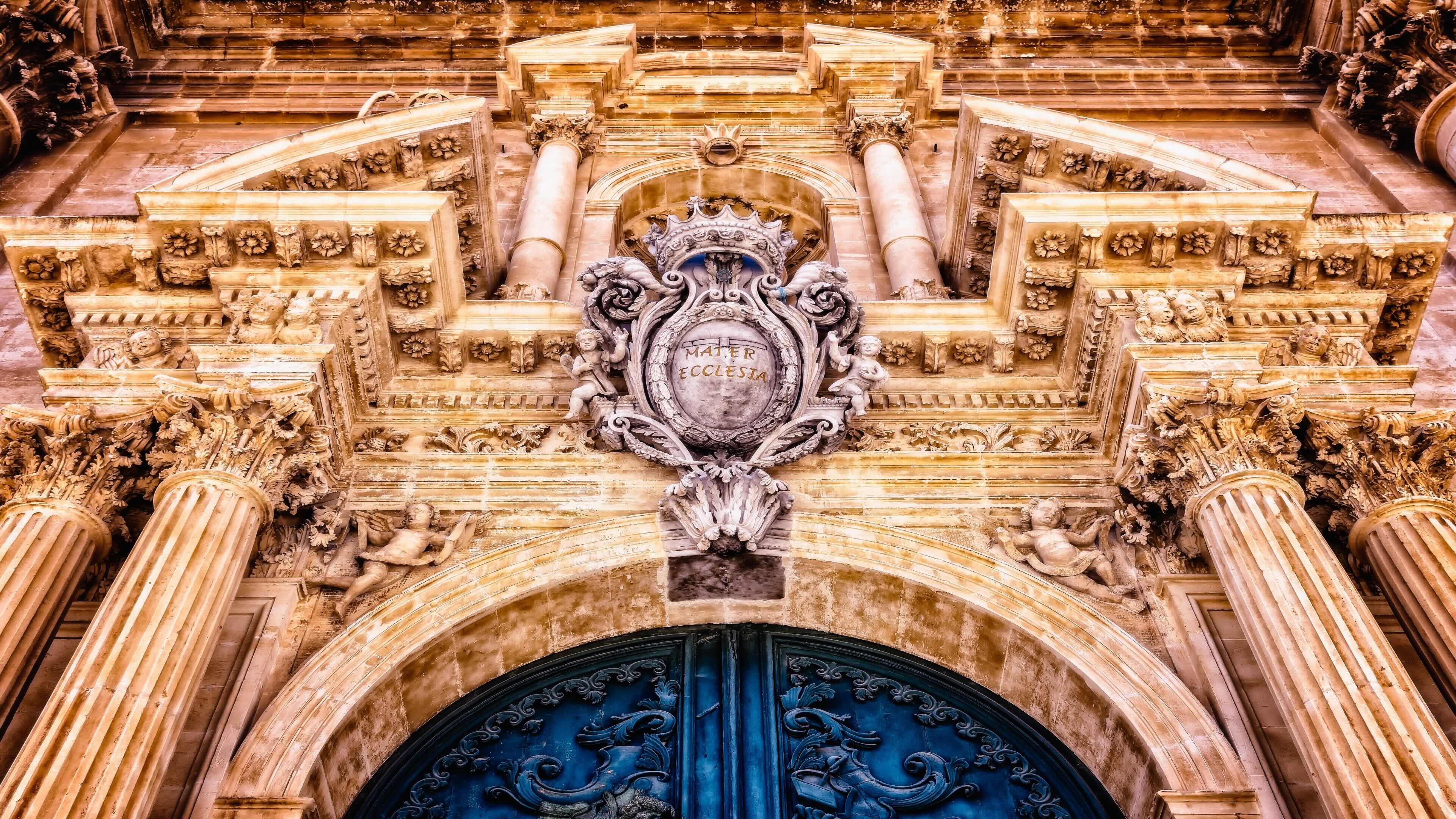} &
    \includegraphics[width=0.28\textwidth]{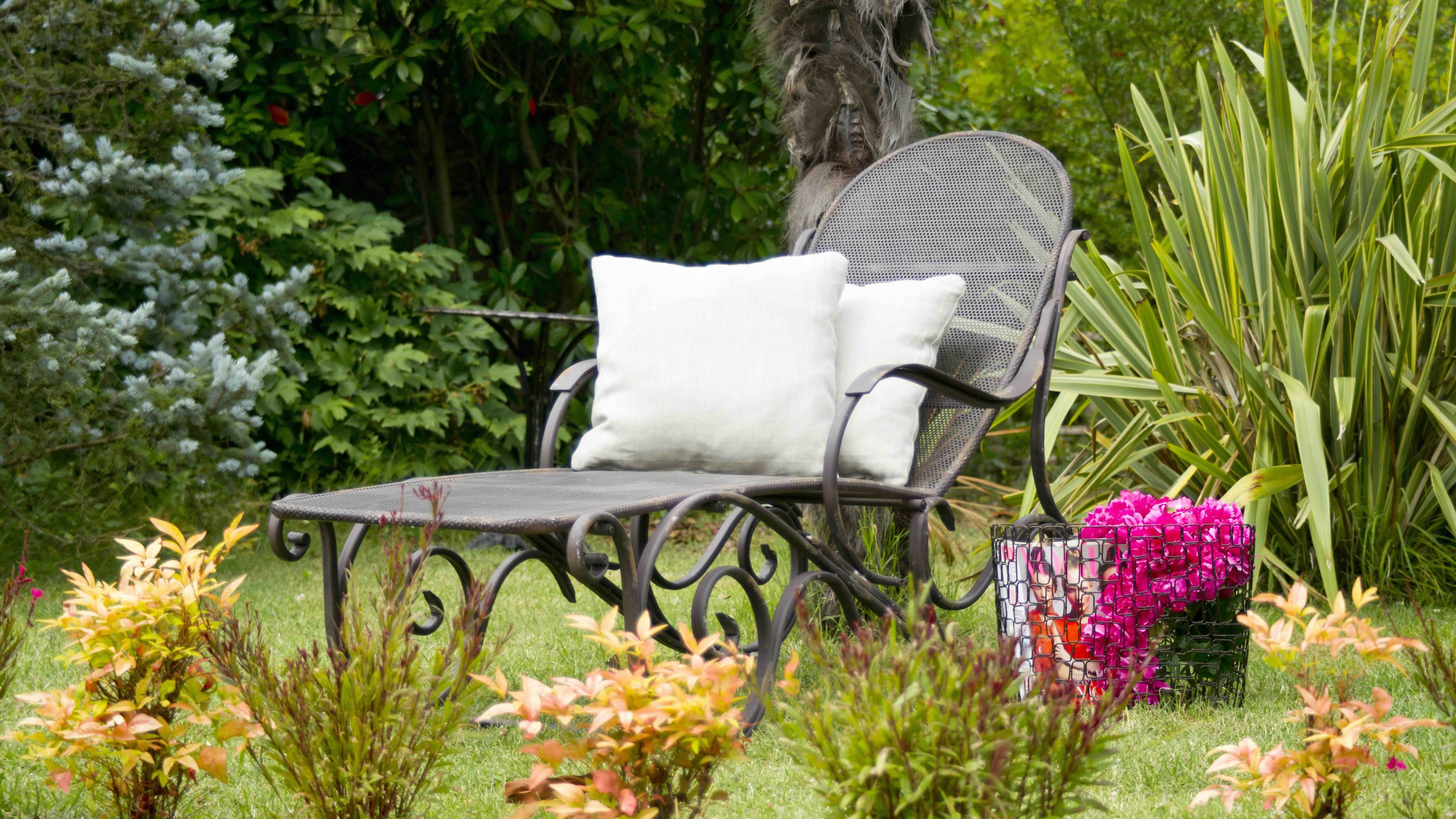} \\[-2pt]
    % ---------------- Row 2: Coarse Depth ----------------
    \rotatebox{90}{\small Depth-Anything V2} &
    \includegraphics[width=0.28\textwidth]{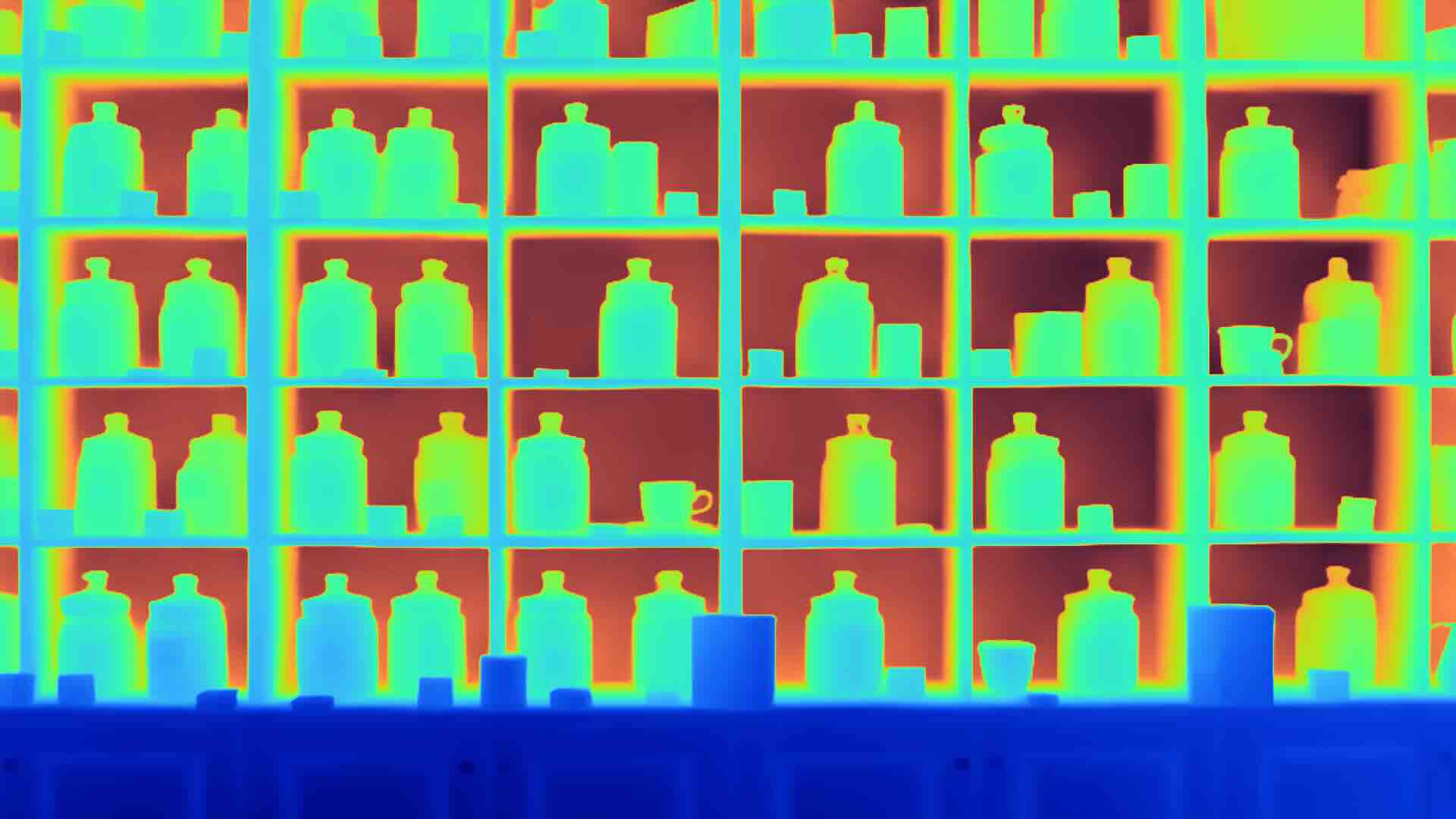} &
    \includegraphics[width=0.28\textwidth]{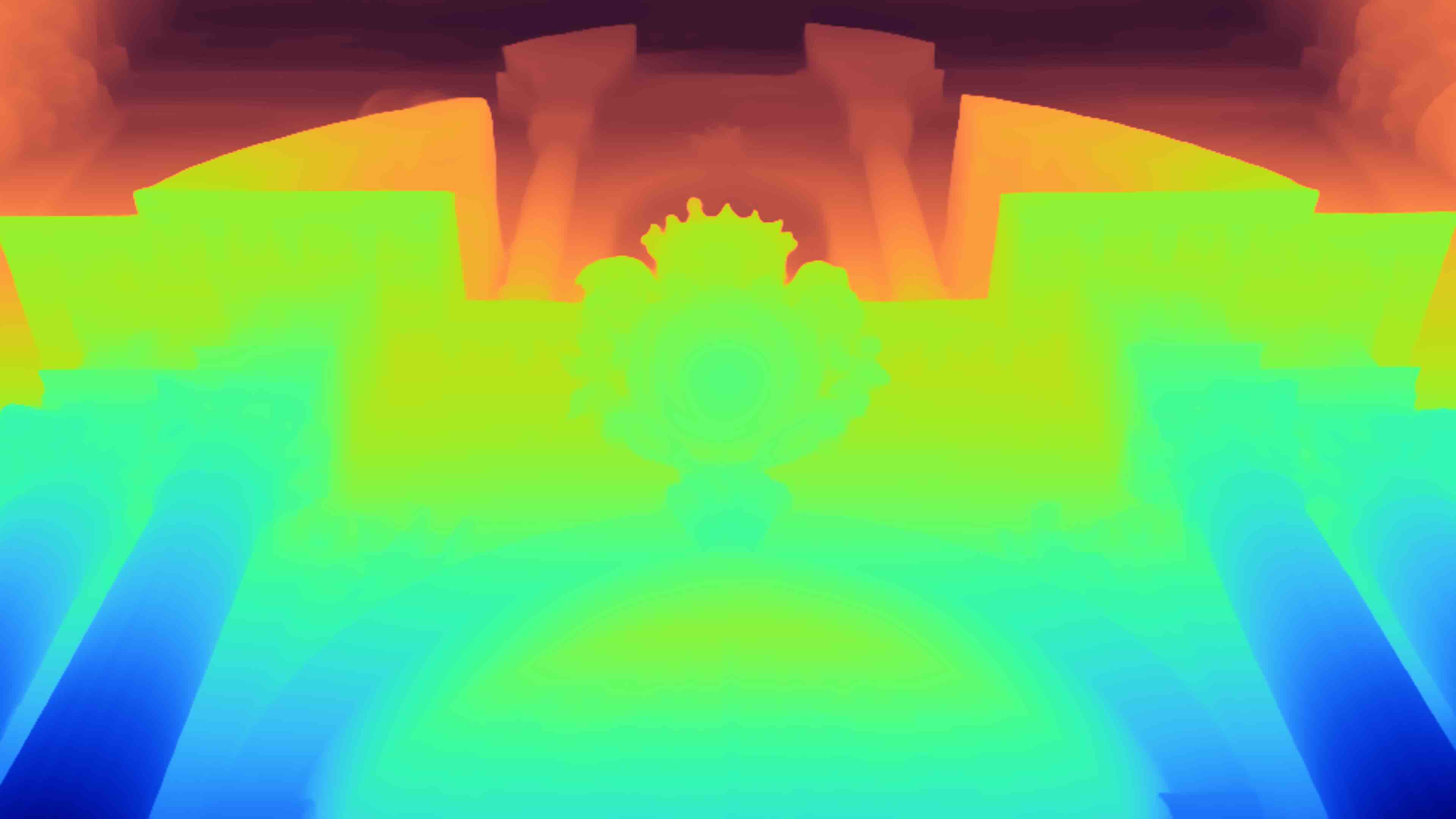} &
    \includegraphics[width=0.28\textwidth]{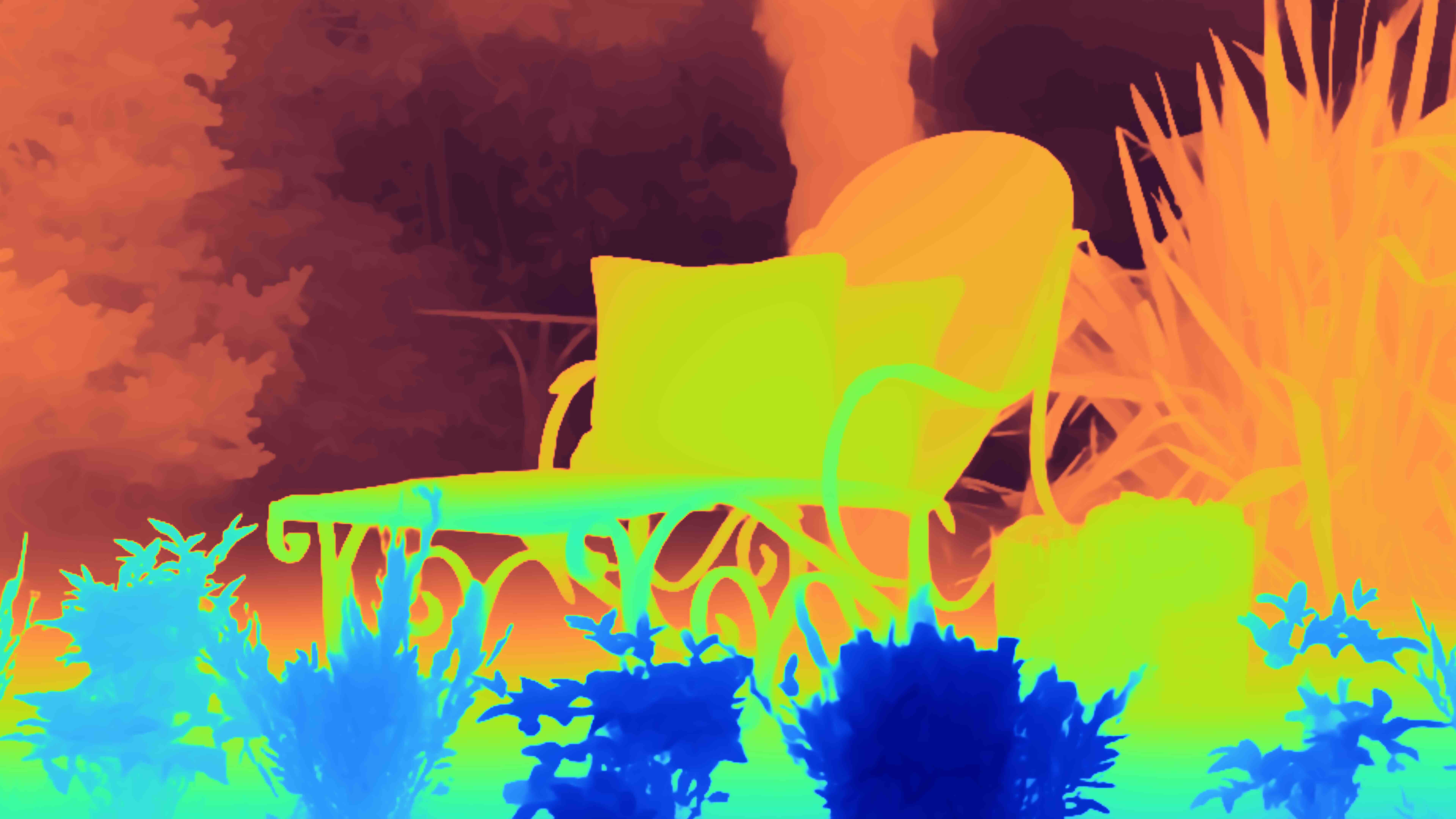} \\[-2pt]
    % ---------------- Row 3: Refined Depth ----------------
    \rotatebox{90}{\hspace{2em}\small Our Depth} &
    \includegraphics[width=0.28\textwidth]{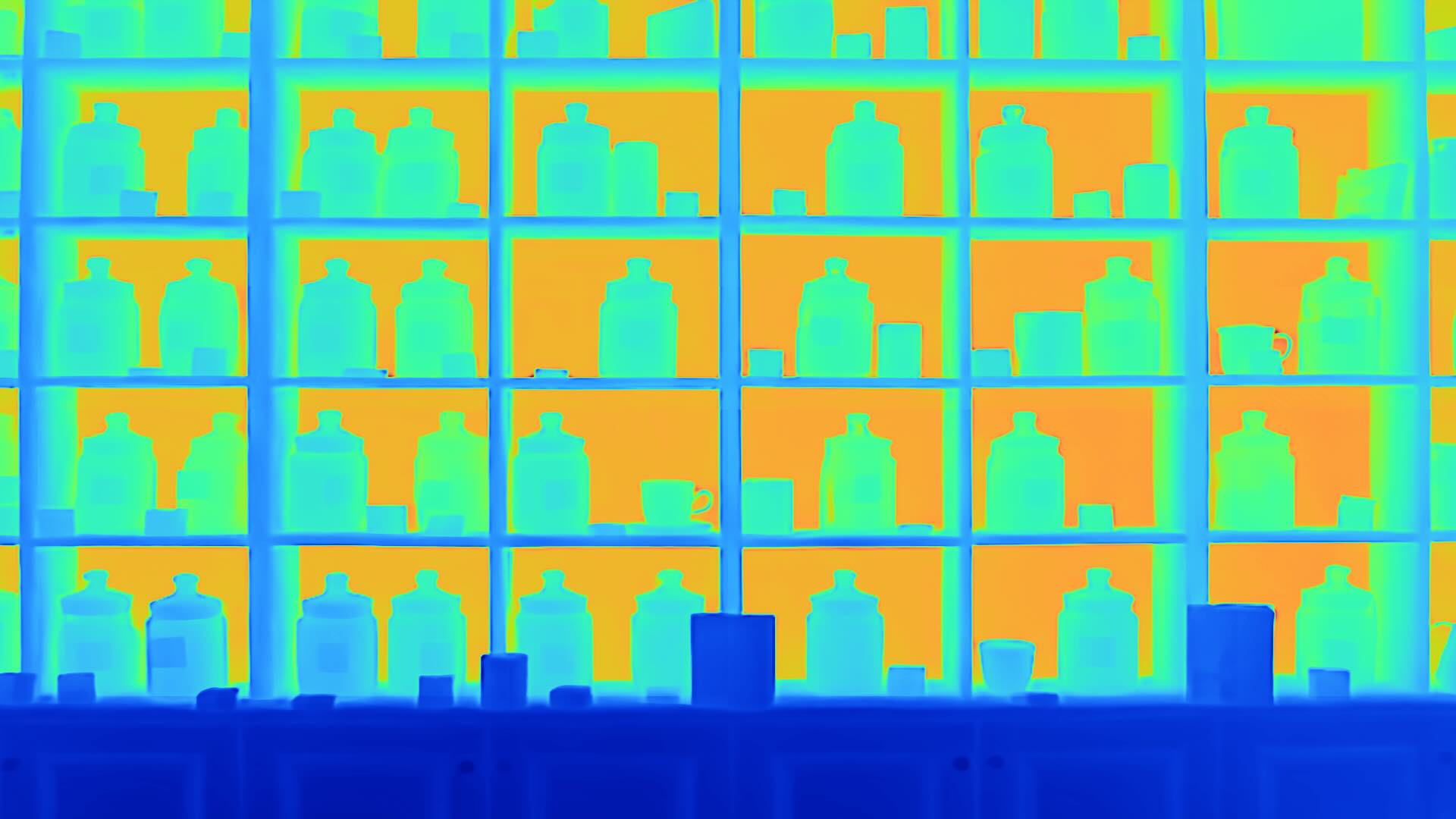} &
    \includegraphics[width=0.28\textwidth]{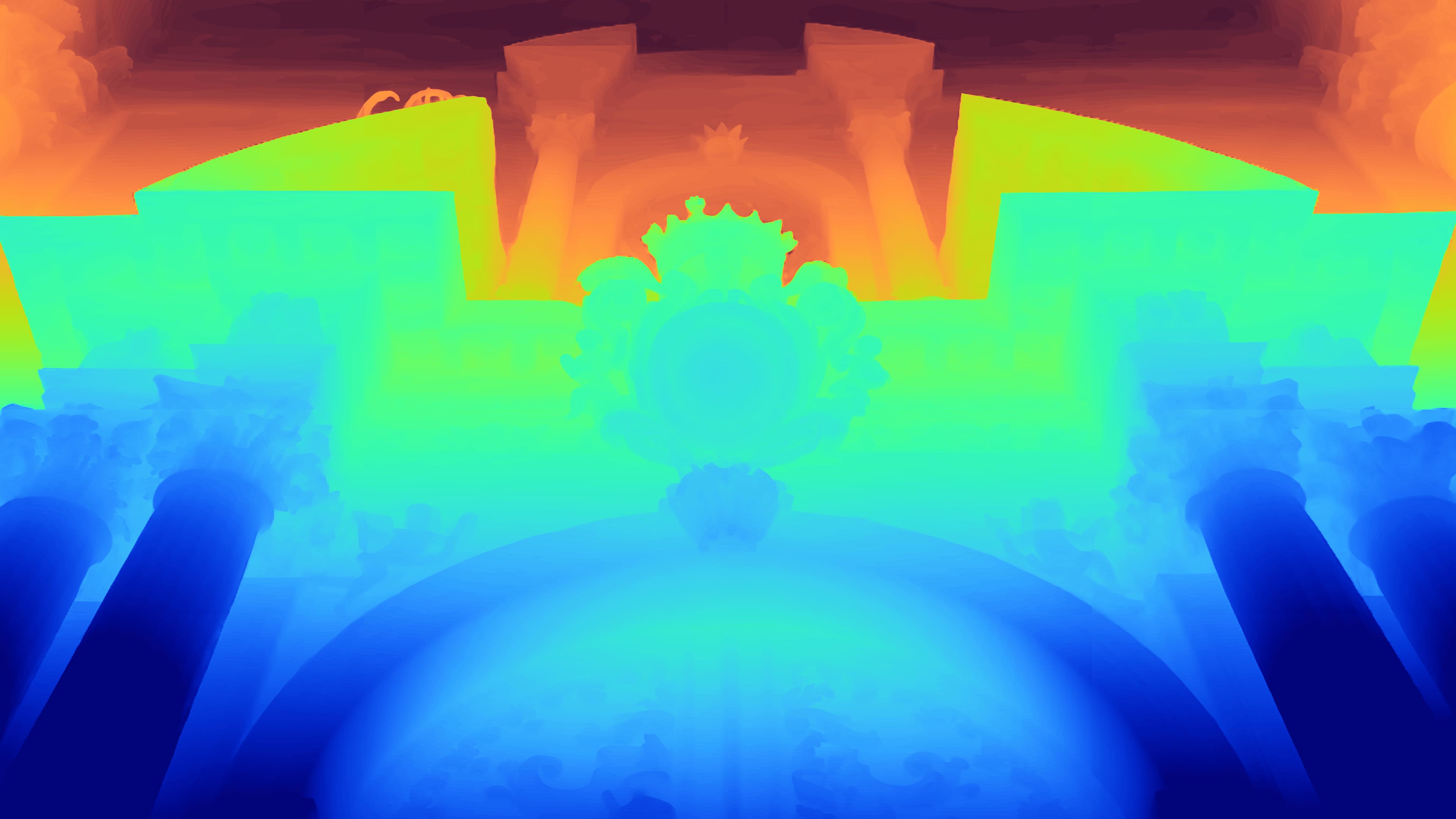} &
    \includegraphics[width=0.28\textwidth]{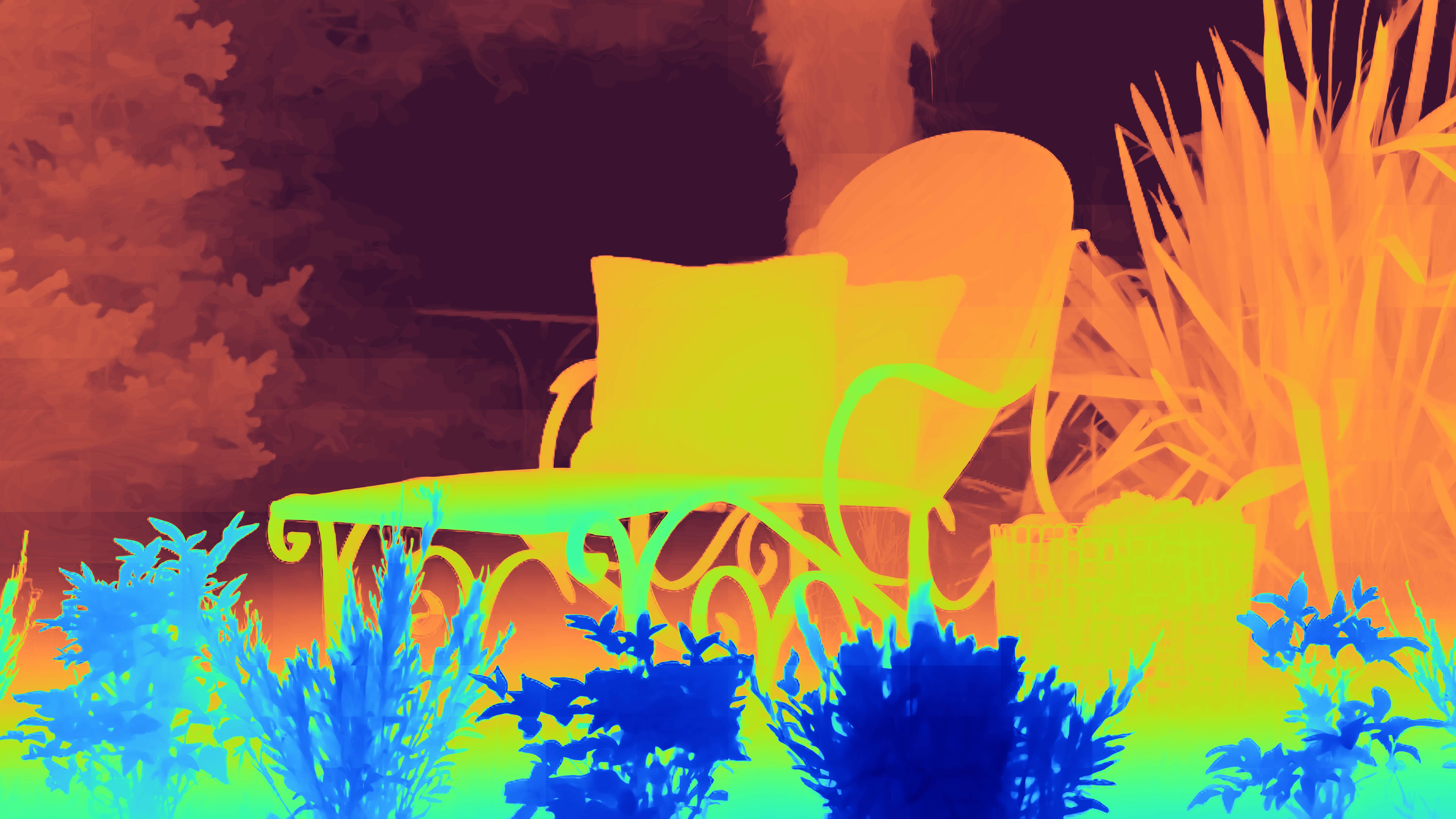} \\[-2pt]
    % ---------------- Row 4: Coarse Normal ----------------
    \rotatebox{90}{\hspace{1.7em}\small Metric3D V2} &
    \includegraphics[width=0.28\textwidth]{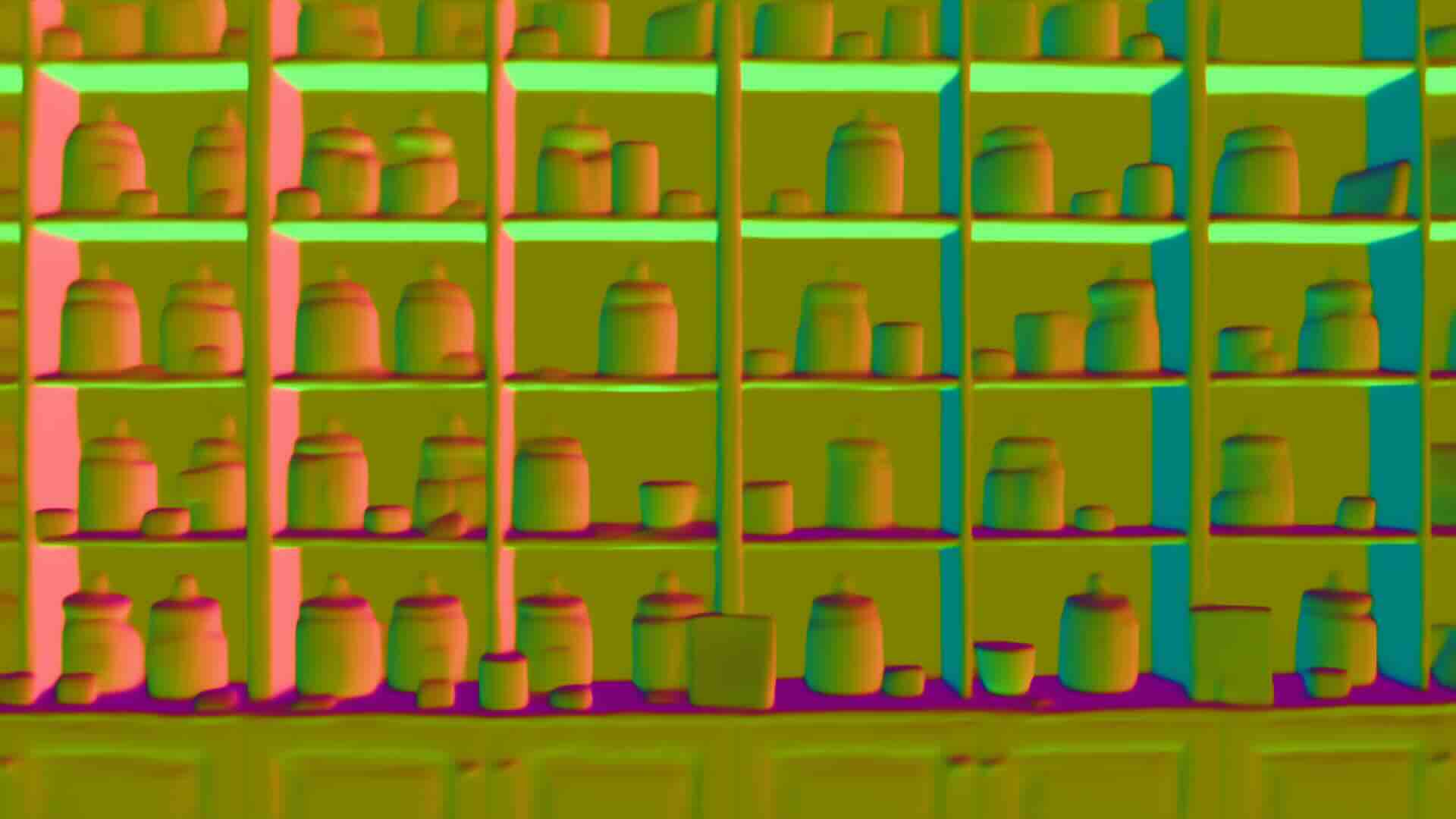} &
    \includegraphics[width=0.28\textwidth]{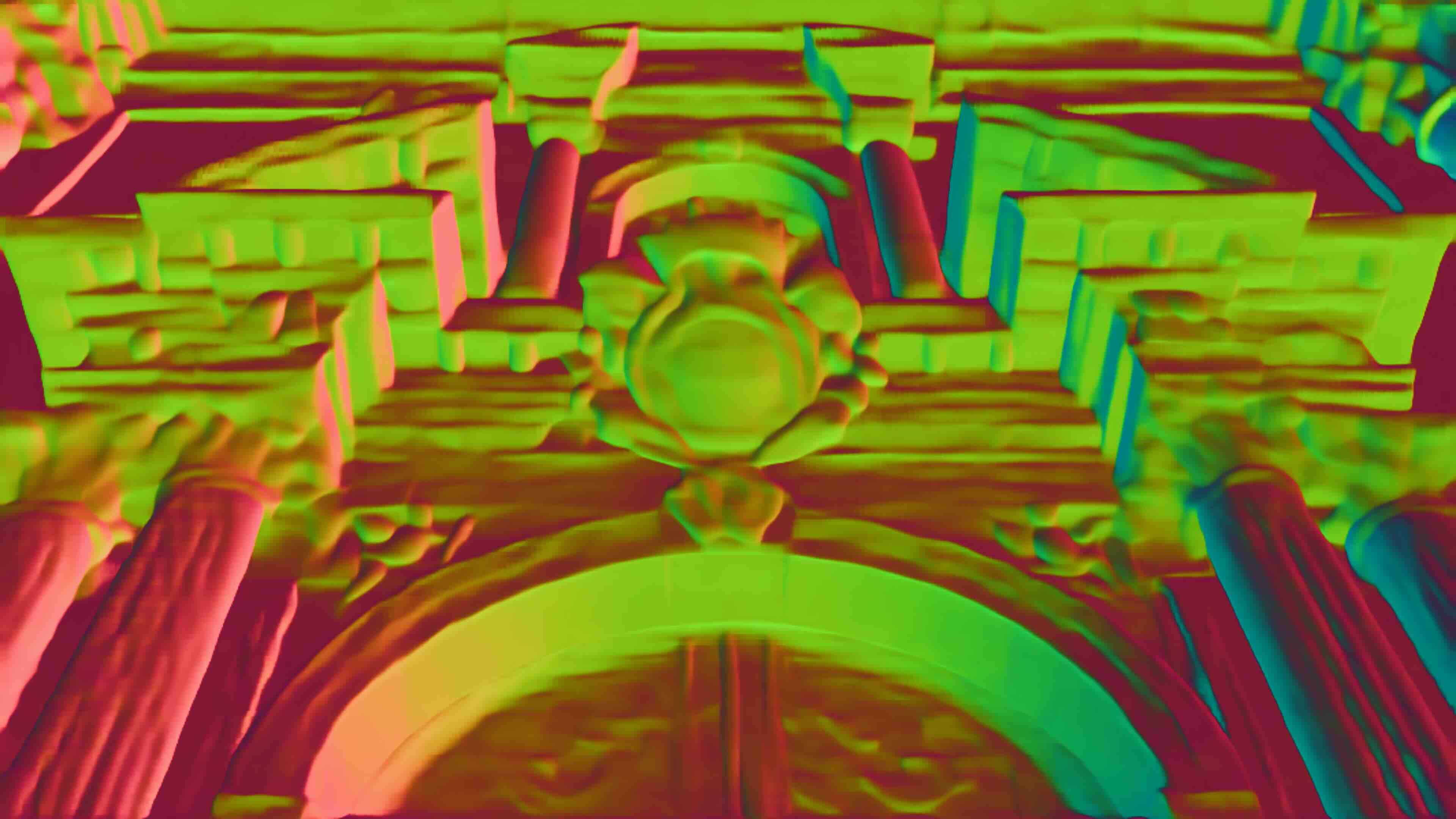} &
    \includegraphics[width=0.28\textwidth]{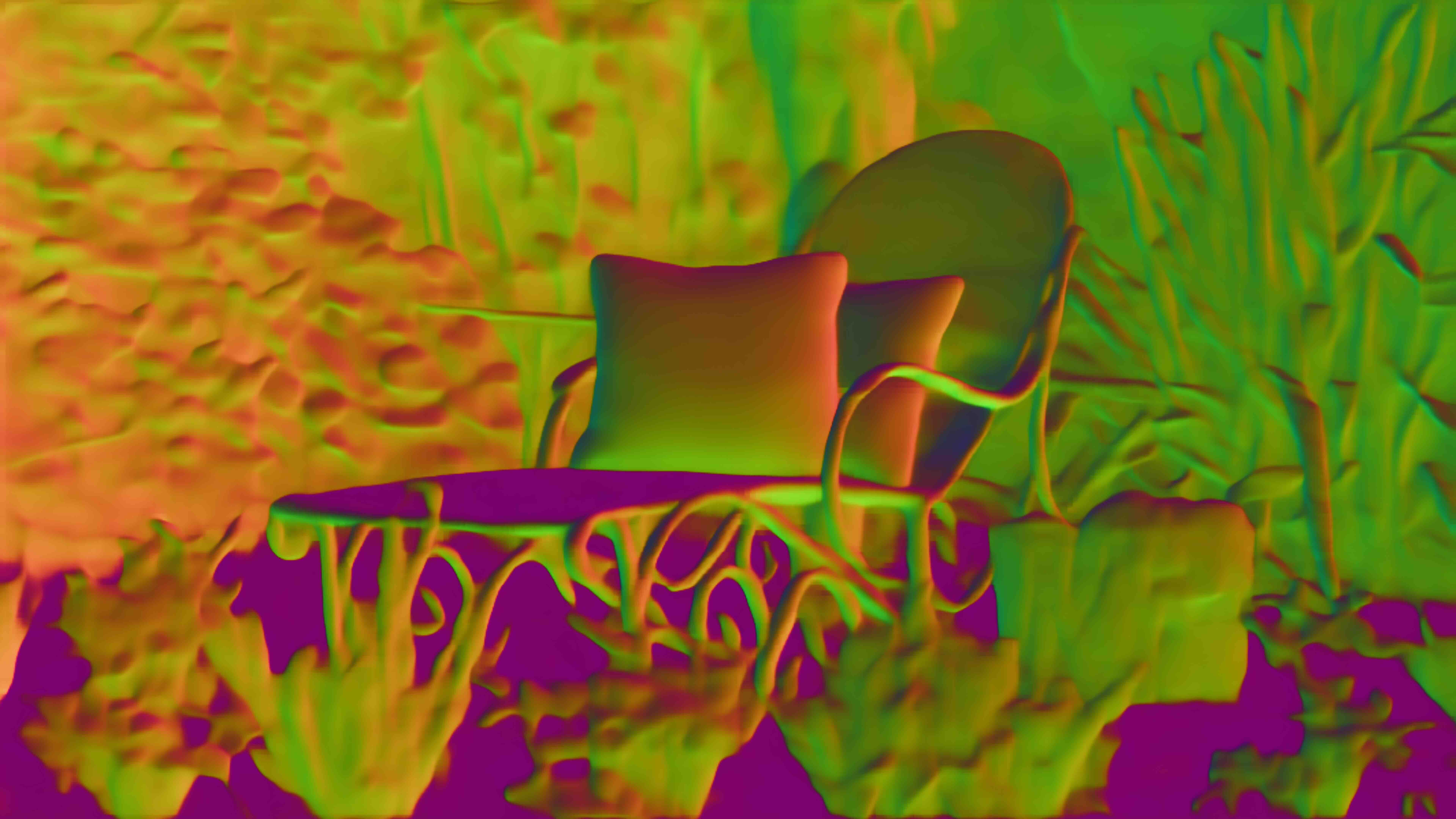} \\[-2pt]
    % ---------------- Row 5: Refined Normal ----------------
    \rotatebox{90}{\hspace{1.7em} \small Our Normal} &
    \includegraphics[width=0.28\textwidth]{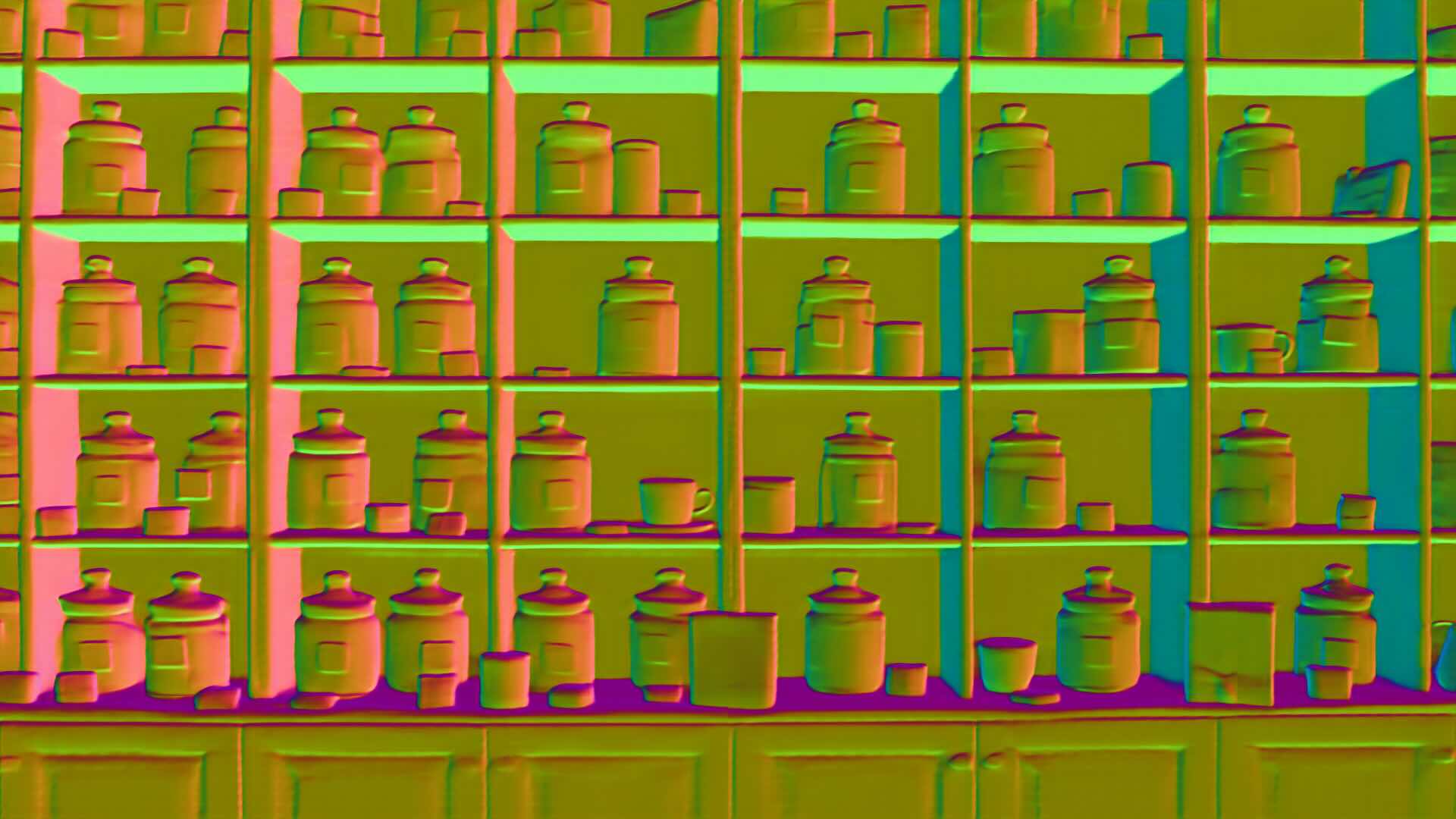} &
    \includegraphics[width=0.28\textwidth]{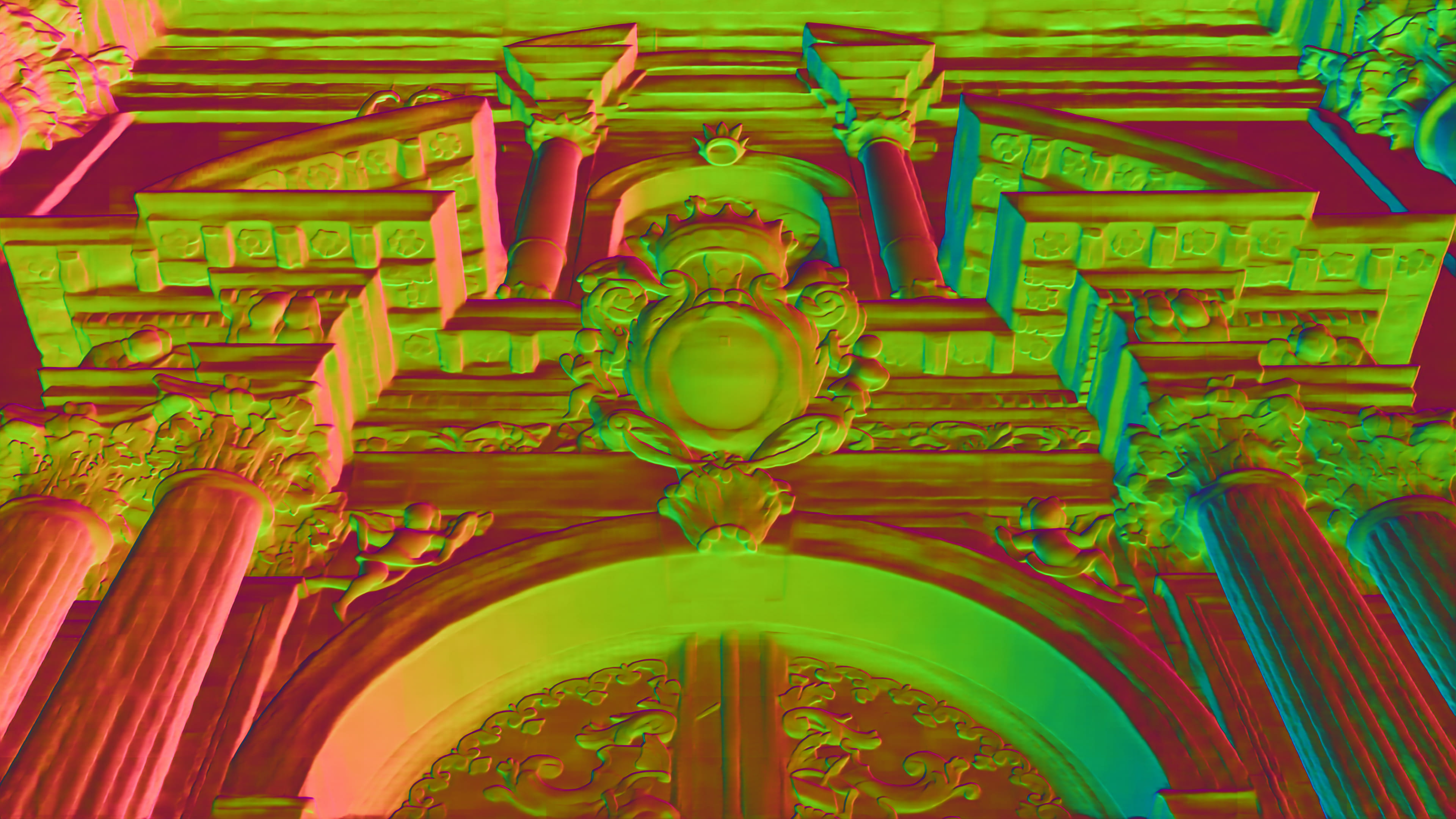} &
    \includegraphics[width=0.28\textwidth]{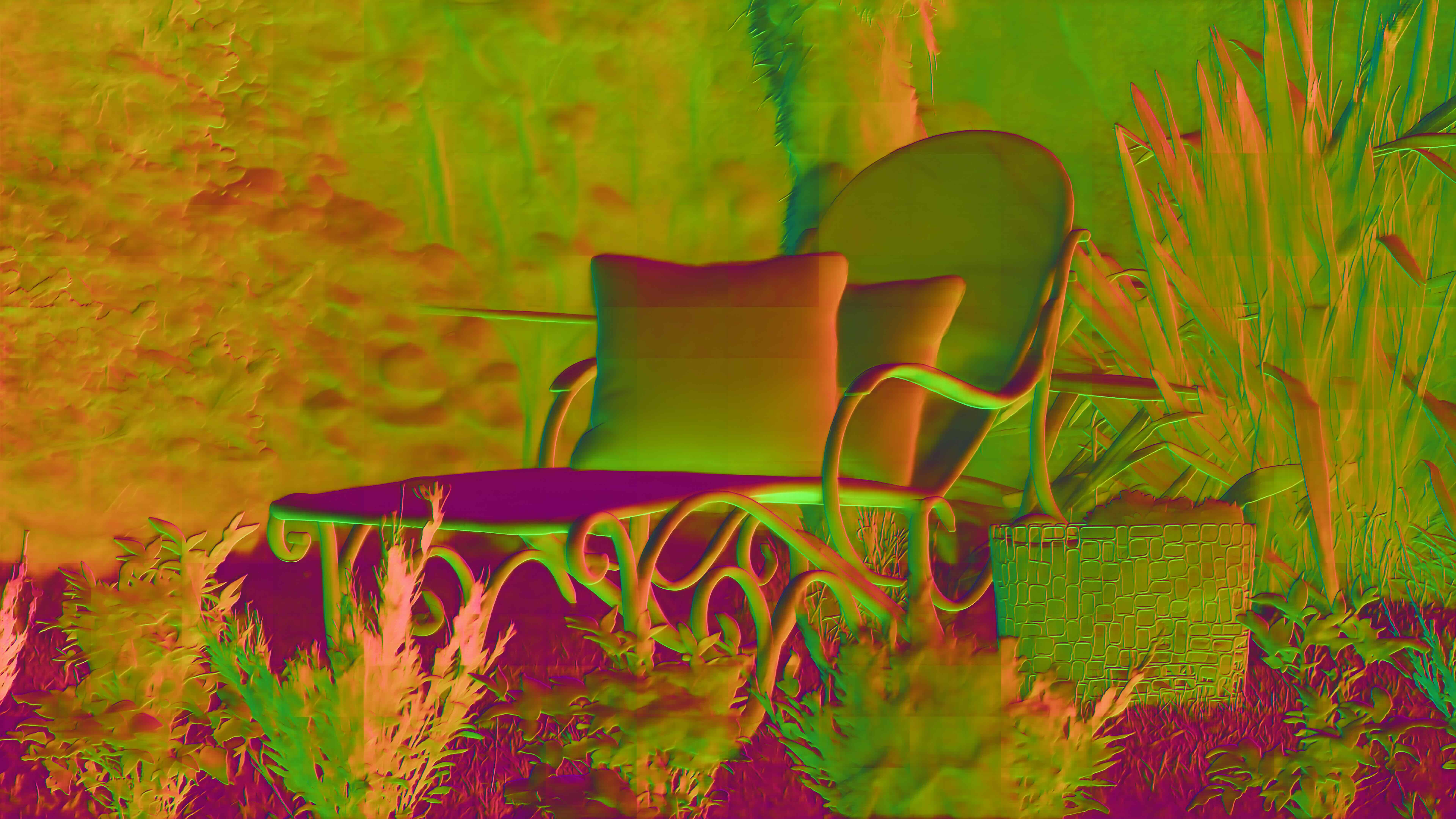} \\[3pt]
    % ---------------- Column labels ----------------
    & \small 2K & \small 4K & \small 8K \\
  \end{tabular}
  \caption{
  \textbf{Extension to arbitrary resolutions.}
  Each column shows a different input resolution (2K, 4K, 8K), and each row corresponds to the RGB input, coarse depth prediction by Depth-Anything V2, our refined depth prediction, coarse normal prediction by Metric3D V2, and our refined normal prediction, respectively.
  }
  \label{fig:any_res}
\end{figure*}

\section{More Qualitative Results}

We further provide qualitative results on the UnrealStereo4K~\cite{u4k} dataset. 
Figure~\ref{fig:u4k_qual_10x5} shows three representative scenes, where each column corresponds to one sample and the rows visualize the RGB input, coarse depth from Depth-Anything V2~\cite{yang2025depthanythingv2}, depth refinements from PatchRefiner~\cite{li2024patchrefiner} and PRO~\cite{kwon2025onelook}, our refined depth, coarse normals from Metric3D V2~\cite{hu2025metric3dv2}, and our refined normals. 
Across diverse scenes and lighting conditions, our method consistently sharpens depth, cleans up noisy regions, and recovers fine geometric structures such as thin objects, furniture details, and small decorations.
On the normal estimation side, our model produces smoother and more coherent normal fields and better fine-details.

\subsection{Improvement over Base Models}
\begin{figure*}[h]
  \centering
  \begin{tabular}{cc}
    % -------- The Two Images --------
    \includegraphics[width=0.48\textwidth]{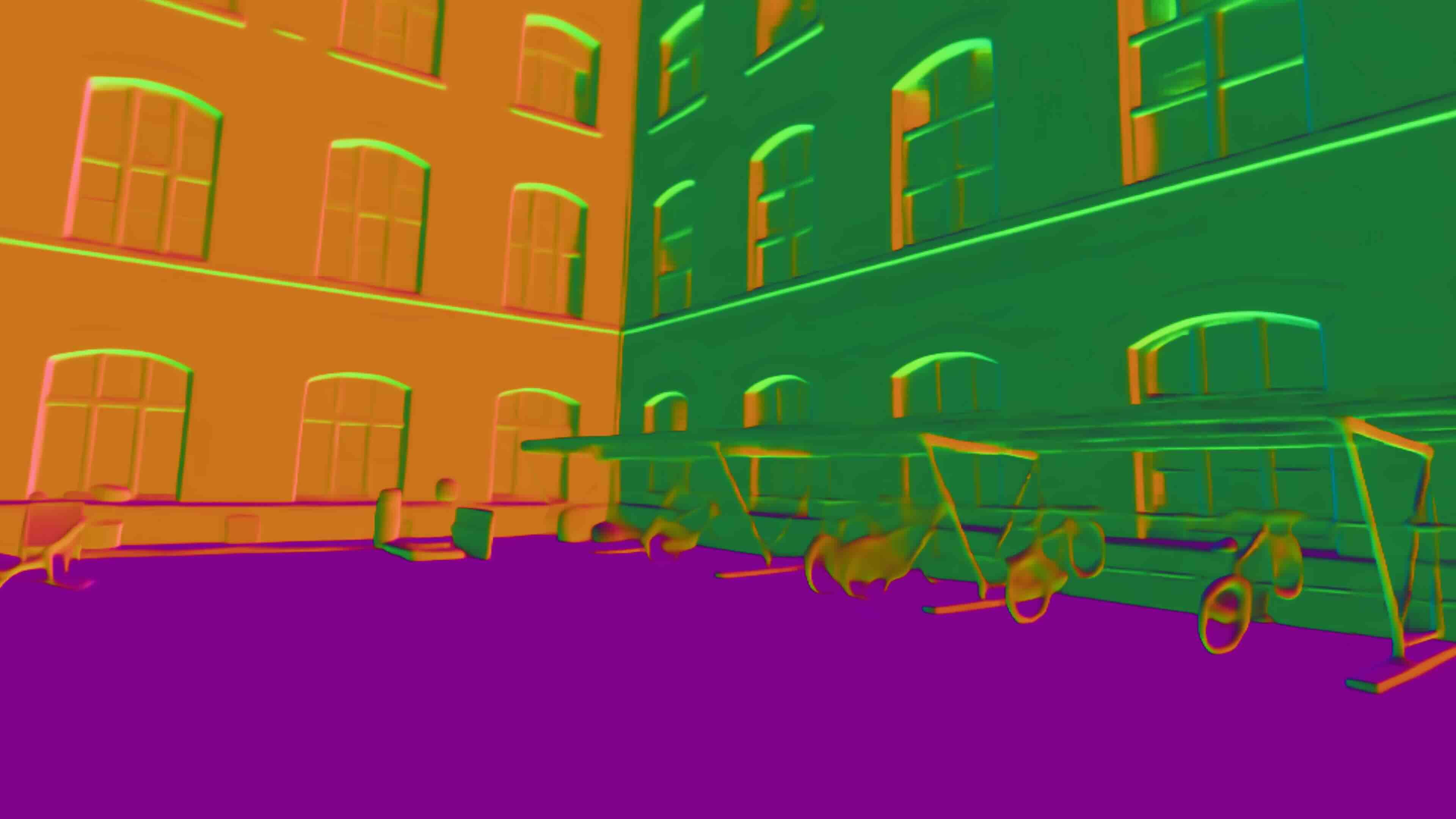} &
    \includegraphics[width=0.48\textwidth]{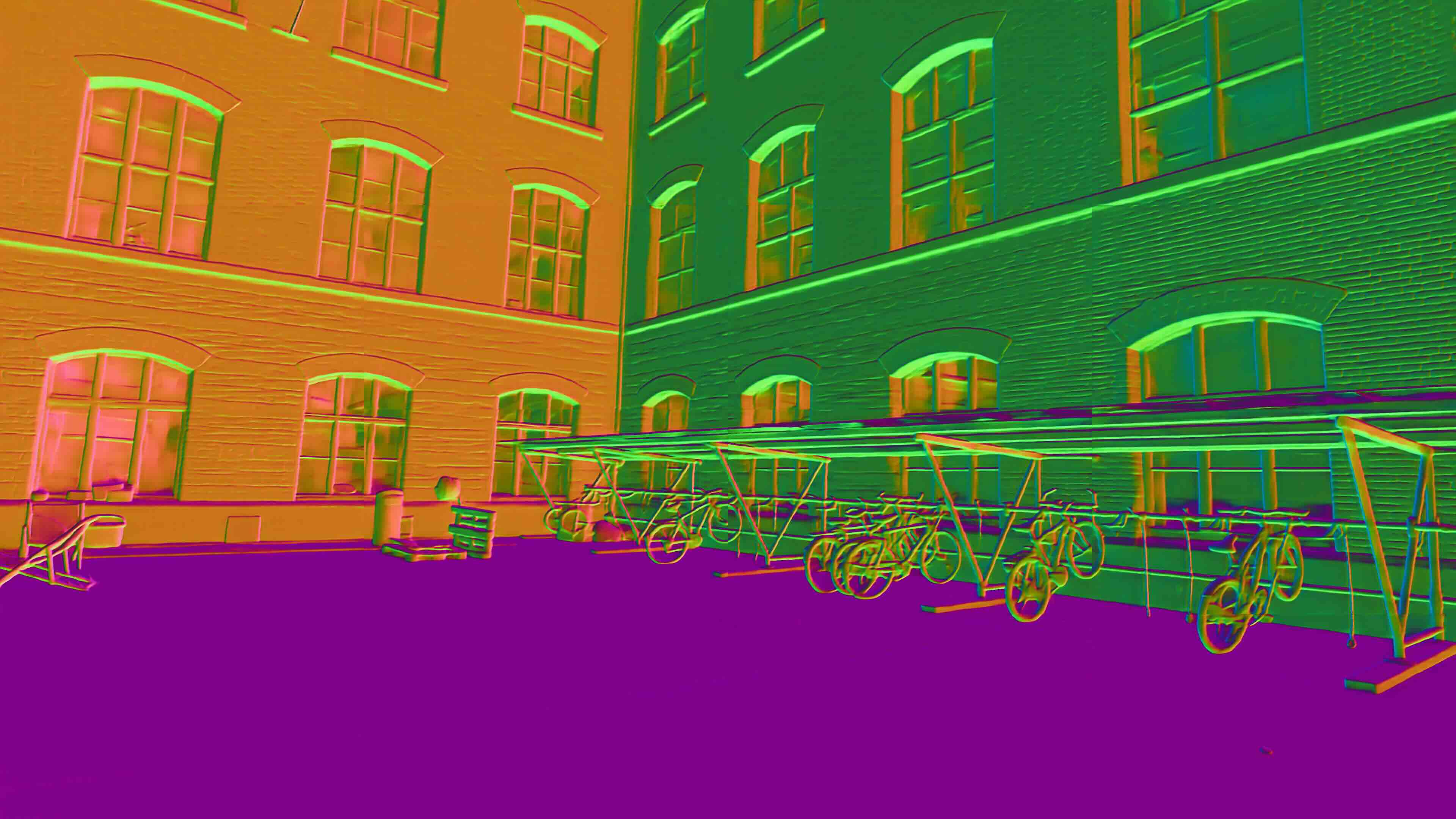} \\
    
    % -------- Sub-captions --------
    {(a) Metric3d v2} &
    {(b) Ours}
  \end{tabular}
  \caption{
    {\textbf{Surface Normal estimation example from ETH3D.}}
  }
  \label{fig:eth_normal}
\end{figure*}

While base models provide an accurate global scale, they are trained on low resolutions and lack the high-frequency details necessary to be good initial models for fine geometry. The marginal numerical gain stems from metrics like AbsRel/$\delta_1$ being dominated by global scale and failing to reflect local precision. Furthermore, since real-world Ground Truth is highly sparse, edge-quality metrics cannot be applied to quantify our gains. Our method provides a critical geometric correction of over-smoothed boundaries (see Fig.~\ref{fig:eth_normal}) that global pixel-wise averages fail to capture. In general, qualitative comparisons better highlight the key gains of our method.

\subsection{Applications: Lens Blur}
\begin{figure*}[h]
  \centering
  \begin{tabular}{cc}
    % -------- The Two Images --------
    \includegraphics[width=0.48\textwidth]{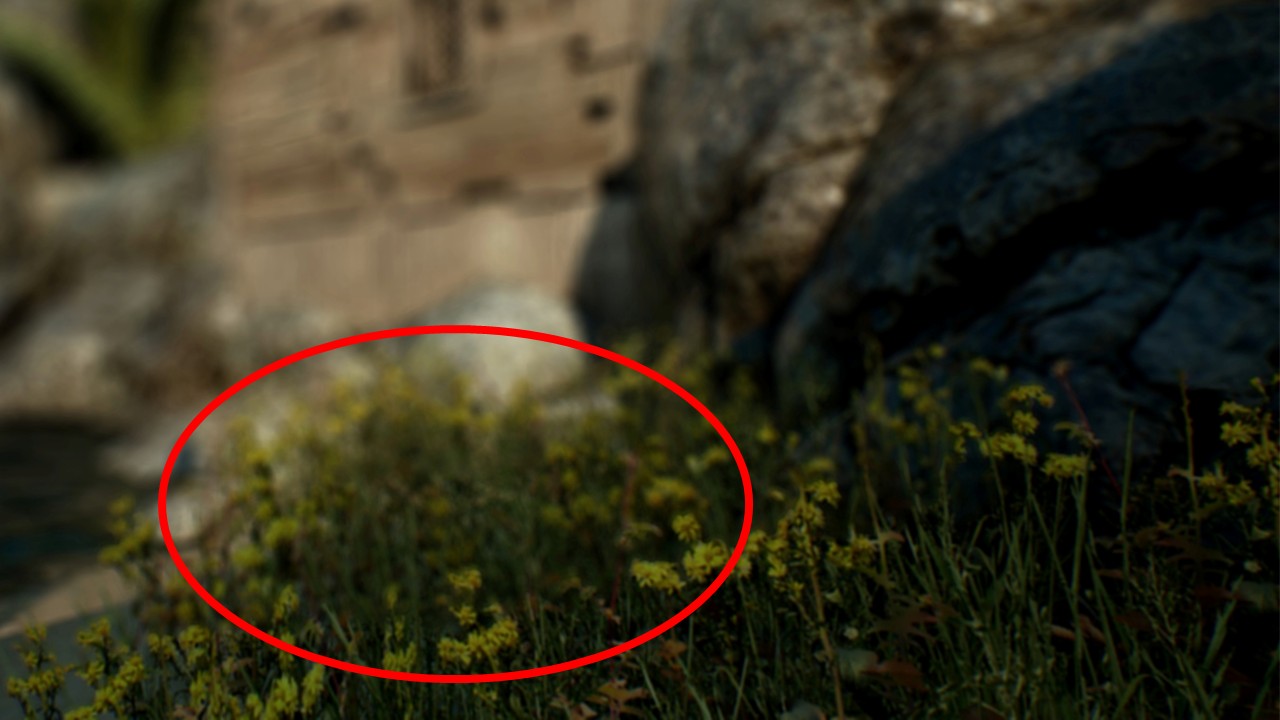} &
    \includegraphics[width=0.48\textwidth]{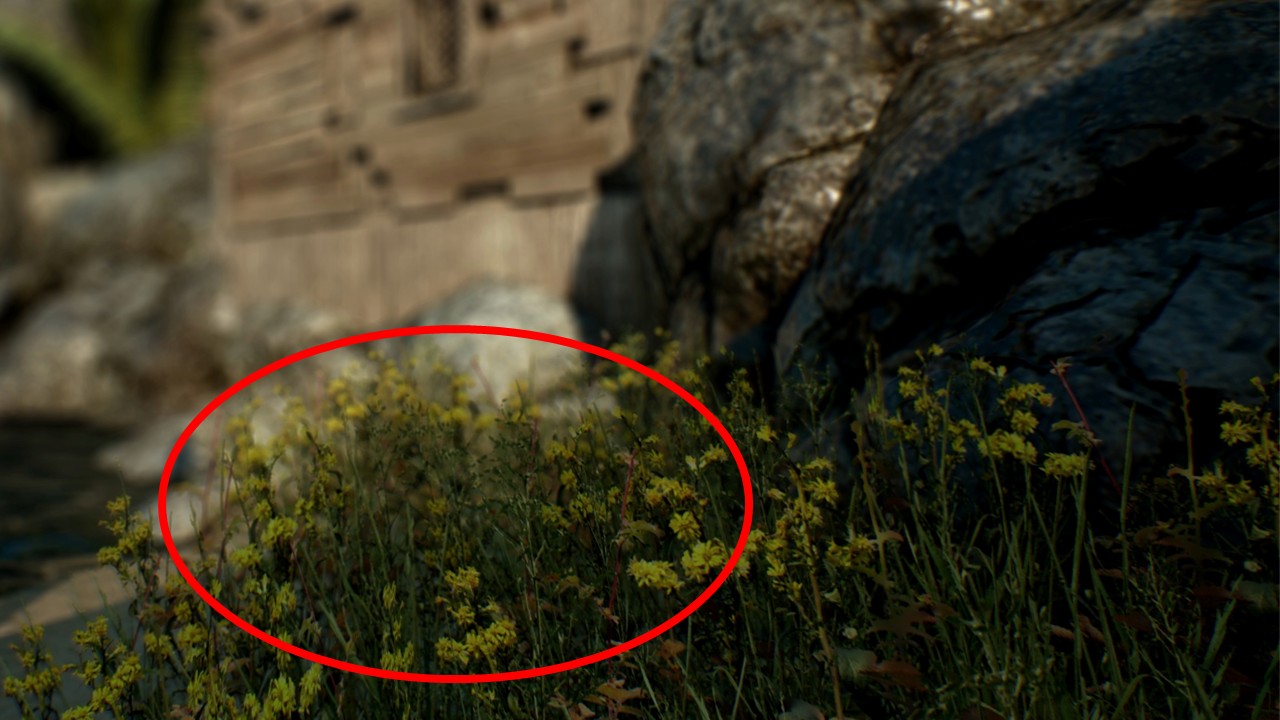} \\
    
    % -------- Sub-captions --------
    {(a) Depth-Anything v2} &
    {(b) Ours}
  \end{tabular}
  \caption{
    \textbf{Application: Lens blur.}
  }

  \label{fig:lens_blur}
\end{figure*}

\begin{figure*}[h]
  \centering
  \setlength{\tabcolsep}{0pt}
  \renewcommand{\arraystretch}{1.0}
  \begin{tabular}{c}
    \includegraphics[width=0.72\textwidth]{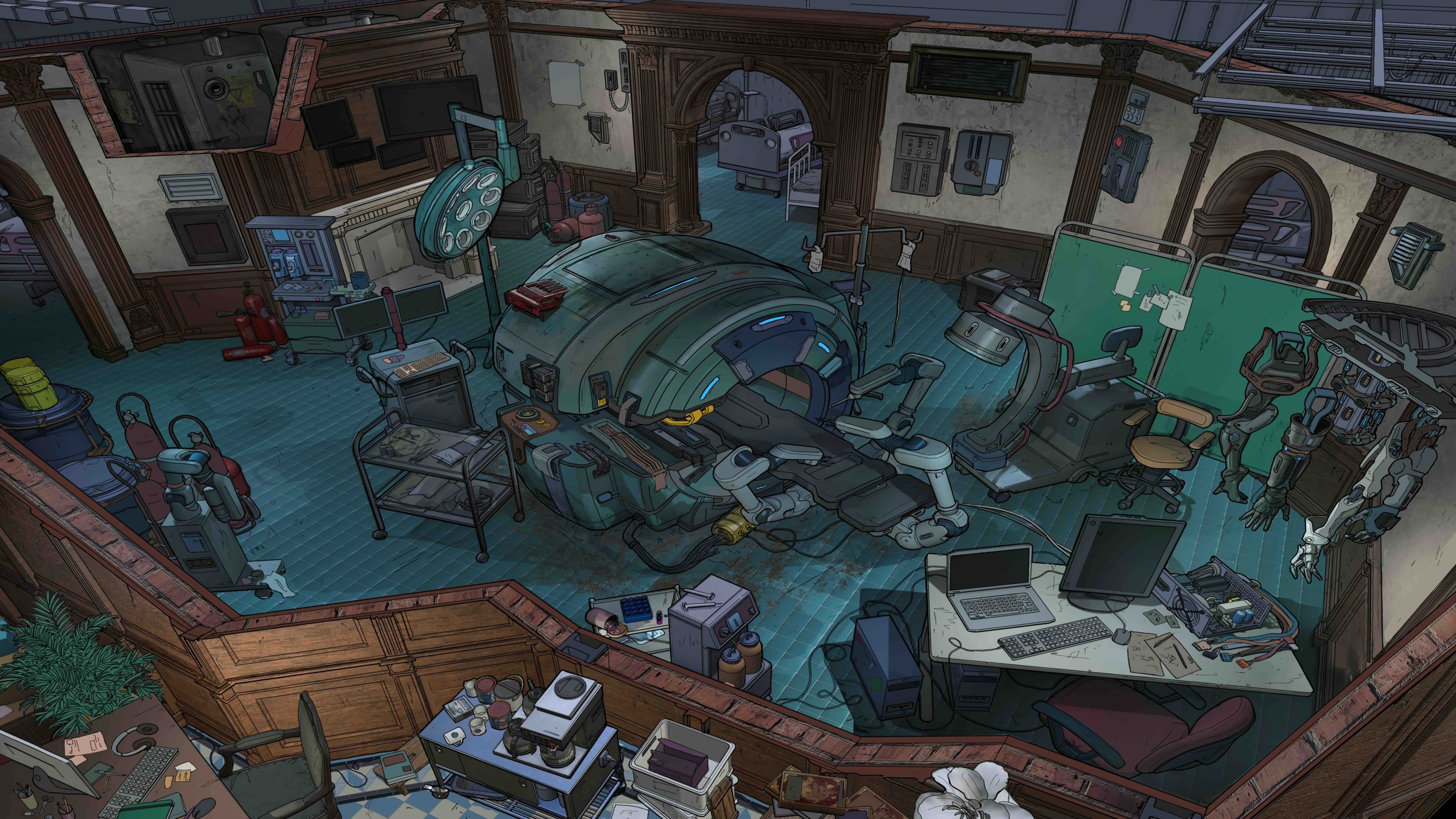} \\[-2pt]
    \includegraphics[width=0.72\textwidth]{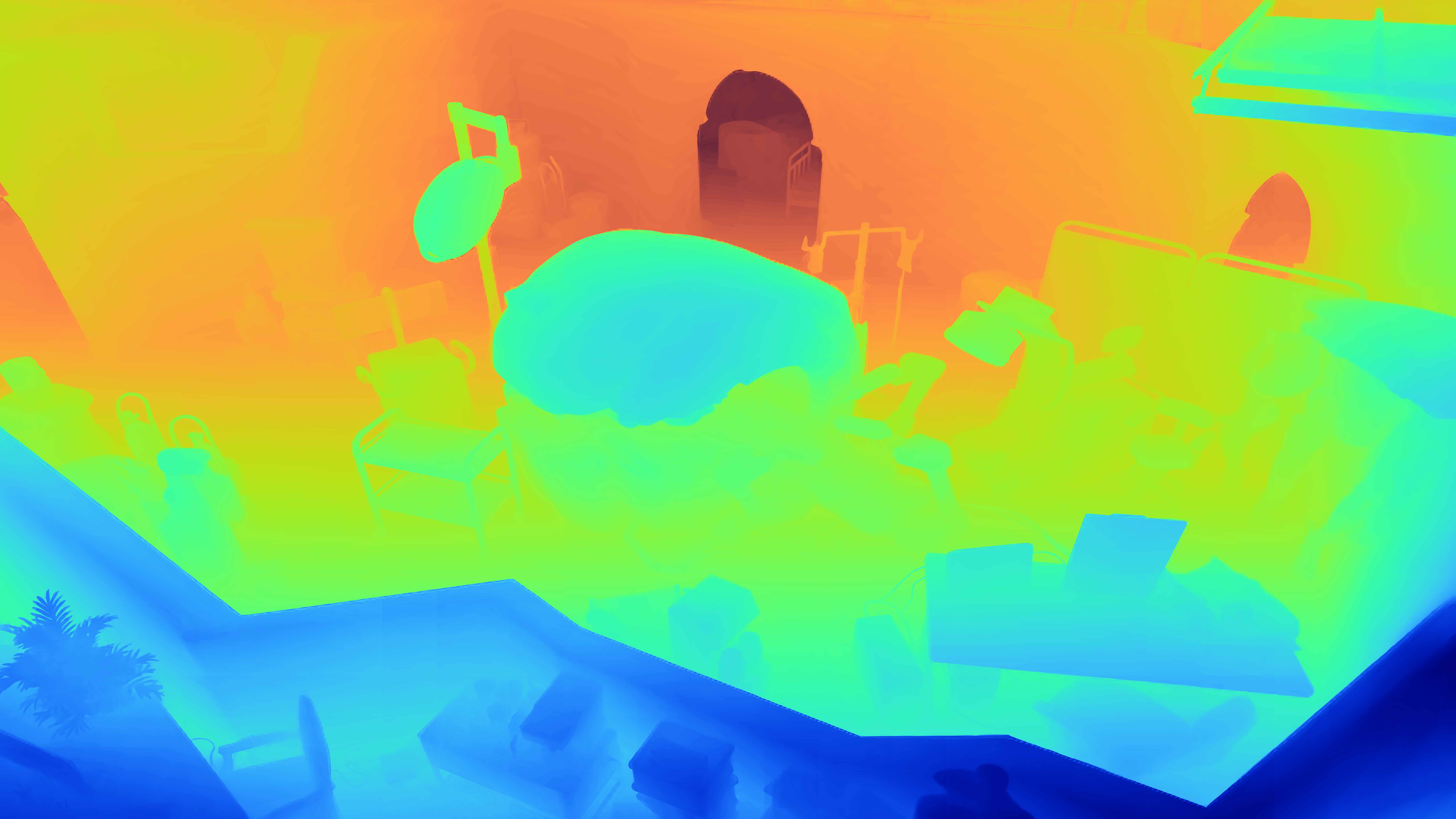} \\[-2pt]
    \includegraphics[width=0.72\textwidth]{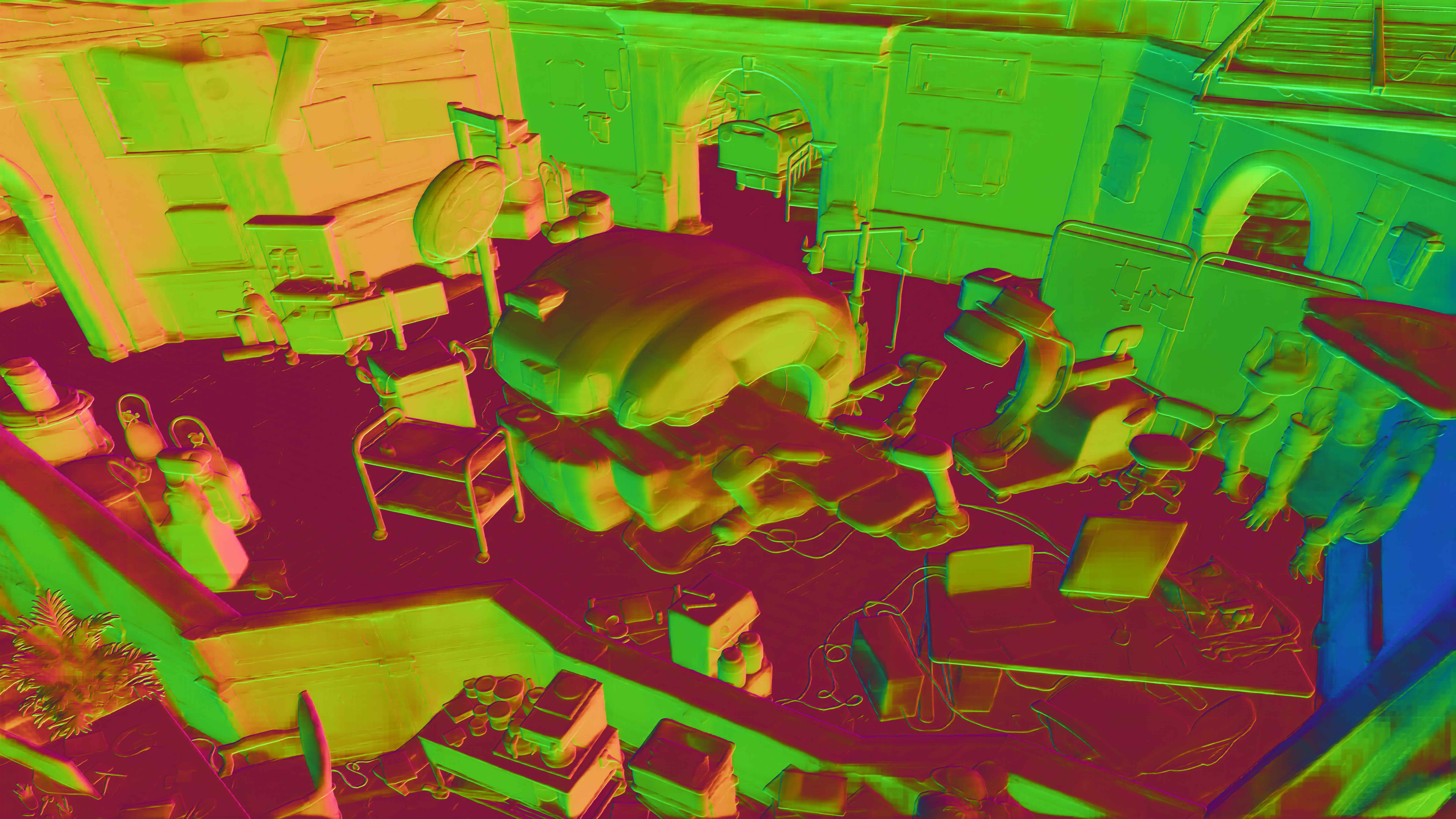} \\[3pt]
  \end{tabular}
  \caption{
  \textbf{Out-of-domain 8K manga-style example.}
  We show an in-the-wild 8K manga-style image, which lies far outside the training distribution, together with our depth and normal predictions. 
  Even under this stylized, out-of-domain setting, our model produces geometrically coherent depth and normals, and the full-resolution visualization allows inspection of fine structures and large-scale consistency.
  }
  \label{fig:8k_solo}
\end{figure*}

\begin{figure*}[h]
  \centering
  \setlength{\tabcolsep}{1pt}
  \renewcommand{\arraystretch}{1.0}
  \begin{tabular}{cccc}
    % ---------------- Row 1: RGB ----------------
    \rotatebox{90}{\hspace{3em} \small RGB } &
    \zoomA{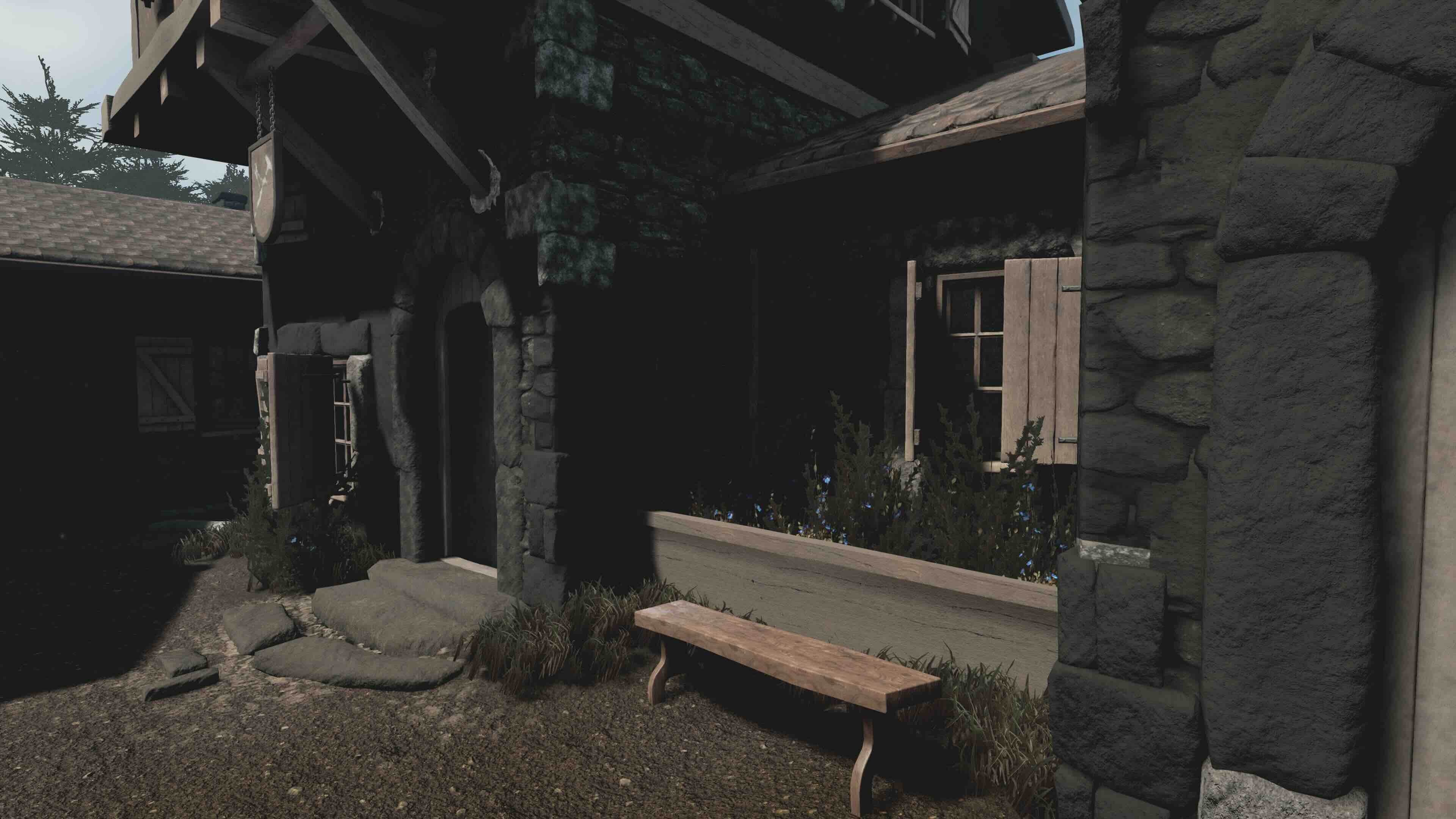} &
    \zoomB{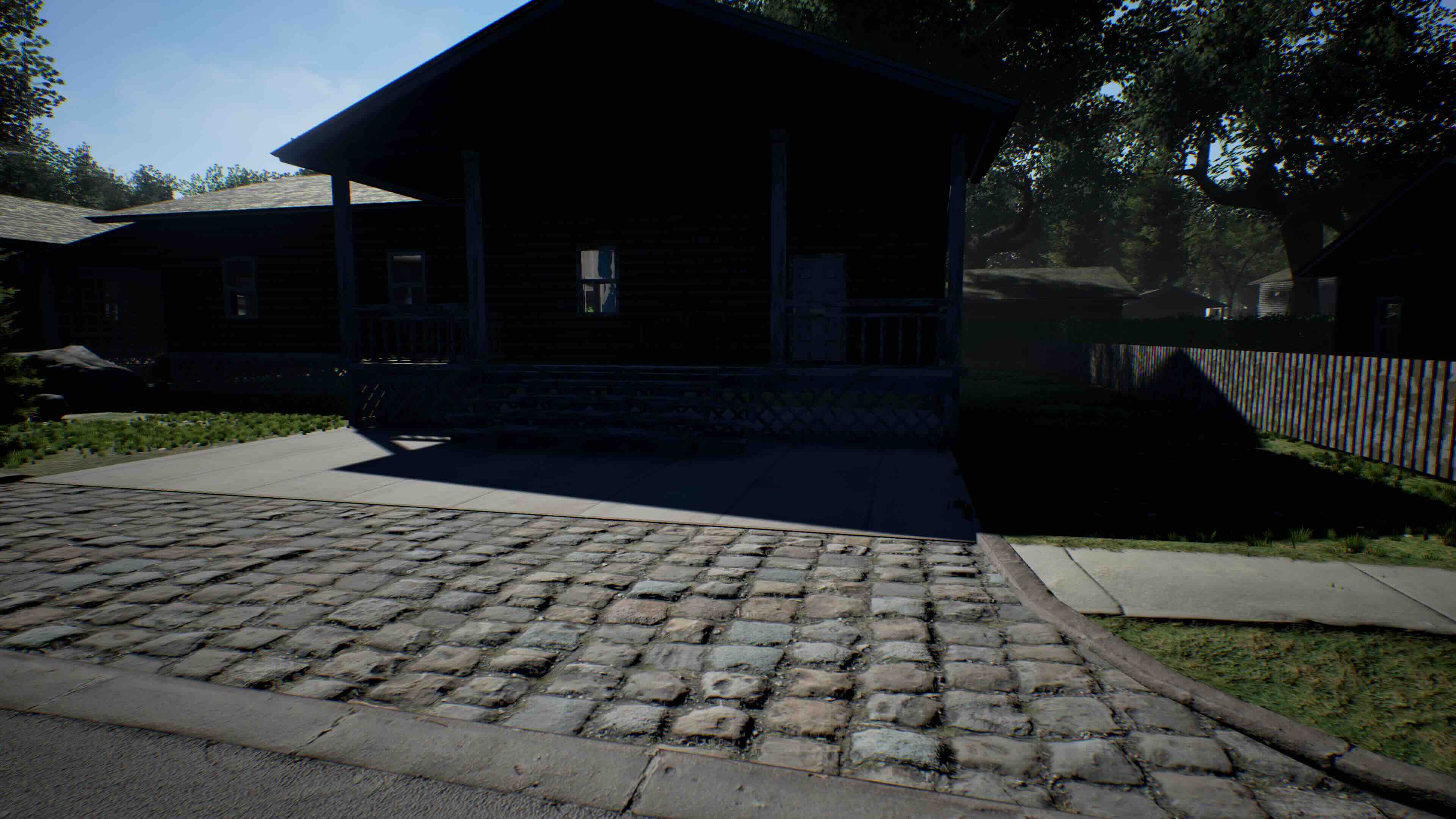} &
    \zoomC{sec/fig/u4k_normal/18.jpg} \\[-2pt]
    % ---------------- Row 2: Coarse Depth ----------------
    \rotatebox{90}{\hspace{0.8em}\small Depth-Anything V2} &
    \zoomA{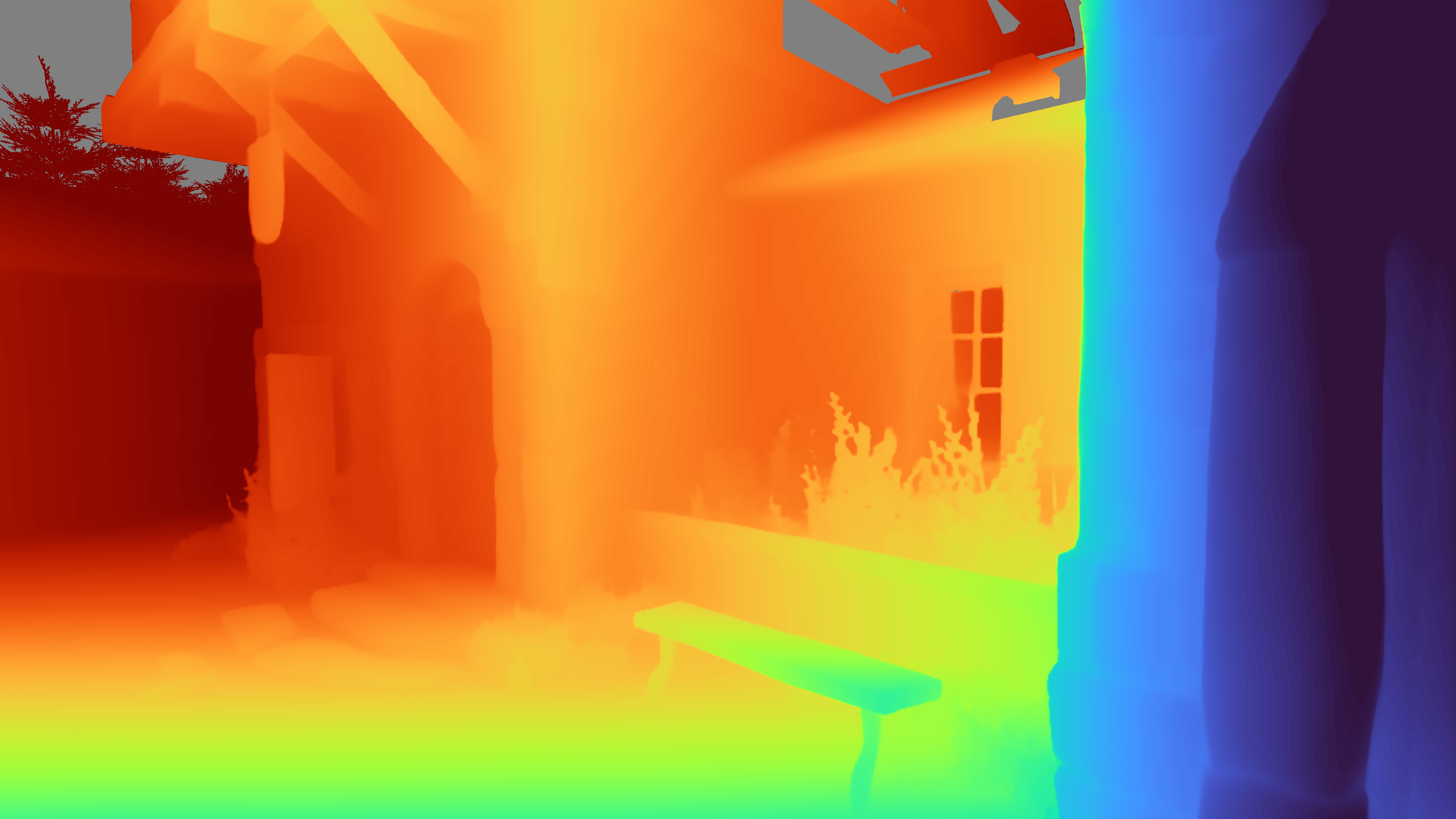} &
    \zoomB{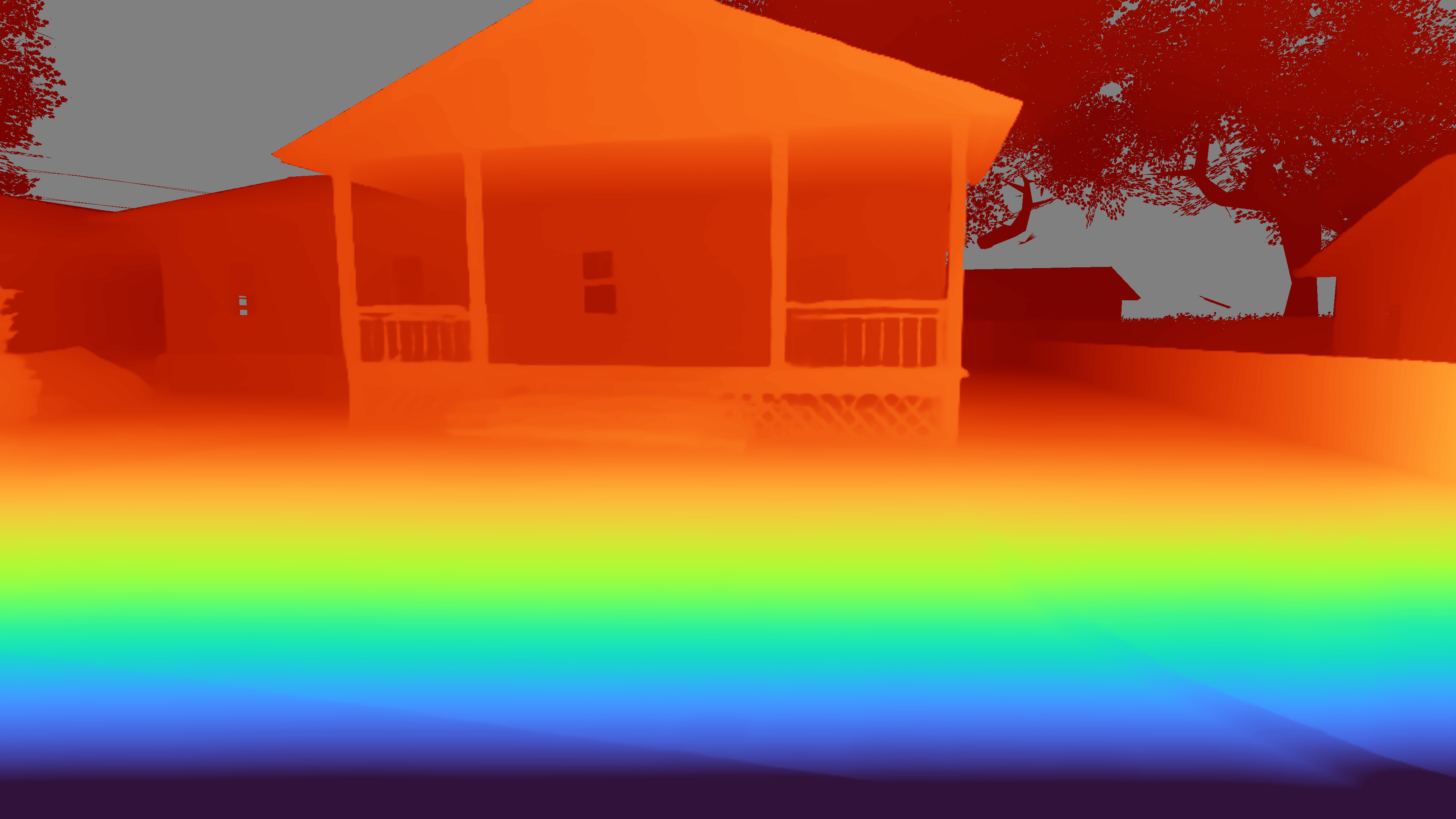} &
    \zoomC{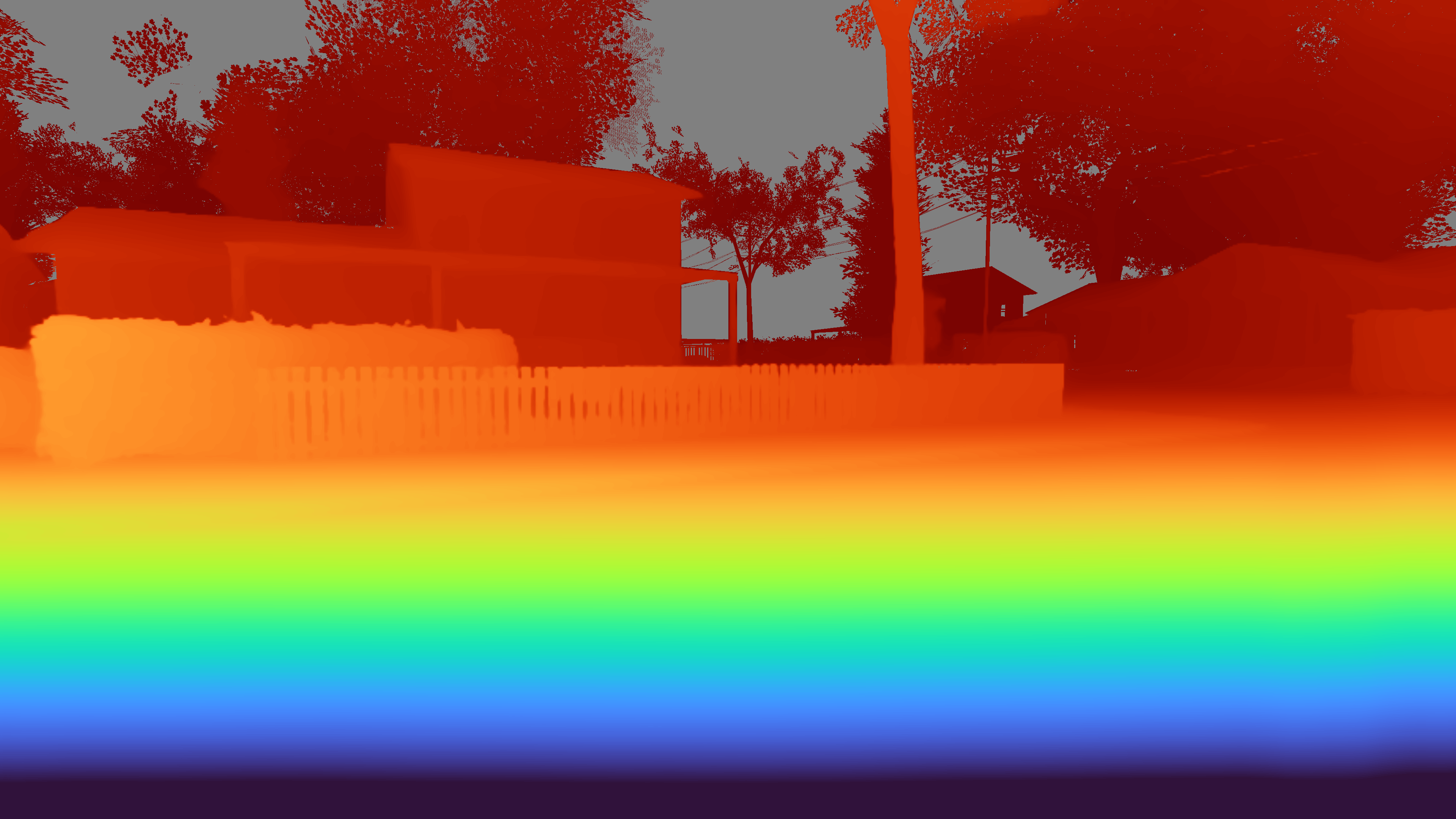} \\[-2pt]
    % ---------------- Row 3: PatchRefiner ----------------
    \rotatebox{90}{\hspace{1.8em}\small PatchRefiner} &
    \zoomA{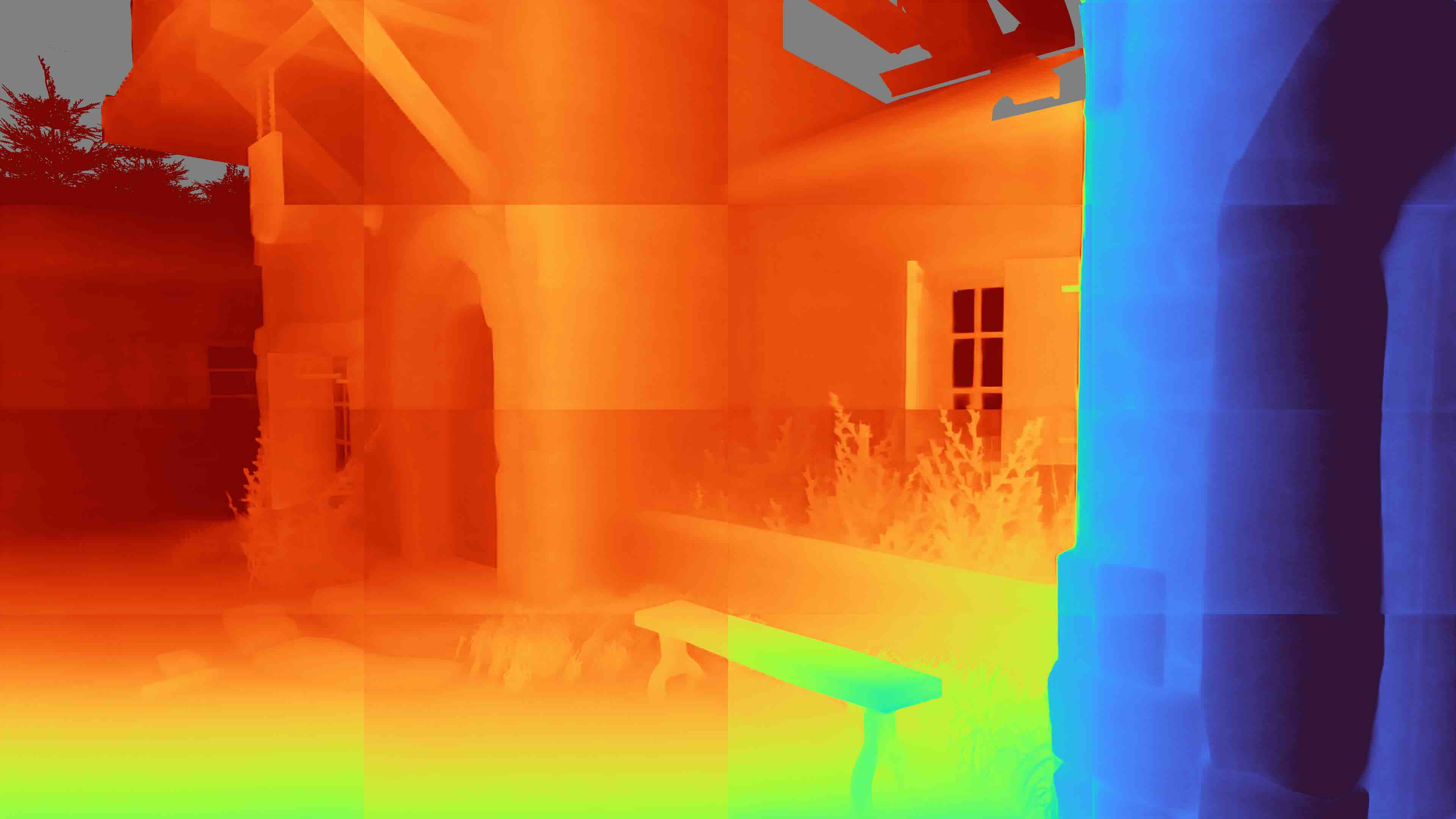} &
    \zoomB{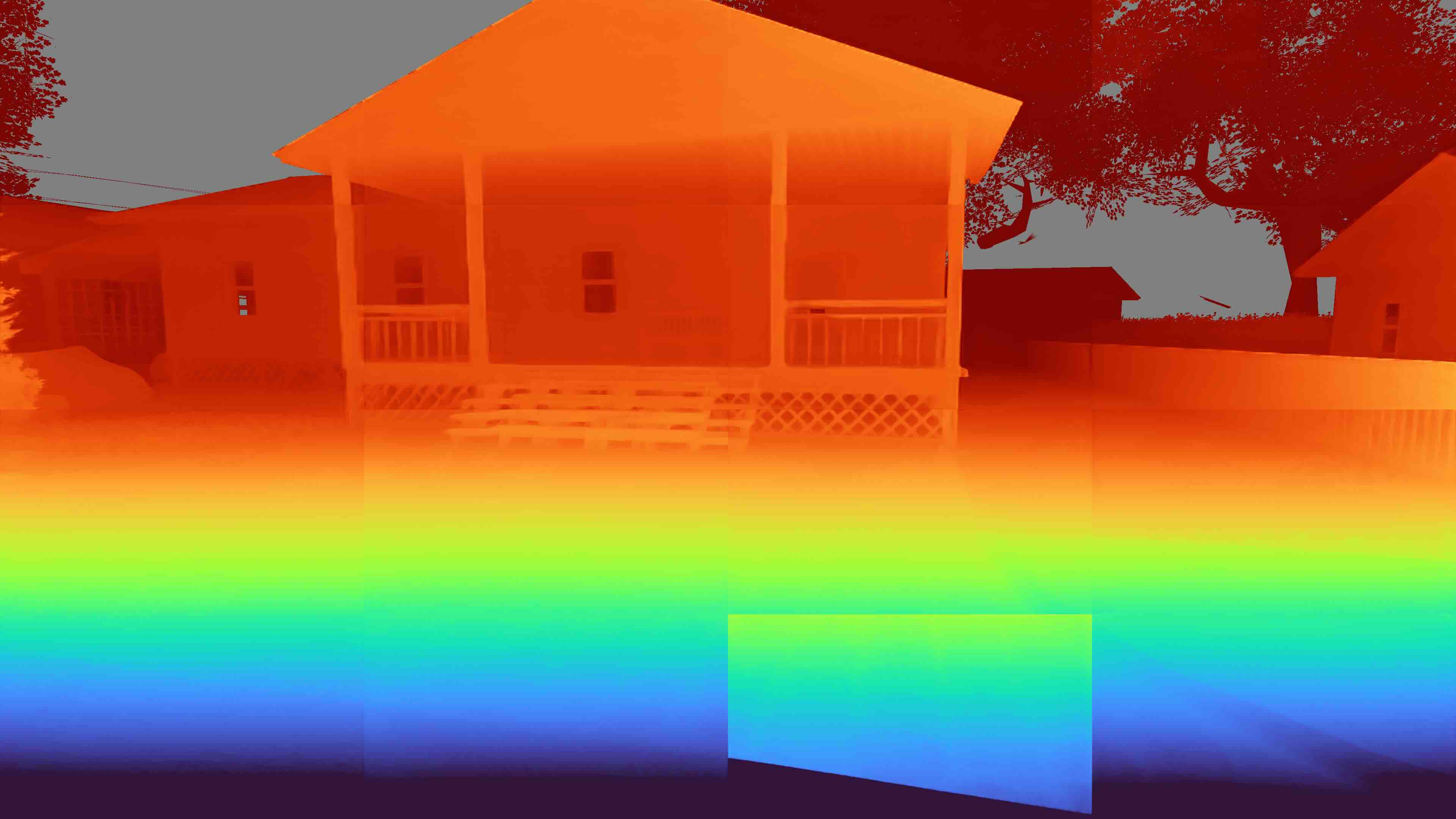} &
    \zoomC{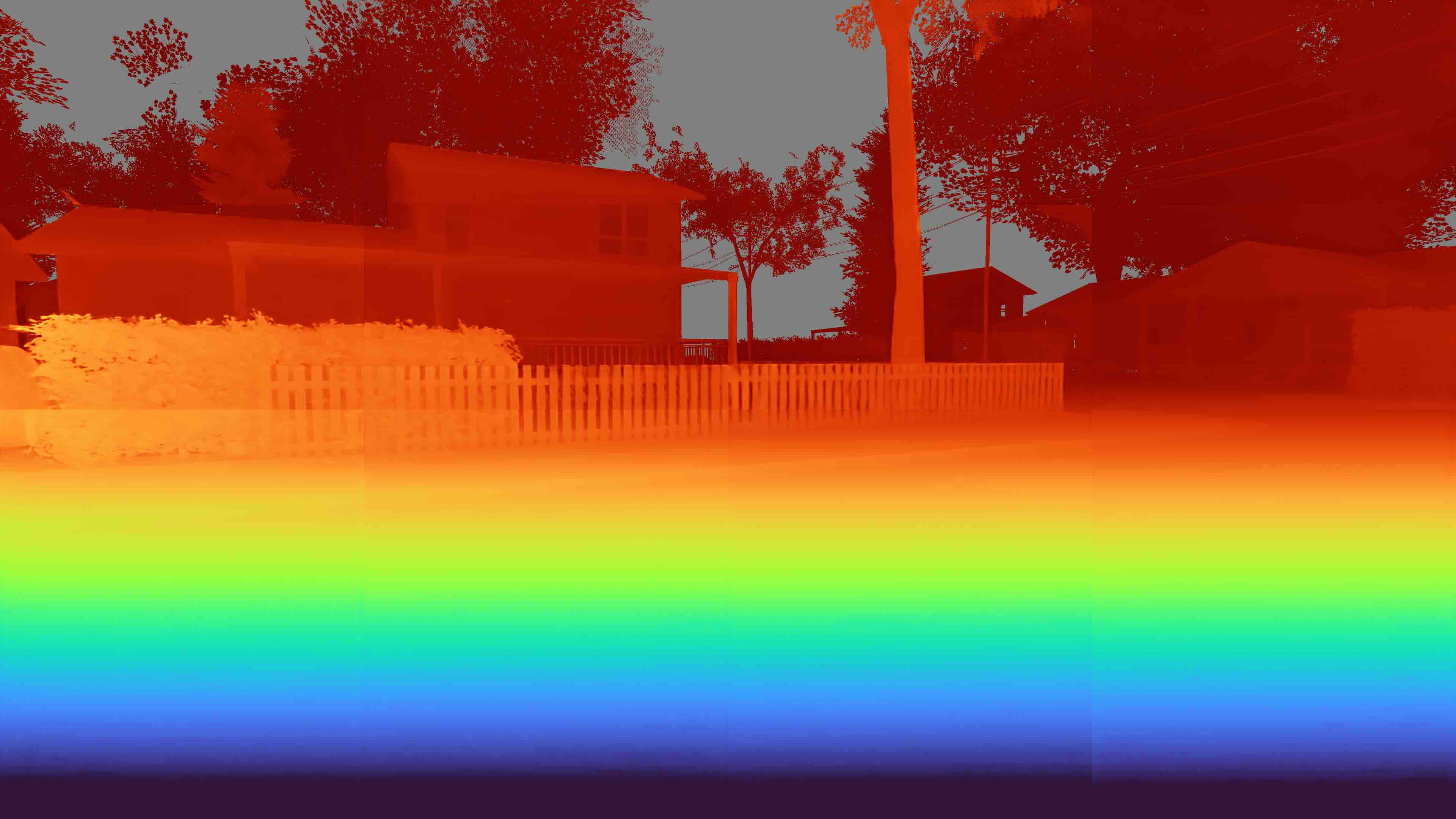} \\[-2pt]
    % ---------------- Row 4: PRO ----------------
    \rotatebox{90}{\hspace{3.5em}\small PRO} &
    \zoomA{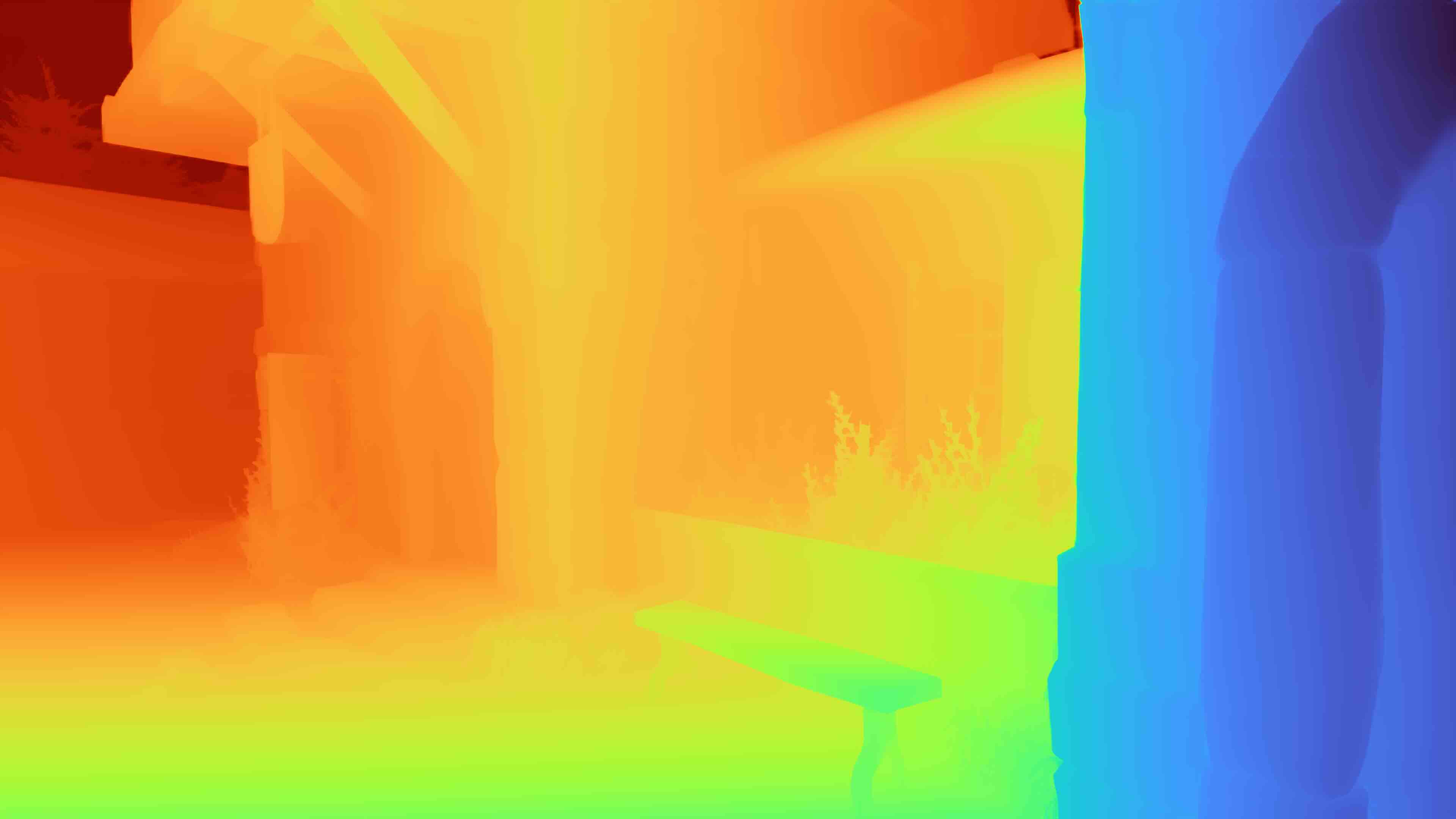} &
    \zoomB{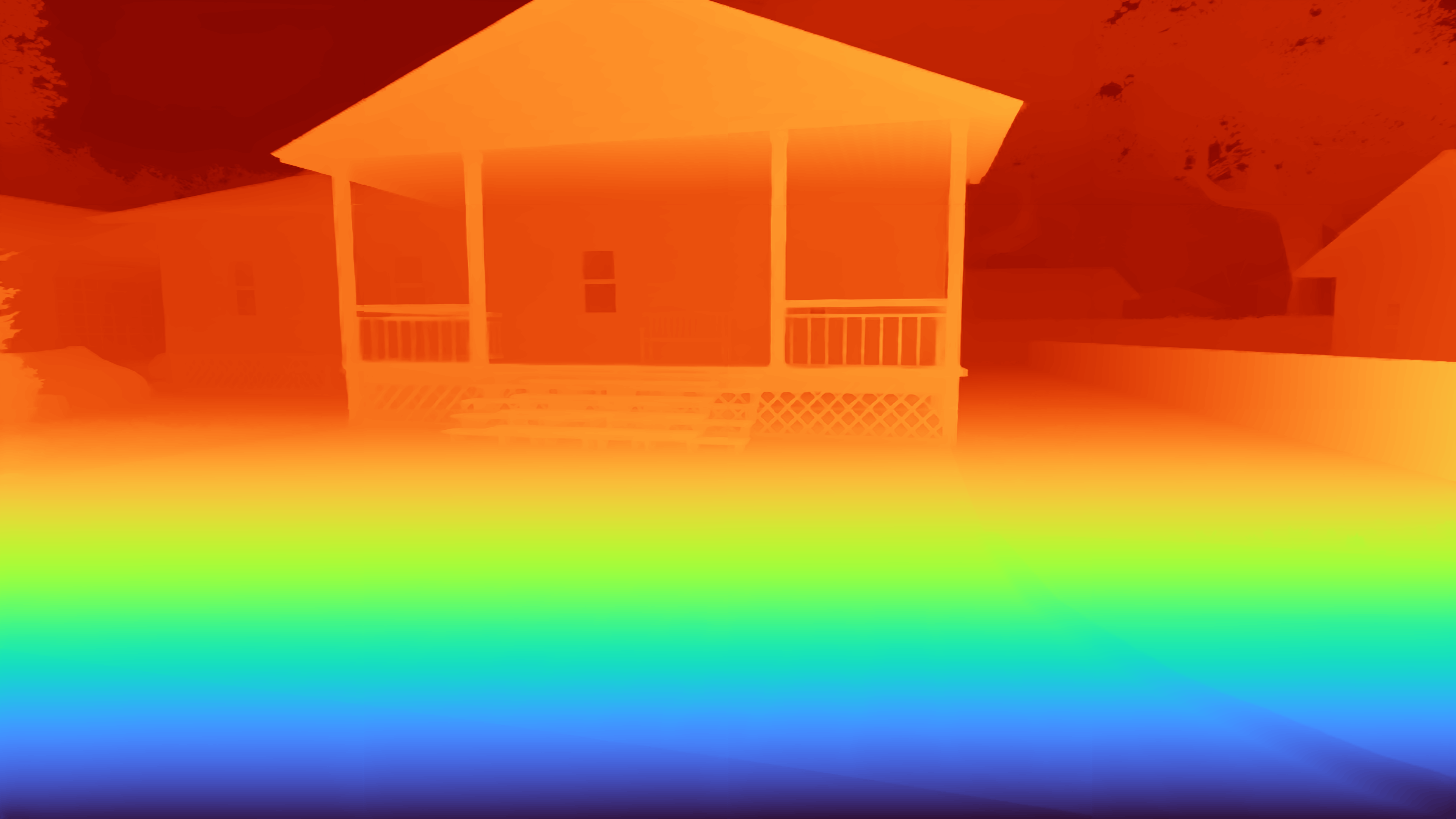} &
    \zoomC{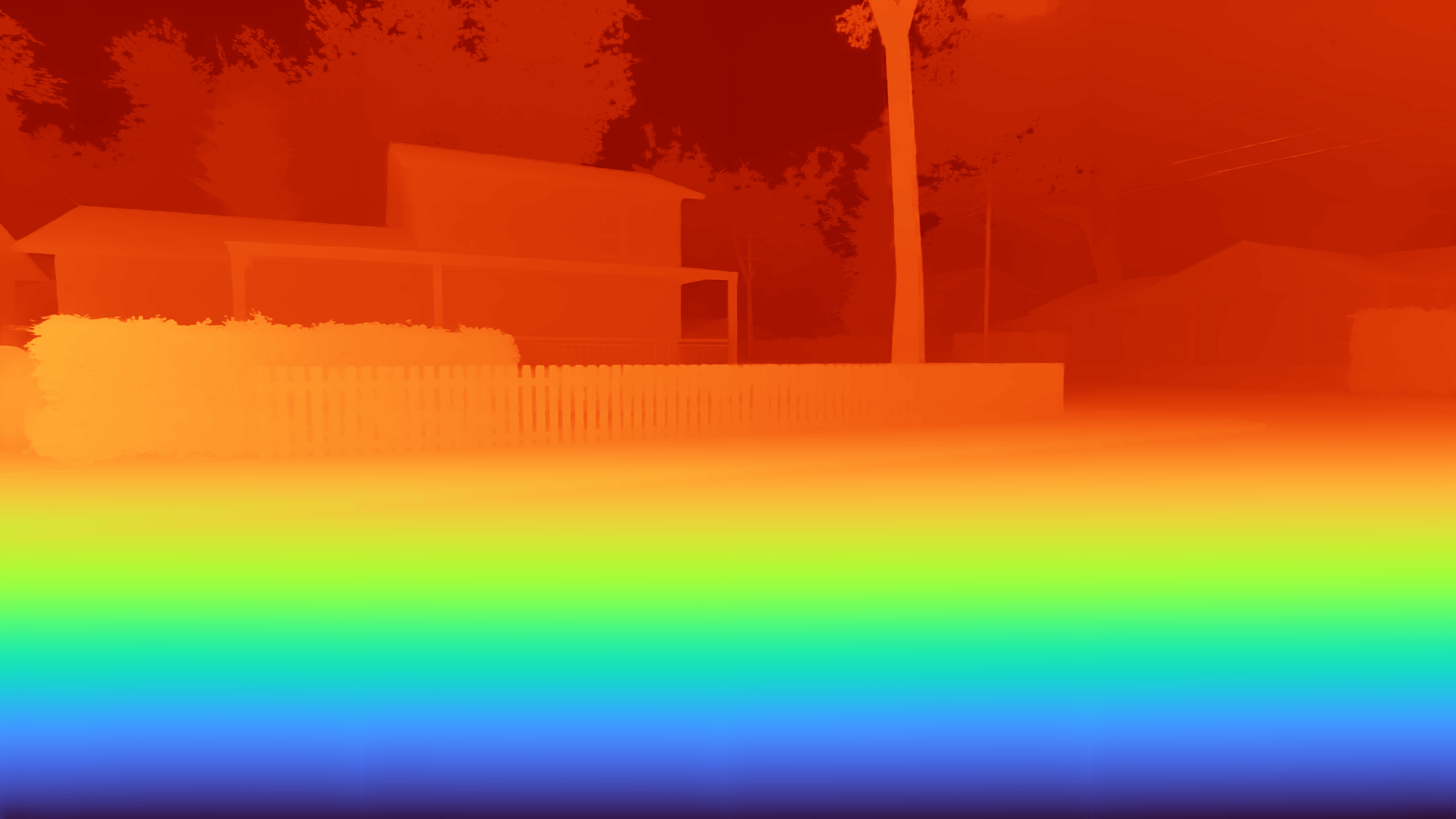} \\[-2pt]
    % ---------------- Row 5: Our Depth ----------------
    \rotatebox{90}{\hspace{2.2em}\small Our Depth} &
    \zoomA{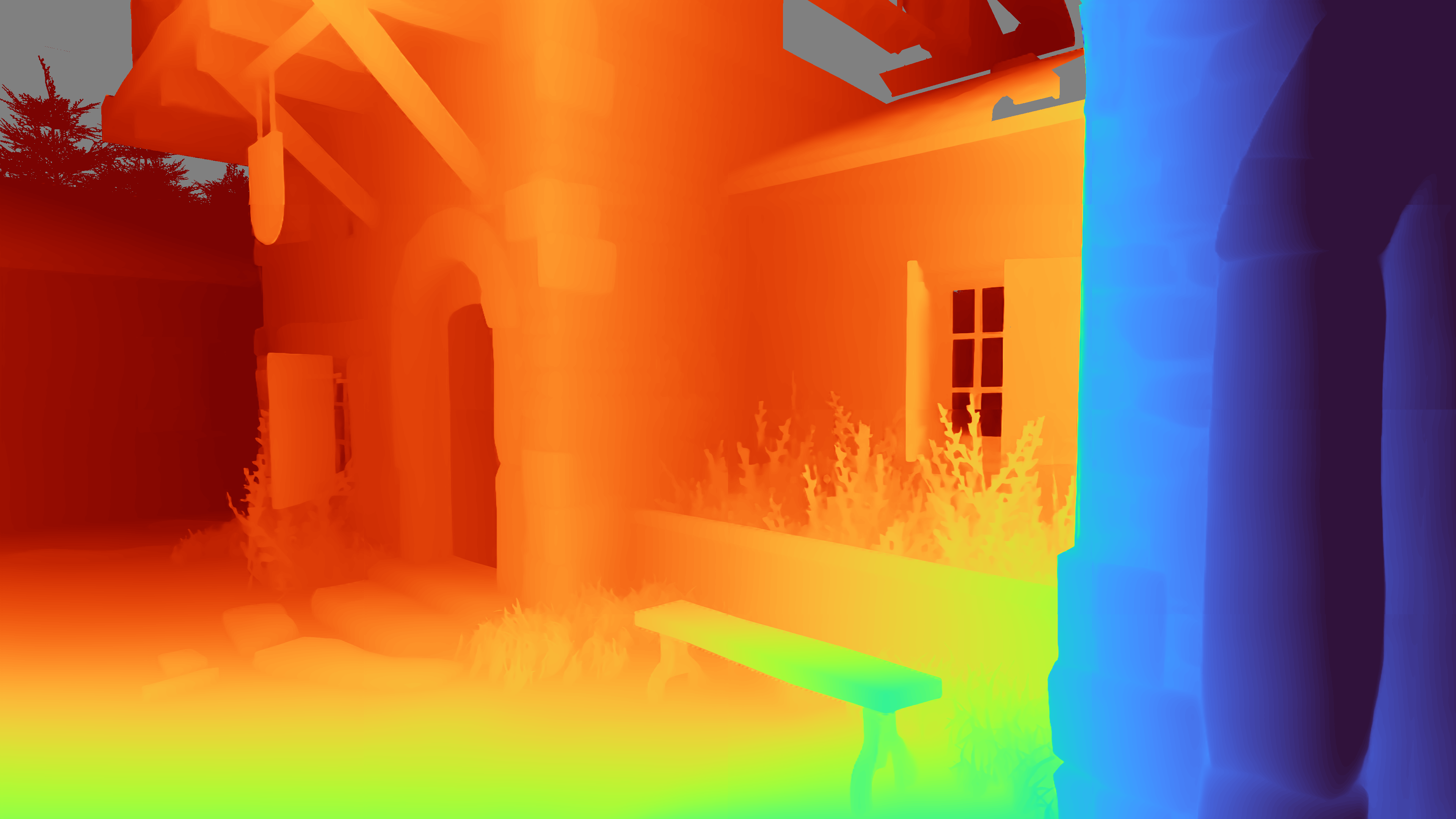} &
    \zoomB{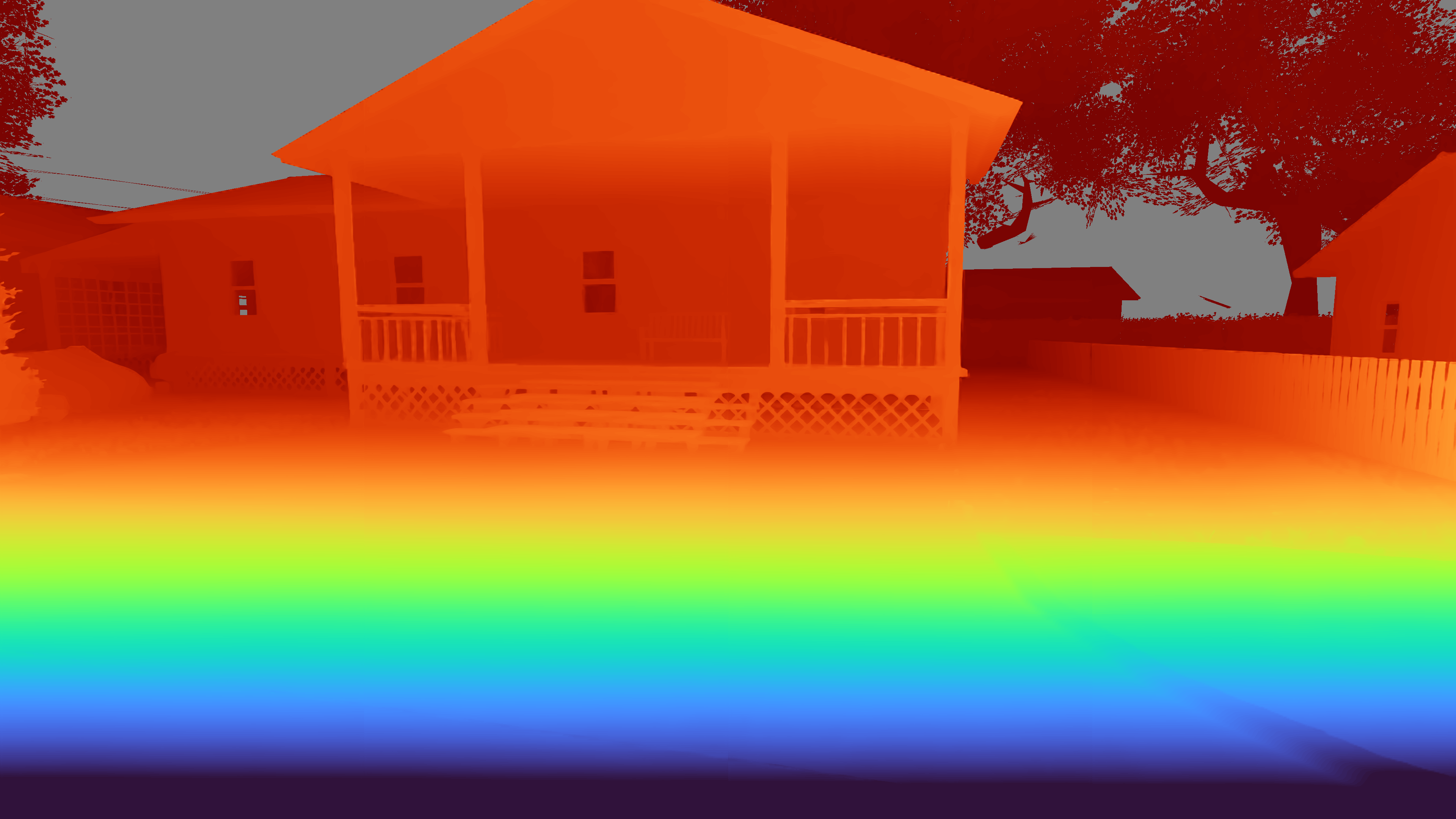} &
    \zoomC{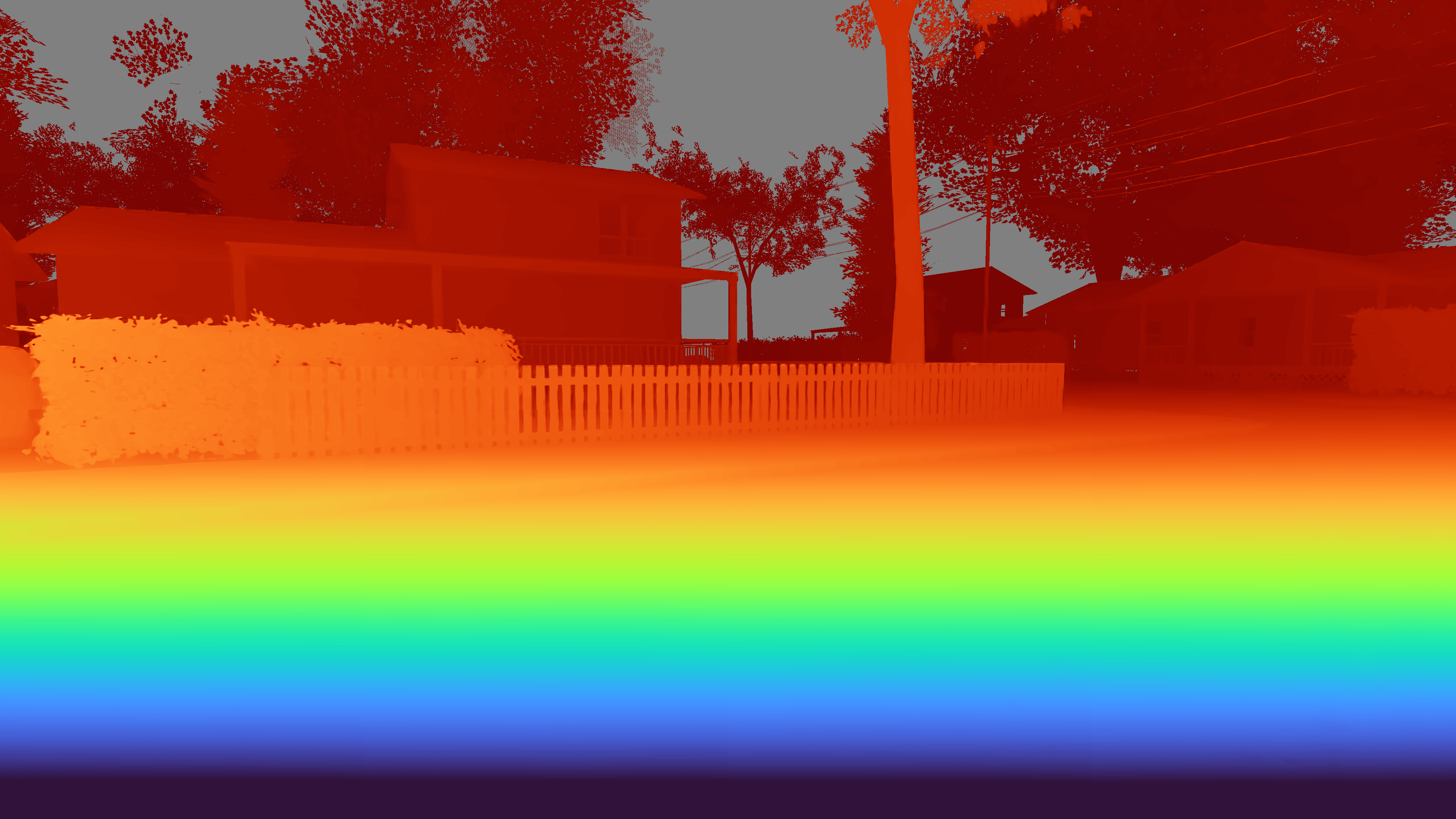} \\[-2pt]
    % ---------------- Row 6: Coarse Normal ----------------
    \rotatebox{90}{\hspace{1.5em}\small Metric3D V2} &
    \zoomA{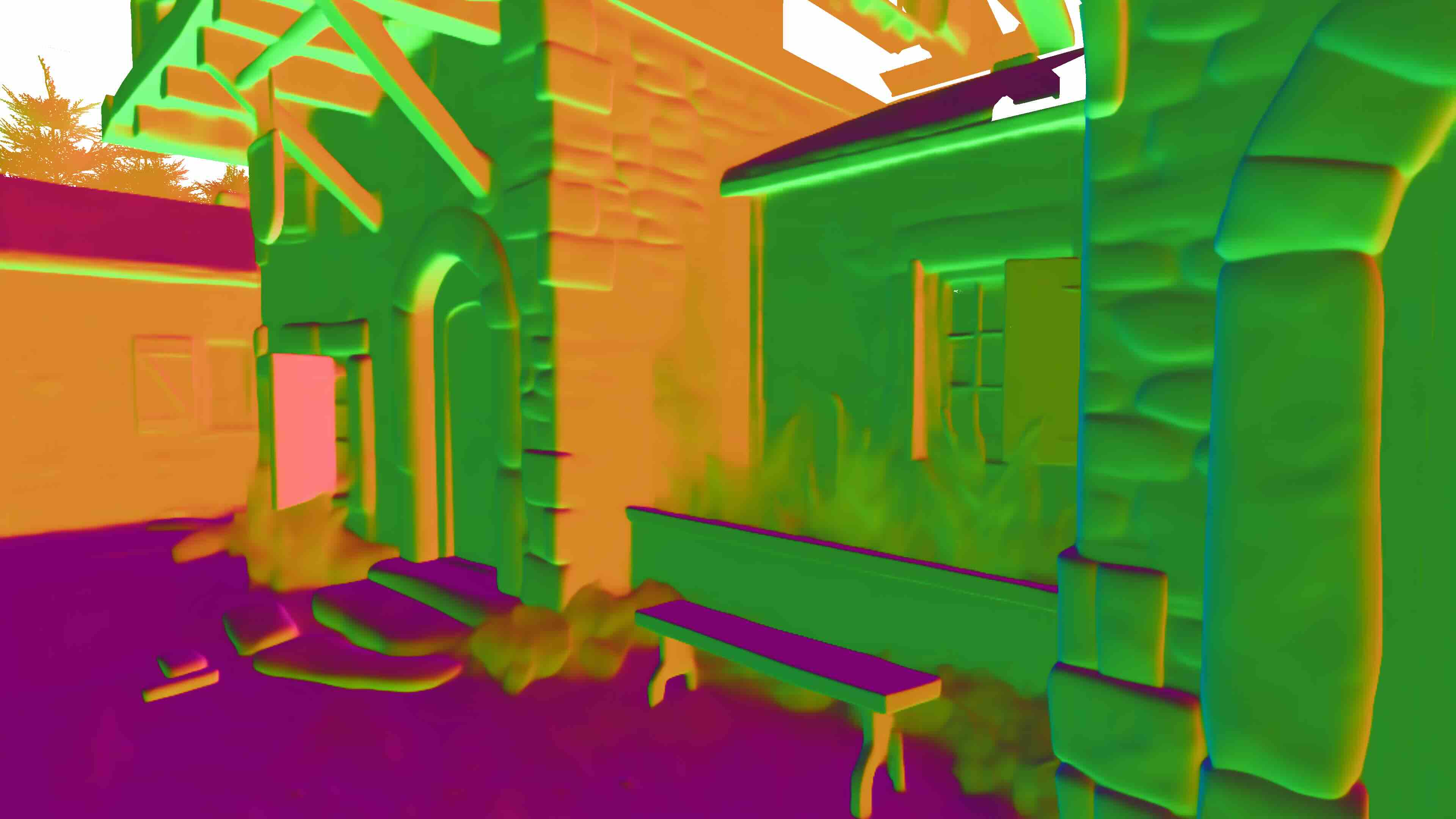} &
    \zoomB{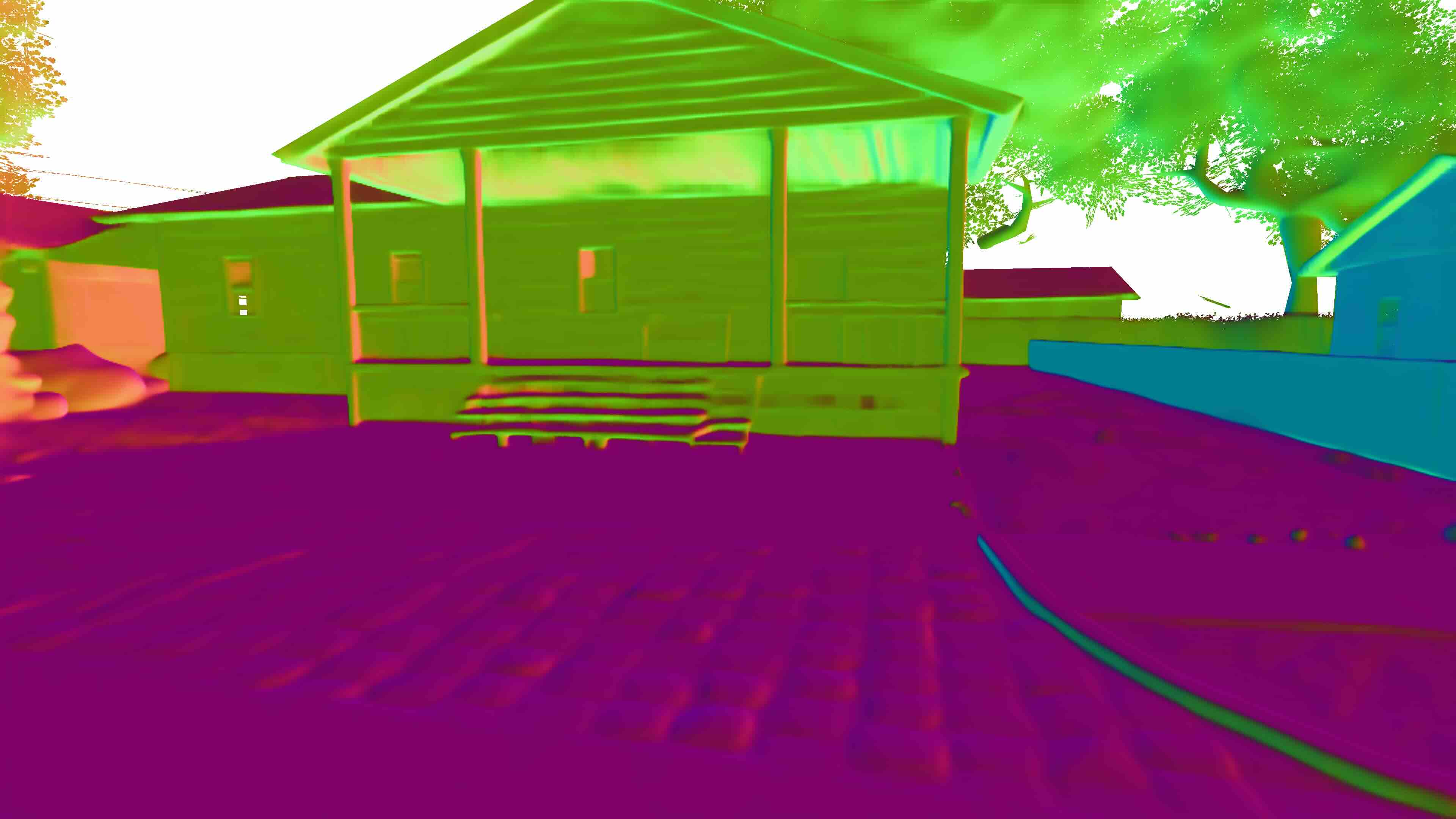} &
    \zoomC{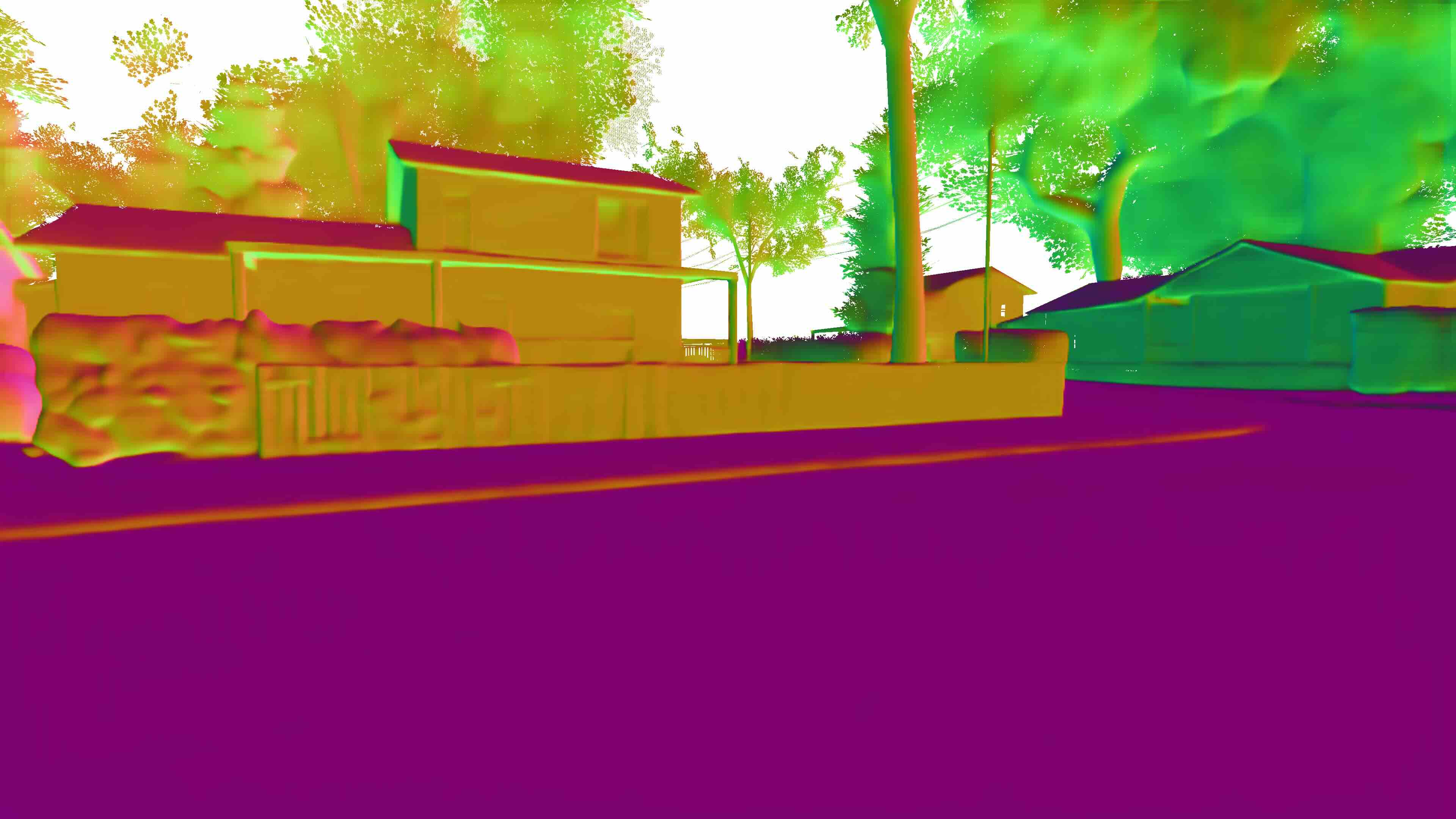} \\[-2pt]
    % ---------------- Row 7: Our Normal ----------------
    \rotatebox{90}{\hspace{2em}\small Our Normal} &
    \zoomA{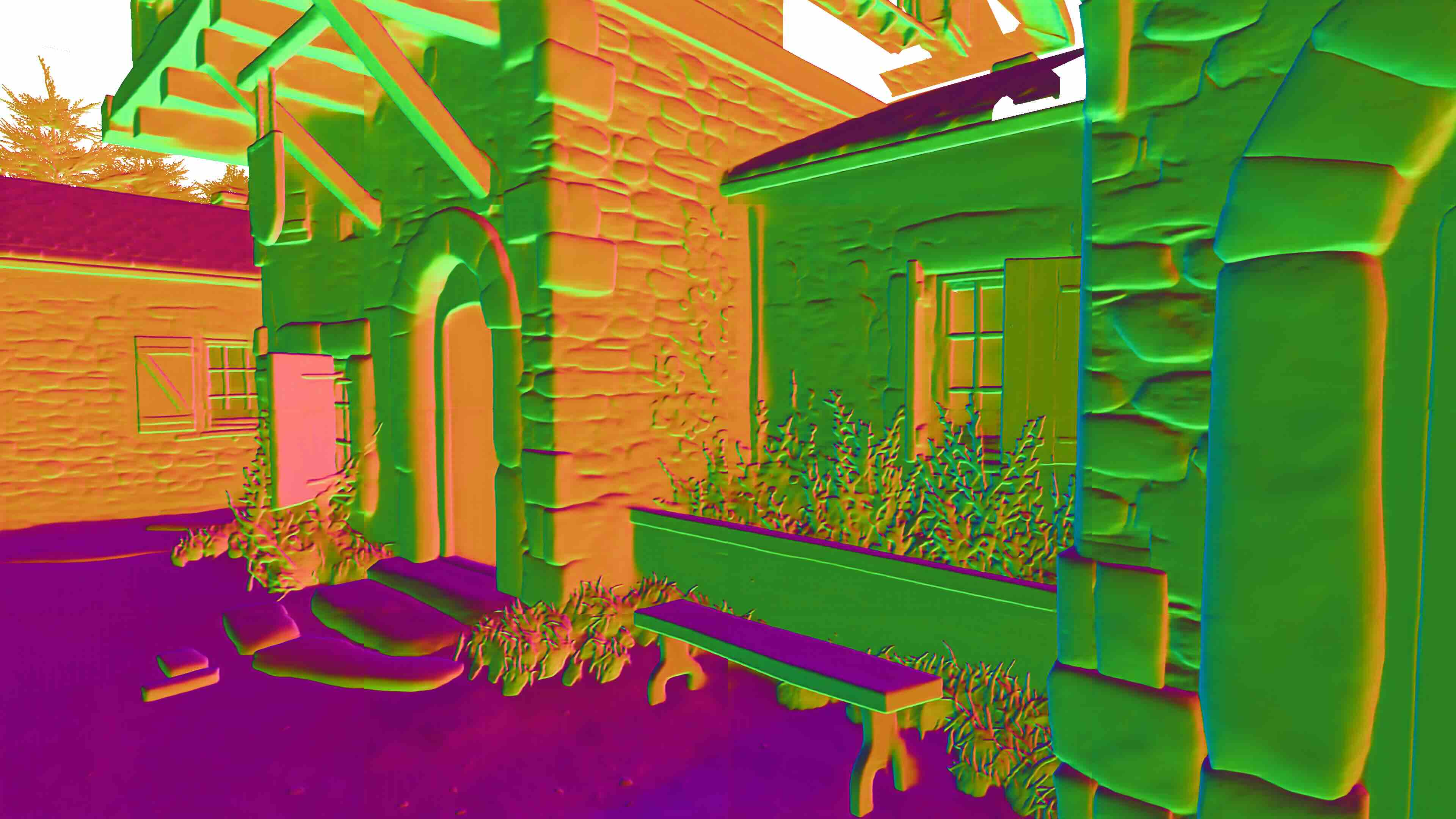} &
    \zoomB{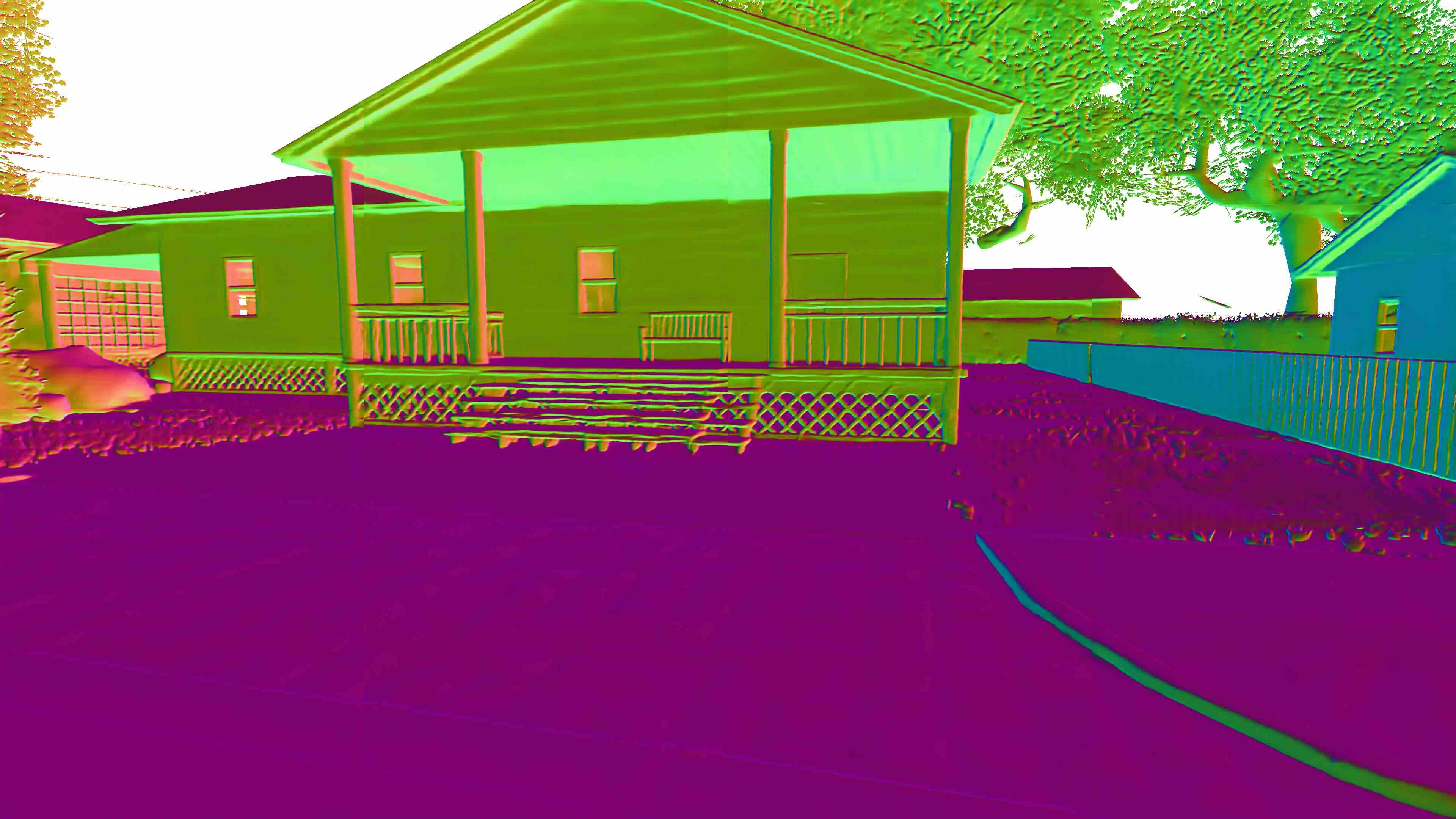} &
    \zoomC{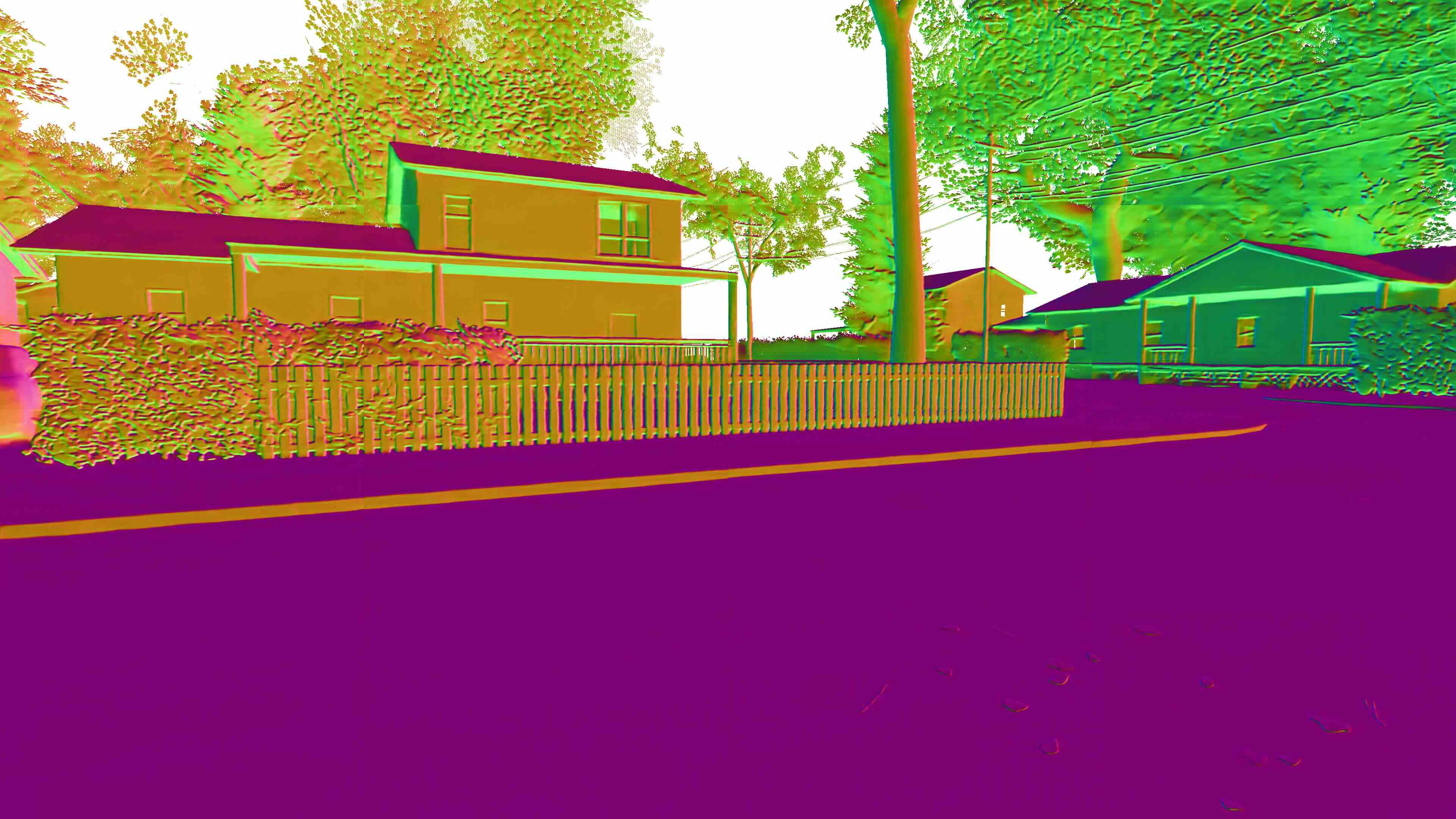} \\[2pt]
    & \small Sample 1 & \small Sample 2 & \small Sample 3 \\
  \end{tabular}
  \caption{
  \textbf{More qualitative results on UnrealStereo4K.}
  We show samples from the UnrealStereo4K~\cite{u4k} dataset. 
  Each column corresponds to one scene, and rows show the RGB input, coarse depth prediction by Depth-Anything V2~\cite{yang2025depthanythingv2}, depth refinements from PatchRefiner~\cite{li2024patchrefiner} and PRO~\cite{kwon2025onelook}, our refined depth prediction, coarse normal prediction by Metric3D V2~\cite{hu2025metric3dv2}, and our refined normal prediction.
  For each column, a consistent zoom-in inset highlights a region of interest across all methods.
  }
  \label{fig:u4k_qual_10x5}
\end{figure*}

We simulate a lens blur effect using depth maps predicted by DepthAnything (DA)-v2 and our method to blur the background. As shown in Fig.~\ref{fig:lens_blur} and in Fig. 4 of the main paper, our method produces sharper fine-grained details (see highlighted regions), which allows part of the plants to be blurred while other parts remain in focus. In contrast, due to over-smoothed depth predictions, DA-v2 fails to separate these structures and thus cannot achieve the same blur effect.

\section{Limitations and Future Work}

While our framework demonstrates strong performance for high-resolution depth and normal estimation, it still has several limitations. First, our model assumes mostly diffuse surfaces and struggles with strongly reflective objects such as mirrors, where the mirror region is misinterpreted as a continuation of the surrounding geometry and yields local depth and normal errors (Fig.~\ref{fig:limitation_mirror}). Second, our current framework is coupled with two specific coarse geometry backbones, namely Depth-Anything V2 for depth and Metric3D V2 for surface normals. As a result, the quality and characteristics of these coarse predictors directly influence the final refined outputs. 

In future work, we aim to relax this dependency and turn our approach into a more generic, plug-and-play refinement module that can be seamlessly attached to a wide range of coarse predictors. 
We believe such a flexible refinement framework would further broaden the applicability of our method across different tasks, datasets, and deployment scenarios.

\end{document}